%% file: main.tex
\documentclass[review,10pt,times,number]{elsarticle}
\usepackage[a4paper,top=4.3cm,bottom=4.3cm,left=4.8cm,right=4.8cm]{geometry}
\usepackage[font=footnotesize]{caption}
\usepackage{setspace}
\usepackage{etoolbox}
\makeatletter%
\patchcmd{\MaketitleBox}{\Large\@title}{\fontsize{14}{17}\selectfont\@title}{}{}
\patchcmd{\MaketitleBox}{\normalsize\elsauthors}{\footnotesize\elsauthors}{}{}
\makeatother

\usepackage[T1]{fontenc}
\usepackage[utf8]{inputenc}
\usepackage{amsmath,amssymb}
\usepackage{graphicx}
\usepackage{booktabs}
\usepackage{algorithm}
\usepackage{algpseudocode}
\usepackage{array}
\usepackage{tabularx}
\usepackage{rotating}
\usepackage{seqsplit}
\usepackage{microtype}
\usepackage{xurl}
\usepackage{xcolor}
\usepackage[hidelinks]{hyperref}
\usepackage[capitalise,noabbrev]{cleveref}
\usepackage{orcidlink}

\input{macros}

\InputIfFileExists{xlabels-main}{}{}

\journal{Pattern Recognition}

\begin{document}

\begin{frontmatter}

\title{Self-prediction residuals improve single-pass detection of wrong-object boxes in grounded vision-language models}

\author[aff1]{\footnotesize Enhao Cui\,\orcidlink{0009-0004-1776-4047}\corref{cor1}}
\ead{2500160612@smail.cczu.edu.cn}
\author[aff1]{\footnotesize Junzhi Yang}
\author[aff1]{\footnotesize Yaoshun Lin}
\author[aff1]{\footnotesize Yuhan Feng}
\author[aff1]{\footnotesize Zijian Chen}
\cortext[cor1]{Corresponding author.}
\affiliation[aff1]{organization={Changzhou University},
            city={Changzhou},
            postcode={213164},
            state={Jiangsu},
            country={China}}

\begin{abstract}
\input{sections/abstract}
\end{abstract}

\begin{keyword}
\input{sections/keywords}
\end{keyword}

\end{frontmatter}

\input{sections/introduction}
\input{sections/related}
\input{sections/method}
\input{sections/setup}
\input{sections/plugin}
\input{sections/measures}
\input{sections/discussion}
\input{sections/conclusion}

\input{sections/declarations}

\setlength{\bibsep}{0pt}
\bibliographystyle{elsarticle-num}
\bibliography{refs}

\input{sections/vitae}

\end{document}


\begin{center}
{\large\bfseries Supplementary Material}\\[6pt]
{\normalsize Self-prediction residuals improve single-pass detection of wrong-object boxes in grounded vision-language models}\\[6pt]
{\footnotesize Enhao Cui, Junzhi Yang, Yaoshun Lin, Yuhan Feng, Zijian Chen}
\end{center}

\noindent Section, table, figure and algorithm numbers without the prefix S refer to the main manuscript, where the detector is called SPRED (self-prediction residual error detector).

\input{sections/supplement}

\setlength{\bibsep}{0pt}
\bibliographystyle{elsarticle-num}
\bibliography{refs}

%% file: macros.tex
\newcommand{\qsmall}{Qwen2.5-VL-3B}
\newcommand{\qlarge}{Qwen2.5-VL-7B}
\newcommand{\qfour}{Qwen3-VL-4B}
\newcommand{\qeight}{Qwen3-VL-8B}
\newcommand{\qivl}{InternVL2.5-8B-MPO}
\newcommand{\qllava}{LLaVA-1.5-7B}

%% file: xlabels-main.tex
\newlabel{sup:repro}{{S1}{1}{}{}{}}
\newlabel{sup:repro@cref}{{[section][1][]S1}{[1][1][]1}{}{}{}}
\newlabel{sup:prompts}{{S1}{1}{}{}{}}
\newlabel{sup:prompts@cref}{{[section][1][]S1}{[1][1][]1}{}{}{}}
\newlabel{tab:prompts}{{S1.1}{3}{}{}{}}
\newlabel{tab:prompts@cref}{{[table][1][1]S1.1}{[1][2][]3}{}{}{}}
\newlabel{tab:costs}{{S1.2}{4}{}{}{}}
\newlabel{tab:costs@cref}{{[table][2][1]S1.2}{[1][4][]4}{}{}{}}
\newlabel{sup:scores}{{S1}{5}{}{}{}}
\newlabel{sup:scores@cref}{{[section][1][]S1}{[1][5][]5}{}{}{}}
\newlabel{sup:diff}{{S2}{6}{}{}{}}
\newlabel{sup:diff@cref}{{[section][2][]S2}{[1][6][]6}{}{}{}}
\newlabel{tab:detectdiff}{{S2.1}{7}{}{}{}}
\newlabel{tab:detectdiff@cref}{{[table][1][2]S2.1}{[1][7][]7}{}{}{}}
\newlabel{tab:labelsvsdetector}{{S2.2}{8}{}{}{}}
\newlabel{tab:labelsvsdetector@cref}{{[table][2][2]S2.2}{[1][8][]8}{}{}{}}
\newlabel{tab:boundsupervised}{{S2.3}{8}{}{}{}}
\newlabel{tab:boundsupervised@cref}{{[table][3][2]S2.3}{[1][8][]8}{}{}{}}
\newlabel{tab:surface}{{S2.4}{9}{}{}{}}
\newlabel{tab:surface@cref}{{[table][4][2]S2.4}{[1][8][]9}{}{}{}}
\newlabel{tab:ablationdiff}{{S2.5}{9}{}{}{}}
\newlabel{tab:ablationdiff@cref}{{[table][5][2]S2.5}{[1][8][]9}{}{}{}}
\newlabel{tab:otherdiff}{{S2.6}{9}{}{}{}}
\newlabel{tab:otherdiff@cref}{{[table][6][2]S2.6}{[1][8][]9}{}{}{}}
\newlabel{tab:residadddiff}{{S2.7}{9}{}{}{}}
\newlabel{tab:residadddiff@cref}{{[table][7][2]S2.7}{[1][8][]9}{}{}{}}
\newlabel{tab:risk80}{{S2.8}{10}{}{}{}}
\newlabel{tab:risk80@cref}{{[table][8][2]S2.8}{[1][8][]10}{}{}{}}
\newlabel{tab:referral}{{S2.9}{10}{}{}{}}
\newlabel{tab:referral@cref}{{[table][9][2]S2.9}{[1][8][]10}{}{}{}}
\newlabel{tab:metrics2}{{S2.10}{11}{}{}{}}
\newlabel{tab:metrics2@cref}{{[table][10][2]S2.10}{[1][11][]11}{}{}{}}
\newlabel{sup:seeds}{{S2}{11}{}{}{}}
\newlabel{sup:seeds@cref}{{[section][2][]S2}{[1][11][]11}{}{}{}}
\newlabel{tab:views}{{S2.11}{12}{}{}{}}
\newlabel{tab:views@cref}{{[table][11][2]S2.11}{[1][11][]12}{}{}{}}
\newlabel{tab:seeds}{{S2.12}{13}{}{}{}}
\newlabel{tab:seeds@cref}{{[table][12][2]S2.12}{[1][12][]13}{}{}{}}
\newlabel{sup:ltt}{{S3}{13}{}{}{}}
\newlabel{sup:ltt@cref}{{[section][3][]S3}{[1][13][]13}{}{}{}}
\newlabel{sup:ltt:rule}{{S3}{13}{}{}{}}
\newlabel{sup:ltt:rule@cref}{{[section][3][]S3}{[1][13][]13}{}{}{}}
\newlabel{RF1}{}
\newlabel{tab:trainres}{{S2.13}{14}{}{}{}}
\newlabel{tab:trainres@cref}{{[table][13][2]S2.13}{[1][13][]14}{}{}{}}
\newlabel{sup:ltt:simulation}{{S3}{15}{}{}{}}
\newlabel{sup:ltt:simulation@cref}{{[section][3][]S3}{[1][15][]15}{}{}{}}
\newlabel{sup:weights}{{S4}{15}{}{}{}}
\newlabel{sup:weights@cref}{{[section][4][]S4}{[1][15][]15}{}{}{}}
\newlabel{sup:ablation}{{S4}{15}{}{}{}}
\newlabel{sup:ablation@cref}{{[section][4][]S4}{[1][15][]15}{}{}{}}
\newlabel{tab:ablation}{{S4.1}{16}{}{}{}}
\newlabel{tab:ablation@cref}{{[table][1][4]S4.1}{[1][16][]16}{}{}{}}
\newlabel{tab:overlapbinormal}{{S4.2}{17}{}{}{}}
\newlabel{tab:overlapbinormal@cref}{{[table][2][4]S4.2}{[1][17][]17}{}{}{}}
\newlabel{sup:tenk}{{S4}{18}{}{}{}}
\newlabel{sup:tenk@cref}{{[section][4][]S4}{[1][18][]18}{}{}{}}
\newlabel{tab:tenk}{{S4.3}{20}{}{}{}}
\newlabel{tab:tenk@cref}{{[table][3][4]S4.3}{[1][20][]20}{}{}{}}
\newlabel{tab:tenkadd}{{S4.4}{21}{}{}{}}
\newlabel{tab:tenkadd@cref}{{[table][4][4]S4.4}{[1][20][]21}{}{}{}}
\newlabel{tab:bicrtenk}{{S4.5}{22}{}{}{}}
\newlabel{tab:bicrtenk@cref}{{[table][5][4]S4.5}{[1][20][]22}{}{}{}}
\newlabel{tab:weights}{{S4.6}{22}{}{}{}}
\newlabel{tab:weights@cref}{{[table][6][4]S4.6}{[1][22][]22}{}{}{}}
\newlabel{tab:sens}{{S4.7}{23}{}{}{}}
\newlabel{tab:sens@cref}{{[table][7][4]S4.7}{[1][22][]23}{}{}{}}
\newlabel{tab:ltt}{{S4.8}{23}{}{}{}}
\newlabel{tab:ltt@cref}{{[table][8][4]S4.8}{[1][23][]23}{}{}{}}
\newlabel{fig:curve}{{S4.1}{24}{}{}{}}
\newlabel{fig:curve@cref}{{[figure][1][4]S4.1}{[1][24][]24}{}{}{}}
\newlabel{sup:natural}{{S5}{24}{}{}{}}
\newlabel{sup:natural@cref}{{[section][5][]S5}{[1][24][]24}{}{}{}}
\newlabel{tab:detect}{{S5.1}{25}{}{}{}}
\newlabel{tab:detect@cref}{{[table][1][5]S5.1}{[1][25][]25}{}{}{}}
\newlabel{sup:plugin}{{S6}{25}{}{}{}}
\newlabel{sup:plugin@cref}{{[section][6][]S6}{[1][25][]25}{}{}{}}
\newlabel{fig:pluginnewitems}{{S6.1}{26}{}{}{}}
\newlabel{fig:pluginnewitems@cref}{{[figure][1][6]S6.1}{[1][26][]26}{}{}{}}
\newlabel{tab:bytypefull}{{S6.1}{27}{}{}{}}
\newlabel{tab:bytypefull@cref}{{[table][1][6]S6.1}{[1][26][]27}{}{}{}}
\newlabel{fig:reviewcost}{{S6.2}{28}{}{}{}}
\newlabel{fig:reviewcost@cref}{{[figure][2][6]S6.2}{[1][26][]28}{}{}{}}
\newlabel{tab:bytypeho}{{S6.2}{29}{}{}{}}
\newlabel{tab:bytypeho@cref}{{[table][2][6]S6.2}{[1][26][]29}{}{}{}}
\newlabel{tab:gaindiff}{{S6.3}{29}{}{}{}}
\newlabel{tab:gaindiff@cref}{{[table][3][6]S6.3}{[1][26][]29}{}{}{}}
\newlabel{tab:bytypenewitems}{{S6.4}{30}{}{}{}}
\newlabel{tab:bytypenewitems@cref}{{[table][4][6]S6.4}{[1][26][]30}{}{}{}}
\newlabel{RF2}{}
\newlabel{tab:plugci}{{S6.5}{31}{}{}{}}
\newlabel{tab:plugci@cref}{{[table][5][6]S6.5}{[1][26][]31}{}{}{}}
\newlabel{tab:addsig}{{S6.6}{32}{}{}{}}
\newlabel{tab:addsig@cref}{{[table][6][6]S6.6}{[1][26][]32}{}{}{}}
\newlabel{tab:equalinfo}{{S6.7}{32}{}{}{}}
\newlabel{tab:equalinfo@cref}{{[table][7][6]S6.7}{[1][26][]32}{}{}{}}
\newlabel{tab:errtype}{{S6.8}{32}{}{}{}}
\newlabel{tab:errtype@cref}{{[table][8][6]S6.8}{[1][30][]32}{}{}{}}
\newlabel{sup:extra}{{S7}{33}{}{}{}}
\newlabel{sup:extra@cref}{{[section][7][]S7}{[1][33][]33}{}{}{}}
\newlabel{sup:morebase}{{S7}{33}{}{}{}}
\newlabel{sup:morebase@cref}{{[section][7][]S7}{[1][33][]33}{}{}{}}
\newlabel{tab:offideal}{{S7.1}{34}{}{}{}}
\newlabel{tab:offideal@cref}{{[table][1][7]S7.1}{[1][33][]34}{}{}{}}
\newlabel{tab:samplebase}{{S7.2}{35}{}{}{}}
\newlabel{tab:samplebase@cref}{{[table][2][7]S7.2}{[1][35][]35}{}{}{}}
\newlabel{sup:tokenreadout}{{S7}{35}{}{}{}}
\newlabel{sup:tokenreadout@cref}{{[section][7][]S7}{[1][35][]35}{}{}{}}
\newlabel{tab:extraadd}{{S7.3}{36}{}{}{}}
\newlabel{tab:extraadd@cref}{{[table][3][7]S7.3}{[1][35][]36}{}{}{}}
\newlabel{tab:pointlists}{{S7.4}{37}{}{}{}}
\newlabel{tab:pointlists@cref}{{[table][4][7]S7.4}{[1][35][]37}{}{}{}}
\newlabel{tab:teachers}{{S7.5}{37}{}{}{}}
\newlabel{tab:teachers@cref}{{[table][5][7]S7.5}{[1][35][]37}{}{}{}}
\newlabel{tab:offidealci}{{S7.6}{37}{}{}{}}
\newlabel{tab:offidealci@cref}{{[table][6][7]S7.6}{[1][35][]37}{}{}{}}
\newlabel{sup:attnref}{{S7}{38}{}{}{}}
\newlabel{sup:attnref@cref}{{[section][7][]S7}{[1][38][]38}{}{}{}}
\newlabel{sup:clean}{{S7}{38}{}{}{}}
\newlabel{sup:clean@cref}{{[section][7][]S7}{[1][38][]38}{}{}{}}
\newlabel{RF3}{}
\newlabel{tab:tokenreadout}{{S7.7}{39}{}{}{}}
\newlabel{tab:tokenreadout@cref}{{[table][7][7]S7.7}{[1][38][]39}{}{}{}}
\newlabel{tab:stable}{{S7.8}{40}{}{}{}}
\newlabel{tab:stable@cref}{{[table][8][7]S7.8}{[1][40][]40}{}{}{}}
\newlabel{sup:pull}{{S7}{40}{}{}{}}
\newlabel{sup:pull@cref}{{[section][7][]S7}{[1][40][]40}{}{}{}}
\newlabel{tab:readctl}{{S7.9}{41}{}{}{}}
\newlabel{tab:readctl@cref}{{[table][9][7]S7.9}{[1][41][]41}{}{}{}}
\newlabel{tab:readctlho}{{S7.10}{42}{}{}{}}
\newlabel{tab:readctlho@cref}{{[table][10][7]S7.10}{[1][41][]42}{}{}{}}
\newlabel{tab:readctlnewitems}{{S7.11}{42}{}{}{}}
\newlabel{tab:readctlnewitems@cref}{{[table][11][7]S7.11}{[1][41][]42}{}{}{}}
\newlabel{fig:residual}{{S7.1}{42}{}{}{}}
\newlabel{fig:residual@cref}{{[figure][1][7]S7.1}{[1][41][]42}{}{}{}}
\newlabel{RF4}{}
\newlabel{tab:secondans}{{S7.12}{44}{}{}{}}
\newlabel{tab:secondans@cref}{{[table][12][7]S7.12}{[1][43][]44}{}{}{}}
\newlabel{tab:sameclass}{{S7.13}{45}{}{}{}}
\newlabel{tab:sameclass@cref}{{[table][13][7]S7.13}{[1][43][]45}{}{}{}}
\newlabel{sup:domains}{{S7}{45}{}{}{}}
\newlabel{sup:domains@cref}{{[section][7][]S7}{[1][45][]45}{}{}{}}
\newlabel{tab:o365}{{S7.14}{46}{}{}{}}
\newlabel{tab:o365@cref}{{[table][14][7]S7.14}{[1][46][]46}{}{}{}}
\newlabel{sup:second}{{S7}{47}{}{}{}}
\newlabel{sup:second@cref}{{[section][7][]S7}{[1][47][]47}{}{}{}}
\newlabel{tab:external}{{S7.15}{48}{}{}{}}
\newlabel{tab:external@cref}{{[table][15][7]S7.15}{[1][47][]48}{}{}{}}
\newlabel{sup:ctl}{{S8}{48}{}{}{}}
\newlabel{sup:ctl@cref}{{[section][8][]S8}{[1][47][]48}{}{}{}}
\newlabel{sec:ctl:estimand}{{S8.1}{48}{}{}{}}
\newlabel{sec:ctl:estimand@cref}{{[subsection][1][8]S8.1}{[1][48][]48}{}{}{}}
\newlabel{eq:twocomp}{{S8.1}{49}{}{}{}}
\newlabel{eq:twocomp@cref}{{[equation][1][8]S8.1}{[1][48][]49}{}{}{}}
\newlabel{sec:ctl:models}{{S8.2}{49}{}{}{}}
\newlabel{sec:ctl:models@cref}{{[subsection][2][8]S8.2}{[1][49][]49}{}{}{}}
\newlabel{sec:ctl:difficulty}{{S8.3}{49}{}{}{}}
\newlabel{sec:ctl:difficulty@cref}{{[subsection][3][8]S8.3}{[1][49][]49}{}{}{}}
\newlabel{tab:ownother}{{S8.1}{50}{}{}{}}
\newlabel{tab:ownother@cref}{{[table][1][8]S8.1}{[1][50][]50}{}{}{}}
\newlabel{tab:ownothermodel}{{S8.2}{50}{}{}{}}
\newlabel{tab:ownothermodel@cref}{{[table][2][8]S8.2}{[1][50][]50}{}{}{}}
\newlabel{fig:controlled}{{S8.1}{51}{}{}{}}
\newlabel{fig:controlled@cref}{{[figure][1][8]S8.1}{[1][51][]51}{}{}{}}
\newlabel{tab:controlled}{{S8.3}{52}{}{}{}}
\newlabel{tab:controlled@cref}{{[table][3][8]S8.3}{[1][51][]52}{}{}{}}
\newlabel{tab:pairsrob}{{S8.4}{52}{}{}{}}
\newlabel{tab:pairsrob@cref}{{[table][4][8]S8.4}{[1][51][]52}{}{}{}}
\newlabel{tab:dprobe}{{S8.5}{53}{}{}{}}
\newlabel{tab:dprobe@cref}{{[table][5][8]S8.5}{[1][51][]53}{}{}{}}
\newlabel{sup:heldout}{{S9}{53}{}{}{}}
\newlabel{sup:heldout@cref}{{[section][9][]S9}{[1][51][]53}{}{}{}}
\newlabel{tab:spredrepfull}{{S9.1}{54}{}{}{}}
\newlabel{tab:spredrepfull@cref}{{[table][1][9]S9.1}{[1][54][]54}{}{}{}}
\newlabel{sup:replication}{{S10}{54}{}{}{}}
\newlabel{sup:replication@cref}{{[section][10][]S10}{[1][54][]54}{}{}{}}
\newlabel{tab:detectho}{{S9.2}{55}{}{}{}}
\newlabel{tab:detectho@cref}{{[table][2][9]S9.2}{[1][54][]55}{}{}{}}
\newlabel{RF5}{}
\newlabel{tab:heldout}{{S9.3}{56}{}{}{}}
\newlabel{tab:heldout@cref}{{[table][3][9]S9.3}{[1][54][]56}{}{}{}}
\newlabel{RF6}{}
\newlabel{tab:residheld}{{S9.4}{57}{}{}{}}
\newlabel{tab:residheld@cref}{{[table][4][9]S9.4}{[1][54][]57}{}{}{}}
\newlabel{tab:heldoutpairs}{{S9.5}{58}{}{}{}}
\newlabel{tab:heldoutpairs@cref}{{[table][5][9]S9.5}{[1][54][]58}{}{}{}}
\newlabel{tab:replication}{{S10.1}{59}{}{}{}}
\newlabel{tab:replication@cref}{{[table][1][10]S10.1}{[1][58][]59}{}{}{}}

%% file: sections/abstract.tex
Grounded vision-language models write boxes as text, and the mean probability of their coordinate tokens says little about whether a box is right. We study a single-pass signal for boxes on another object, which matter when only the object counts. A linear map, fitted per model to its own answers without labels, reads a box from the hidden state before writing. Its disagreement with the written box, the self-prediction residual, flags boxes on another object although the read-out is far less accurate; the residual's weight was fixed with labels on two models. On a replication list of 1{,}200 RefCOCOg test items, the residual, averaged over six models, raises the detection of these errors for each of fifteen single-pass scores, by 0.026 to 0.049 in area under the receiver operating characteristic curve (AUROC) on a label-free probe and probes trained on 2,000 labelled items. Over all errors, adding it to the label-free probe moves AUROC, averaged over the models, by less than 0.02 (95\% interval). With a fifth of answers reviewed, the share of all errors caught moves from 0.518 to 0.524, that of boxes on another object from 0.505 to 0.607 (post hoc): the residual changes which errors are found, not how many, and in cost-weighted recall it breaks even when such a box costs about 0.86 times another error, and its gain is positive in point estimate at equal costs. These gains come from RefCOCOg lists of MS-COCO images. %

%% file: sections/keywords.tex
visual grounding \sep error detection \sep vision-language models \sep hidden-state probe \sep self-prediction residual \sep selective prediction

%% file: sections/introduction.tex
\section{Introduction}
\label{sec:intro}

Grounded vision-language models (VLMs) answer a referring expression with a bounding box written as text~\cite{bai2025qwen25vl,ax2511.21631}. Some uses of such a box need it only to pick out the object: a grasp or a crop handed to the next step may tolerate a loose box but not a box around another object. Evaluations of this kind count a prediction as right when it falls in the annotated region, as in the pointing game~\cite{zhang2018excitation}. For such uses we treat a box on another object as the failure to catch. Deciding which outputs to trust is the problem of classification with a reject option~\cite{fumera2000reject} and of failure prediction~\cite{hendrycks2017msp,jiang2018trustscore}. Averaged over the coordinate tokens, the model's own probability says little about whether a box is right~\cite{ax2606.25760,ax2607.05978}: it separates wrong from right boxes with an area under the receiver operating characteristic curve (AUROC) of 0.612 to 0.665 on the models we test, on the replication list of \cref{sec:setup}; weighting the digits by place value helps~\cite{li2026pace}, and a digit-weighted coordinate uncertainty is among our baselines (\cref{tab:detect}). Querying the model again, by resampling~\cite{wang2023selfconsistency,kuhn2023semantic}, with perturbed views~\cite{zhang2026mvp} or with another model~\cite{ax2609.37488}, at least doubles the cost (\cref{tab:detect,tab:views}). Checks of whether the answer changes are also weak on errors that the model repeats, as reported for text models~\cite{ax2505.17656,orgad2025know}: on the four Qwen models, 0.208 to 0.476 of the errors on the new items keep their box (intersection over union, IoU, at least 0.5) under all seven perturbed views (\cref{tab:stable}). Probes on hidden states are cheap at test time but are usually trained on correctness labels~\cite{azaria2023saplma,ax2605.10893}, which need gold boxes for the training items.

We study a signal that needs neither labels nor a second pass. A linear map, the \emph{read-out}, is fitted once per model to the model's own greedy answers and reads a box from the hidden state before the coordinates are written; the disagreement between this box and the written one, $1-\mathrm{IoU}$, is the \emph{self-prediction residual} (\cref{sec:method}). It targets \emph{wrong-object errors}, boxes on another object (IoU with the gold box below 0.1); 72.6\% of those on the replication list cover another annotated object, so most are boxes around real objects (\cref{sec:measure:pull}). Added to a probe trained without labels, the residual forms a single-pass \emph{detector}, which we call SPRED, for self-prediction residual error detector (\cref{sec:method:probe}). With a fifth of the answers sent to review, SPRED catches 0.607 of the wrong-object errors on the replication list, against 0.505 for its probe alone, 0.309 for coordinate log-prob and 0.200 at random, while the share of all errors caught moves only from 0.518 to 0.524 (means over the models, post hoc). The residual thus changes which errors are found rather than how many; in cost-weighted recall it breaks even when a wrong-object error costs about 0.86 times another error, and its gain is positive in point estimate at equal costs (replication list, post hoc; \cref{sec:discussion}).

The read-out acts as a weak second localizer. What sets it apart from the nearest ways of getting a second opinion is where that opinion comes from. Another localizer or another model needs a further model and a further pass at test time. Detectors that flag a disagreement between a probe and the output need gold answers and compare discrete classes~\cite{sun2025arith,elshangiti2026count,ax2609.11216}, box heads that read the state before the answer are trained on gold boxes~\cite{ax2610.13239,ax2610.35053}, and predictors fitted to a model's own outputs~\cite{ax2502.13329,belrose2023tunedlens} are not compared with the answer the model writes. Like training-free localizers inside the model~\cite{kang2025heads,phukan2025contextuallens}, the read-out needs no gold boxes and no other model and is computed in the same pass; it differs from them in being a map fitted to the model's own boxes and in its use, a continuous disagreement $1-\mathrm{IoU}$ with the written box (\cref{sec:method:design}). \Cref{tab:position} sets it apart from nearby signals.

\begin{table}[!t]
\centering
\caption{The residual and the nearest kinds of error signal for grounded VLMs. Gold: gold boxes needed for training. Passes: forward passes of the grounding model at test time (LOCUS: with its verification). Fitted to: target of the trained part. Compared with: what the written answer is compared with to score it. The first three columns give what a check needs, the last two what it computes. Score: continuous error score, a discrete flag, a choice among boxes, or a prediction about the answer or its tokens. Re-asking: resampling, perturbed views or other models; the cited methods use the agreement to improve accuracy, and it is evaluated as an error score here.}
\label{tab:position}
\setlength{\tabcolsep}{3pt}
\begin{tabular}{@{}>{\raggedright\arraybackslash}p{3.1cm}cc>{\raggedright\arraybackslash}p{1.6cm}>{\raggedright\arraybackslash}p{1.6cm}>{\raggedright\arraybackslash}p{1.7cm}@{}}
\toprule
Signal & Gold & Passes & Fitted to & Compared with & Score\\
\midrule
Coordinate log-prob~\cite{ax2606.25760} (cf.~\cite{li2026pace}) & no & 1 & --- & --- & continuous\\
Re-asking the model or others~\cite{wang2023selfconsistency,zhang2026mvp,ax2609.37488} & no & $\ge 2$ & --- & other answers & continuous or choice\\
Correctness probes~\cite{azaria2023saplma,ax2605.10893} & yes & 1 & correctness & --- & continuous\\
Sampled-answer probes~\cite{kossen2024sep,ax2604.19444} & no & 1 & sampled answers & --- & continuous\\
Probes of the model's own output~\cite{ax2502.13329} & no & 1 & own outputs & --- & answer property\\
Layer-wise predictor agreement~\cite{chen2019whitebox} & yes & 1 & labels & layer predictions & continuous\\
Probe--output disagreement~\cite{sun2025arith,elshangiti2026count,ax2609.11216} & yes & 1 & correct answers & output or its probe & discrete\\
Training-free localizers in the model~\cite{kang2025heads,phukan2025contextuallens} & no & 1 & not fitted & --- & box or map\\
Box heads (Jag, LOCUS)~\cite{ax2610.13239,ax2610.35053} & yes & 1 / ${\approx}5.6$ & gold boxes & --- / head's box, to choose a box & choice\\
Logit lens on a layer~\cite{belrose2023tunedlens} & no & 1 & not fitted & --- & token prediction\\
Tuned and future lenses~\cite{belrose2023tunedlens,pal2023futurelens} & no & 1 & own outputs & --- & token prediction\\
Self-prediction residual (ours) & no & 1 & own answers & read-out & continuous\\
\bottomrule
\end{tabular}
\end{table}

Our contributions are:
\begin{itemize}
\item \emph{A label-free signal and what it marks.} The self-prediction residual compares the written box with a box read from the state before the answer by a linear map fitted to the model's own answers; it needs no labels to fit, leaves the model frozen and runs in the generation pass (white-box access to hidden states is required). Among wrong answers only, it separates boxes on another object from the other errors with an AUROC of 0.726 on the replication list, averaged over the six models, and it raises a classifier on the geometry of the written box by 0.061 (\cref{sec:plug:add}). On wrong-object errors the read-out lies partway towards the annotated object. The signal is largely about the item (post hoc), yet on replication-list items that one of five models answers correctly and another with a box on another object, the residual picks the model that erred in 0.608 of the cases (post hoc): it is a second reading from the checked model's pass that needs no gold boxes, no other model and no further pass (\cref{sec:measure,sec:discussion}).

\item \emph{SPRED and a plug-in rule.} With a fixed weight the residual is added to a label-free two-teacher probe, which gives SPRED, a single-pass detector that needs no labels beyond that weight (\cref{sec:method:probe}). The same rule adds the residual to other single-pass scores, and on wrong-object errors it adds more than the other label-free signals tested in its place (\cref{sec:plug:add}).

\item \emph{Gains on RefCOCOg lists, on six models.} On a RefCOCOg replication list of 1{,}200 items (\cref{sec:setup}), the residual raises the detection of wrong-object errors, averaged over the models, for each of fifteen single-pass scores, which, like this measure, were fixed before the list was scored; the held-out list, also from RefCOCOg and scored before they were fixed, shows the same (post hoc); on the two-teacher probe and the six probes trained on 2,000 labelled items the residual adds 0.026 to 0.049. With the top-scoring fifth sent to review, the share of these errors caught rises from 0.469 to 0.575, averaged over the model--score cells (\cref{sec:plug:add}). When three of the models are prompted for a point instead, so that only the object counts, the residual, here a point-to-point distance, raises the two-teacher probe's detection of all errors by 0.035 on average over the replication and held-out lists (post hoc), most on \qsmall{} (0.070, against 0.015 on \qlarge{} and 0.020 on \qfour{}; \cref{sec:plug:add}).
\end{itemize}

\emph{Scope.} The residual addresses wrong-object errors, 27.8\% to 48.3\% of all errors per model on the replication list, and the gains above were measured on RefCOCOg lists of MS-COCO images.

%% file: sections/related.tex
\section{Related work}
\label{sec:related}

\paragraph{Visual grounding and its reliability}
Referring expression comprehension (REC) has moved from graph-based and language-guided detectors~\cite{ke2025graphrec,cheng2026plrvg,wang2027plvl} to multimodal language models that write boxes as text~\cite{bai2025qwen25vl,ax2511.21631}. Generalized REC lets an expression refer to several objects or to none~\cite{he2023grec}; our lists keep one target per expression, so a box on another object is defined geometrically, by IoU $<0.1$ with the annotated box. Their errors are studied mostly as something to reduce: by enhancing visual tokens against hallucination~\cite{huang2027tave}, by aggregating multiple views when coordinates are unstable~\cite{zhang2026mvp}, or by selecting among the boxes of several models~\cite{ax2609.37488}. Fewer works score an individual answer: a benchmark of 27 post-hoc uncertainty scores for grounding in graphical user interfaces, hidden-state and density scores included~\cite{ax2606.25760}, the attention of coordinate tokens~\cite{ax2607.05978}, the spread of sampled coordinates, calibrated into a risk-controlled decision~\cite{wang2026safeground}, which needs several samples per item, and the attention maps of the pass that wrote a box, scored by an IoU regressor trained on gold boxes or by head entropy, to choose among sampled boxes~\cite{yang2026acs}.

\paragraph{Error detection from internal states}
Comparing a prediction with what intermediate layers predict is an established idea in classification: meta-models on linear probes of intermediate layers score confidence~\cite{chen2019whitebox}; they are trained with labels and their output is a class. Hidden states of language models predict whether an output is correct and can encode more than the output shows~\cite{orgad2025know}, and such disagreements between output and probe have several sources~\cite{liu2023dissonance}. Disagreement has also been used to flag single answers: for arithmetic~\cite{sun2025arith} and for counting in VLMs~\cite{elshangiti2026count}, when a probe fitted to the model's own outputs and a probe fitted to the correct answers disagree, and in a synthetic binding task by comparing a probe fitted to the correct bindings with the output token~\cite{ax2609.11216}. Each needs gold answers for at least one probe and compares discrete classes; the residual needs none, and its gap to the written box is a continuous $1-\mathrm{IoU}$. Supervised probes such as SAPLMA~\cite{azaria2023saplma} learn from correctness labels; probes on the query state predict properties of the model's own output before generation~\cite{ax2502.13329}, for VLMs supervised probes on pre-generation states predict hallucination~\cite{ax2603.05465}, and a trajectory of token probabilities and visual features across depth estimates the correctness of a text answer in one pass, with separators calibrated on 500 labelled examples per task~\cite{fime2026trajectory}. Without correctness labels, the Mahalanobis distance flags unusual inputs~\cite{lee2018mahalanobis}, and two further label-free scores of a single pass are compared in \cref{sup:morebase}; semantic-entropy probes replace labels with a teacher computed from sampled answers~\cite{kossen2024sep}, as does one-sample agreement~\cite{ax2604.19444}, and HaloScope learns a truthfulness classifier from unlabelled generations through a subspace of their embeddings~\cite{du2024haloscope}, scoring whole text answers.

\paragraph{Box heads and lenses}
Frozen VLMs also contain localizers that need no training: a few attention heads whose text-to-image maps give a box~\cite{kang2025heads}, the output head applied to image representations~\cite{jiang2025vlinterp}, and similarity maps between middle-layer embeddings of the answer tokens and the image patches~\cite{phukan2025contextuallens}. These produce an answer or score a text answer; they are not compared with a box that the model writes. Jag adapts the backbone together with a regression head that reads the state of the last input token~\cite{ax2610.13239}; LOCUS trains such a head jointly with the backbone on gold boxes, adds an IoU-aware confidence and uses the head's box as a draft that the coordinate tokens verify, in about 5.6 passes on average~\cite{ax2610.35053}. Both are trained on gold boxes, and LOCUS uses the agreement to choose a box. OLA decodes ordinal labels from the prompt-end state with label-supervised lenses~\cite{li2026ola}. In VLMs, ranking hidden dimensions by their contribution to coordinate tokens locates where coordinates are emitted~\cite{ax2609.32757}. %

The tuned lens trains, for each layer, an affine map followed by the model's own output head so that the layer reproduces the model's final next-token distribution at the same position; it needs no labels~\cite{belrose2023tunedlens}. At the last prompt token it therefore predicts the first token of the answer, not a box. The lens that corresponds to our setting predicts later tokens from one earlier state, as the future lens does~\cite{pal2023futurelens}; the read-out is such a predictor for a box, fitted by ridge regression to the box the model writes. \Cref{sec:method:design} compares it with a box decoded by the output head through the logit lens~\cite{belrose2023tunedlens}. A token-style read-out of the future-lens kind, fitted to the same targets, is no better on wrong-object errors (\cref{sup:tokenreadout}).

\paragraph{Consistency checks and box confidence}
Agreement among sampled answers~\cite{wang2023selfconsistency,kuhn2023semantic} is a standard uncertainty signal, and VLM answers are also checked by masking the most attended image regions in a second pass~\cite{park2026vauq}. BICR trains a correctness probe on the state of the last prompt token with a ranking term that asks for higher confidence with the real image than with a blanked one, and runs one pass at test time~\cite{ax2605.10893}; it needs correctness labels and, for training only, one extra pass per item with the blanked image. We train it on the same 2,000 labelled items as our other supervised probes (the original used about 20,000; \cref{sup:scores,sec:plug:detector}). Detectors also predict the IoU of their boxes~\cite{jiang2018iounet}. Without labels, the overlap between the proposal and the final box of a two-stage detector, and the stability of a detector's boxes under input noise, have been combined with classification uncertainty to choose images to annotate~\cite{kao2018localization}; the residual and the two-teacher probe are their counterparts for a model that writes its box as text. Each check costs at least one further pass at test time, and errors that a model repeats partly escape it: for text models, such self-consistent errors do not decrease with model size and also weaken supervised probes on the model's own states~\cite{ax2505.17656}.

\paragraph{Selective prediction and risk control}
Classifiers with a reject option abstain on the least certain inputs~\cite{fumera2000reject}, and failure prediction ranks errors by the maximum softmax probability~\cite{hendrycks2017msp}, by agreement with a nearest-neighbour classifier~\cite{jiang2018trustscore} or by the uncertainty of a pre-trained backbone~\cite{zhou2027edl}. Learn-then-Test turns a score into a rule with a risk guarantee~\cite{angelopoulos2021ltt}. The residual supplies the ranking; a guaranteed threshold still needs labelled calibration items (\cref{sec:method:use}).

%% file: sections/method.tex
\section{The self-prediction residual}
\label{sec:method}

\subsection{Problem and cost}
\label{sec:method:setup}

\begin{figure}[!t]
\centering
\includegraphics[width=\textwidth]{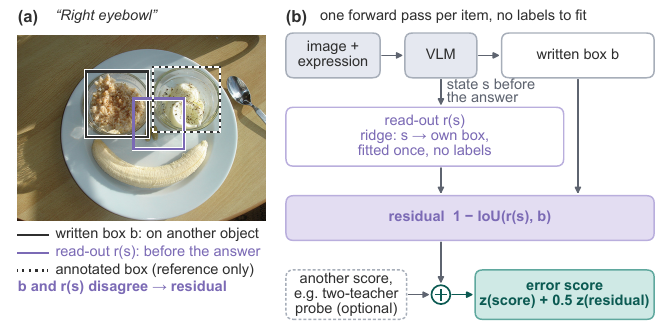}
\caption{The self-prediction residual. (a) A wrong-object error of \qlarge{} on a new item (``Right eyebowl'' [sic]): written box $b$ (black), box $r(s)$ read out from the state before the answer (lavender), annotated box (white dashes, for reference only). (b) One forward pass: a ridge map fitted to the model's own boxes gives $r(s)$; the residual $1-\mathrm{IoU}(r(s), b)$ enters an error score as $0.5\,z(\text{residual})$. Photograph: ``Start the day with a smile'' by Kim Anderson, CC BY 2.0 (MS-COCO).}
\label{fig:method}
\end{figure}

A grounded VLM $f$ receives an image $I$ and a referring expression $q$ and writes a box $\hat b = f(I, q)$. The answer is an error when $\mathrm{IoU}(\hat b, b^\ast) < 0.5$ for the gold box $b^\ast$. An error score $s(I, q)$ should be higher for errors; it is evaluated by the area under the receiver operating characteristic curve (AUROC) and by the error rate among the answers it keeps (selective prediction, \cref{sup:ltt}). The gold box is used only for evaluation. Cost is counted in forward passes of any model at test time: generating $\hat b$ is one pass, and every check that queries the grounding model again, or runs another model such as a separate detector, adds at least one. The residual and the detector below read the hidden states of the generation pass through linear maps, so the test stays at one pass (\cref{fig:method}).

For each item we keep two hidden states of the language model at every fourth layer and the last: $h^{\mathrm{last}}_\ell$, the state of the last prompt token before the first coordinate token is written, and $h^{\mathrm{coord}}_\ell$, the mean state over the coordinate tokens of $\hat b$. Both are standardized with the statistics of the training items, which are unlabelled referring expressions of the same source.

\subsection{Read-out and residual}
\label{sec:method:residual}

The residual asks whether the state before writing agrees with the box that is written. A ridge map $r$, the \emph{read-out}, maps $h^{\mathrm{last}}_\ell$ to a box in coordinates normalized by the image size,
\begin{equation}
r = \arg\min_{W,\, c} \sum_{j} \bigl\lVert W h^{\mathrm{last}}_{\ell,j} + c - \hat b_j\bigr\rVert^2 + \lambda \lVert W\rVert^2,
\label{eq:readout}
\end{equation}
fitted once per model on its training items with the model's own greedy boxes $\hat b_j$ as targets, so no gold box enters; the layer $\ell$ and $\lambda \in \{10^2, 10^3, 10^4, 10^5\}$ are chosen by five-fold cross-validation for the mean IoU between read-out and own box. The \emph{self-prediction residual} is
\begin{equation}
u = 1 - \mathrm{IoU}\bigl(r(h^{\mathrm{last}}_\ell), \hat b\bigr),
\label{eq:residual}
\end{equation}
set to 1 when $\hat b$ cannot be parsed. The name describes the procedure: a model's hidden state predicts the box the same model writes. The read-out is thus a second reading of the object from the same pass, a weak localizer whose disagreement with the written box flags errors; much of the signal belongs to the item (\cref{sec:measure:own}). On items like the training items the read-out and the written box agree and $u$ is small. A large $u$ marks a written box that departs from the state before writing.

\subsection{Design choices}
\label{sec:method:design}

Three choices define the read-out. \emph{It is fitted.} A box decoded from an intermediate layer by the model's own output head through the logit lens, the unfitted baseline of~\cite{belrose2023tunedlens}, would need no fitting, but at the layer fixed in advance most items yield no parseable coordinates, and at the layer best on the test labels, which favours it, its disagreement with the written box detects wrong-object errors worse than the residual on every model, by 0.135 to 0.242 on the replication list, every interval below zero. Among other training-free localizers, on the new items of \qsmall{} and \qlarge{} the residual adds more on wrong-object errors to the two-teacher probe than a reference for attention-head localizers~\cite{kang2025heads}, approximated from stored head centroids, with heads partly selected post hoc (\cref{sup:attnref}). \emph{The map is linear and coarse.} A read-out far less accurate than the written box can still flag that box's errors on targets wide compared with its error, because only their disagreement is used. The read-out is coarse (accuracy at IoU $\ge0.5$ of 0.391 to 0.467, against 0.853 to 0.876 for the written box, three models on the replication list). \emph{It is fitted to the model's own answers.} No gold box enters, so the map can be fitted on unlabelled items of a new model or domain. Fitting it to gold boxes instead, as label-supervised lenses do~\cite{li2026ola}, brings no consistent gain on the four models tested: it changes the detector's AUROC by $-$0.009 to 0.004, also on \qllava{}, whose boxes are right on only 0.619 of replication-list items. %

\subsection{Adding the residual to other scores}
\label{sec:method:combine}

To add the residual to a score $g$, each is standardized with the mean and standard deviation of its out-of-fold values on the training items, and the residual enters with half weight,
\begin{equation}
s = z(g) + 0.5\,z(u).
\label{eq:combine}
\end{equation}
The standardization uses no labels, and each test item is scored on its own. The weight 0.5 was chosen once, with labels and for the detection of all errors, while examining the six sets of \qsmall{} (which include its new items) and the new items of \qfour{}, and is kept unchanged for every other model, score and set; fitting the read-out, the probe and the standardization for a new model or domain needs no labels. The AUROC is flat around the weight: on the new items of five models, the AUROC at 0.5 is within 0.0010 of that at each model's binormal-optimal weight (\cref{sup:ablation,sup:weights}). Wrong-object errors became the main measure of the plug-in comparisons, and the fifteen scores were chosen after the held-out list was scored for SPRED and before the replication list was scored. %

\subsection{A label-free base and SPRED}
\label{sec:method:probe}

The residual needs a base score that ranks errors of every type. As a base that needs no labels either, we distil consistency checks, which query the model again~\cite{wang2023selfconsistency,kuhn2023semantic}, into the generation pass. For an unlabelled training item, a view $v$ gives a second box $\hat b_v$ mapped back into the original frame, and $e_v = 1 - \mathrm{IoU}(\hat b, \hat b_v)$, set to 1 when $\hat b_v$ cannot be parsed. Two views need no gold box: a content shift (\cref{alg:detector,sup:prompts}) and one sampled answer (temperature 0.7, top-$p$ 0.9). A ridge probe $g$ maps $h^{\mathrm{last}}_\ell$ or $h^{\mathrm{coord}}_\ell$ to their mean $t = (e_{\mathrm{shift}} + e_{\mathrm{samp}})/2$, with feature, layer and penalty chosen by five-fold cross-validation using whether $t$ is among the top 15\% of the training items as selection label. We call $g$ the \emph{two-teacher probe}, after its two views; it needs no labels. Training a probe on a teacher computed from extra passes follows semantic-entropy probes~\cite{kossen2024sep} and one-sample agreement targets~\cite{ax2604.19444}; the teacher here is box-valued, and teacher variants are compared in \cref{tab:teachers}. The \emph{detector} is \cref{eq:combine} with this probe as $g$; we call it SPRED (self-prediction residual error detector). It uses no label beyond the fixed weight, and its test-time cost is one pass; \cref{alg:detector} lists its training and use.

\begin{algorithm}[!t]
\caption{Training and use of SPRED. No gold box enters any step.}
\label{alg:detector}
\footnotesize
\begin{algorithmic}[1]
\Require grounding model $f$; unlabelled training items $(I_j, q_j)$, $j=1,\dots,n$
\For{each training item $j$}
  \State generate $\hat b_j = f(I_j, q_j)$ and keep the hidden states $h_j$
  \State $\hat b_{j,\mathrm{shift}} \gets$ box for the image padded on the left by a strip in its mean colour (\cref{sup:prompts}), shifted back
  \State $\hat b_{j,\mathrm{samp}} \gets$ box of one sampled answer
  \State $t_j \gets (e_{\mathrm{shift}} + e_{\mathrm{samp}})/2$, with $e_v = 1 - \mathrm{IoU}(\hat b_j, \hat b_{j,v})$
\EndFor
\State fit the probe $g$ to $(h_j, t_j)$; choose features, layer and $\lambda$ by five-fold cross-validation
\State fit the read-out map $r$ to $(h^{\mathrm{last}}_{\ell,j}, \hat b_j)$ (\cref{eq:readout}); choose layer and $\lambda$ by five-fold cross-validation
\State store the mean and standard deviation of the out-of-fold $g$ and $u$ (\cref{eq:residual})
\Statex \textit{Test item $(I, q)$, one pass:}
\State generate $\hat b = f(I, q)$ and keep $h$; $u \gets 1 - \mathrm{IoU}\bigl(r(h^{\mathrm{last}}_\ell), \hat b\bigr)$
\State \Return $s = z(g(h)) + 0.5\,z(u)$; abstain when $s$ exceeds the threshold of \cref{sup:ltt:rule} %
\end{algorithmic}
\end{algorithm}

\subsection{Before use: target size and thresholds}
\label{sec:method:use}

The read-out locates an object only to about a tenth of the image side, so the residual can separate a box on another object only when the target is wide compared with this read-out error $\sigma$, the median distance between the centres of the read-out and the own box on the training items, which needs no labels and is larger for read-outs fitted on Objects365 and remote-sensing images than on RefCOCO. The gain of the residual falls as the target gets narrower in units of $\sigma$ (\cref{sec:measure:not}). We report this as a description, not as a label-free switch for single items. Since this dependence was found post hoc, we offer a check that needs no labels, for new domains to confirm: estimate $\sigma$ on unlabelled items of the domain, in coordinates normalized by the image size, and count the written boxes whose width, taken as the square root of their area in the same units, is below $2\sigma$; if more than half are, the residual is unlikely to help, a threshold set post hoc between the shares observed. Such boxes are at most 0.015 of the items of any model on the RefCOCO family, but 0.599 to 0.665 on Objects365. The dependence was seen with the read-out fitted on Objects365. With the maps fitted on RefCOCO and applied unchanged, the residual adds little at any width, 0.013 below $2\sigma$ and 0.005, not detected, above (seven probes, three models; \cref{tab:o365}), so the read-out should be fitted on the domain. Taken together, the residual is worth adding to the probe when a box on another object costs more than other errors, its cost-weighted recall breaking even at a cost ratio of about 0.86 (replication list, post hoc; \cref{sec:discussion}), or when only the object counts (\cref{sec:plug:add}), and fewer than half of the written boxes are narrower than $2\sigma$. The residual needs white-box access to the hidden states. Computed in the generation pass, SPRED takes 0.997, 1.001 and 1.126 times the time of generation alone on \qsmall, \qlarge{} and \qfour{}, with almost no extra memory (\cref{sup:repro}); \cref{tab:position,tab:costs} set its requirements beside those of other checks. Without labels the residual and SPRED only rank answers; a threshold whose risk is controlled with high probability~\cite{angelopoulos2021ltt} needs labelled calibration items from the deployment domain, and, calibrated on RefCOCO with an 8\% target, its guarantee fails on Objects365 for the two Qwen2.5-VL models (\cref{sup:ltt}). %

%% file: sections/setup.tex
\section{Experimental setup}
\label{sec:setup}

\paragraph{Data} \Cref{tab:lists} gives the training and test lists. For each model the read-out and the probe are fitted on the unlabelled training list,\footnote{2{,}000 rows covering 1{,}963 distinct items (37 listed twice); fits use all rows.} of which at least 1{,}997 rows have a usable answer for each model; test images are disjoint from the training images. Headline results are on the replication list, a second RefCOCOg test sample on images used in no other list: the method, its hyperparameters and the thresholds were fixed before any score on this list was computed; analyses added afterwards are marked post hoc. \Cref{tab:replication} lists the comparisons on this list with their results, and values in the text are from this list unless another is named. The held-out list was first evaluated after the method, its weight and every reference score had been fixed (\cref{sup:heldout}). The new items were drawn at random with a fixed seed from images not used before the method was designed; unlike the held-out list, they were evaluated repeatedly, and those of \qsmall{} and \qfour{} were examined while the method was designed. Ref-L4 re-checked the annotation of part of the new and RefCOCO+ items; we use its flags to rescore these items on confirmed annotations only (\cref{sup:pull}). %

\input{tables/lists}

\paragraph{Models, errors and intervals} The models are \qsmall, \qlarge, \qfour, \qeight, \qivl{}, whose language model is InternLM2.5, and \qllava{} from a third family, its image padded to a square and nothing tuned on it. All receive the official grounding prompt and are decoded greedily in bf16. Each model has its own read-out and probes, fitted on its own training items; the supervised probes use the same cross-validation grid with correctness labels. An error is a box with IoU $<0.5$, scored against all other answers; for error types each type is scored against the correct answers (against all other answers: \cref{sup:morebase}): \emph{wrong object} (IoU $<0.1$), \emph{partial overlap} (0.1--0.3) and \emph{wrong extent} (0.3--0.5). Intervals are 95\% intervals of the paired difference from a bootstrap over images, in which all items of a sampled image enter together (10,000 resamples unless stated; \cref{sup:diff}); when that interval includes zero we say that no difference is detected, which does not show that the two scores are equal, and say that one score is \emph{ahead} only when it excludes zero. Against the supervised linear probe we also report the one-sided 95\% lower bound of the difference; the detector is within 0.03 when that bound exceeds $-0.03$, a margin set during development, before the held-out list was scored, and close to the half-width of a single model's interval for this difference on the held-out list (\cref{tab:heldout}). Intervals treat the six models as fixed and do not generalize over models. Intervals that cover refits (fold seeds, redrawn training items or teacher answers) are nested over images and refits or combined from both standard errors, as the captions state (\cref{sup:seeds}; model and software versions: \cref{sup:repro}). %

\paragraph{Scores} All scores use the generation pass, and \cref{sup:scores} lists them with their sources. The residual is added to fifteen single-pass scores. Nine are \emph{hidden-state probes}: three label-free probes fitted on the training items to the sampling teacher (the \emph{sampling-teacher probe}, as in semantic-entropy probes~\cite{kossen2024sep,ax2604.19444}), to the shift teacher or to both (the two-teacher probe), and six probes trained with the gold boxes of the training items, namely a linear probe on correctness (the \emph{supervised probe} unless stated otherwise; the accuracy probe of~\cite{kossen2024sep}), a linear probe on $1-\mathrm{IoU}$, the perceptron of SAPLMA~\cite{azaria2023saplma}, a pre-generation probe~\cite{ax2603.05465} and two linear probes on wrong-object labels, which \emph{the seven hidden-state probes} exclude. The six others are the mean log-probability of the coordinate tokens (\emph{coordinate log-prob})~\cite{ax2606.25760,ax2607.05978}; \emph{in-box attention}, the attention of the coordinate tokens to the image tokens inside the written box averaged over all layers (a variant of multi-token localized attention, MTLA~\cite{ax2607.05978}), and the same attention over the whole image, both for the four Qwen models; the Mahalanobis distance of the hidden state~\cite{lee2018mahalanobis}; and two ridges on written-box geometry, one fitted to the teacher, one with gold boxes. Our re-asking baseline, the box's disagreement with its versions under seven perturbed views (content shifts, resampled answers, JPEG, noise), takes eight passes.

%% file: tables/lists.tex
\begin{table}[!t]
\centering
\caption{Training and test lists. Items: number of items, with the number of images for the two RefCOCOg test lists; for Objects365 and VRSBench, items scored per model. Models: all six unless named. Use: what the list serves and when it was first scored.}
\label{tab:lists}
\setlength{\tabcolsep}{2pt}
\begin{tabularx}{\textwidth}{@{}>{\raggedright\arraybackslash}p{1.75cm}>{\raggedright\arraybackslash}p{1.85cm}>{\raggedright\arraybackslash}p{2.5cm}>{\raggedright\arraybackslash}X@{}}
\toprule
List & Items & Source & Models; use\\
\midrule
Training & 2{,}000 rows & RefCOCO val~\cite{yu2016refcoco} & fits the read-out and the probes without labels; labels only for the supervised references\\
Replication & 1{,}200 (901 images) & RefCOCOg test~\cite{mao2016refcocog} & headline results, on images used in no other list\\
Held-out & 1{,}200 (983 images) & RefCOCOg test & scored after the method, its weight and the reference scores were fixed\\
New items & 552 + 600 & RefCOCO testB; RefCOCOg val & scored repeatedly; examined during design for \qsmall{} and \qfour{}\\
RefCOCO+ & 1{,}200 & RefCOCO+ testA; val~\cite{yu2016refcoco} & \qsmall, \qlarge, \qivl\\
Six sets & new items + four sets & RefCOCO testA, testB, val; RefCOCOg val & \qsmall\\
Objects365 & 996 to 1{,}000 & Ref-L4~\cite{chen2024refl4} & \qsmall, \qlarge, \qfour; images outside MS-COCO\\
VRSBench & 997 to 1{,}000 & VRSBench~\cite{li2024vrsbench} & \qsmall, \qfour; remote-sensing images\\
\bottomrule
\end{tabularx}
\end{table}

%% file: sections/plugin.tex
\section{Results: the residual as a plug-in and in SPRED}
\label{sec:plug}

\subsection{Adding the residual to other single-pass scores}
\label{sec:plug:add}

\begin{figure}[!t]
\centering
\includegraphics[width=\textwidth]{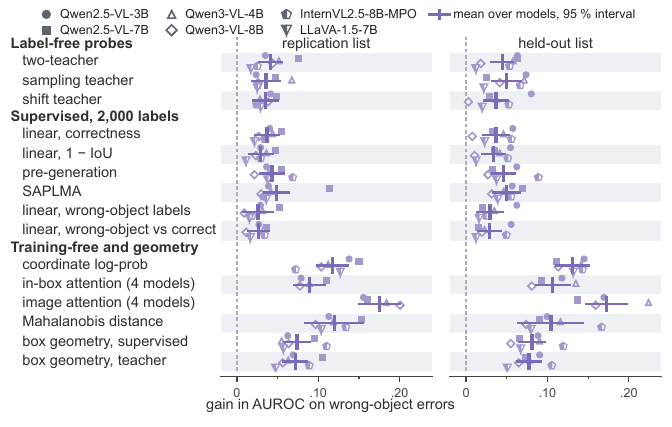}
\caption{Gain in AUROC on wrong-object errors when the residual is added with weight 0.5 to each of fifteen single-pass scores, replication and held-out lists. Markers: models (attention scores: four Qwen models). Tick and line: mean over the models with its combined interval over images and ten training refits. Dashed: no gain. New items in \cref{fig:pluginnewitems}.}
\label{fig:plugin}
\end{figure}

We add the residual with the fixed weight of \cref{eq:combine} to every single-pass score of \cref{sec:setup}, fifteen in all: nine hidden-state probes and six other scores, namely coordinate log-prob, the Mahalanobis distance, two attention scores and two box-geometry scores (\cref{fig:plugin,tab:plugingain}); they share one residual and one set of items, so they are not fifteen independent tests (\cref{sec:method:combine} states when the set was fixed). On the replication list the mean gain in AUROC for wrong-object errors over the models has an interval above zero for each of the fifteen scores, as it does on the held-out list, which was scored before this measure and the scores were fixed (post hoc); the interval combines the resampling of images with ten refits on redrawn training items. The largest gains are on weak scores such as coordinate log-prob; on the six supervised probes trained with 2,000 labels, including the two trained on wrong-object labels themselves, they are 0.026 to 0.049, on supervised two-position ridge probes 0.030 and 0.025 (\cref{tab:replication}), and on the two-teacher probe 0.042. Averaged over the scores, the gain has a combined interval above zero on six of the six models, and in single model--score cells it is positive in every distinct cell of the replication list. %

Replication-list values outside \cref{tab:replication} were computed afterwards with the same fits and are marked post hoc, as are the held-out comparisons other than those on SPRED (\cref{sup:plugin}); post hoc comparisons are not corrected for multiplicity. Other wrong-object boundaries give the same direction (\cref{sup:morebase}). %

A second signal of any kind might raise a score. Under the same rule, no other label-free signal of the pass that we tested adds as much on wrong-object errors: the residual adds more than coordinate log-prob, the Mahalanobis distance, attention and box geometry, and 0.022 more than the two-teacher probe, which comes closest (\cref{tab:addsig}). For review, when the fifth of the answers with the highest score is returned and a reviewed error counts as caught, adding the residual raises the share of wrong-object errors caught from 0.469 to 0.575, averaged over the model--score cells, and, over all error types, the share caught by 0.032. For SPRED against its probe (post hoc), the share of wrong-object errors caught rises from 0.505 to 0.607, while the share of errors of any type caught moves only from 0.518 to 0.524: the residual changes which errors are found, not how many. \Cref{tab:spredrep} gives SPRED's share per model, \cref{fig:reviewcost} over review budgets. %

The gain is specific to the error type the residual targets: paired within each resample (post hoc), it exceeds the gain on partial-overlap and on wrong-extent errors on average over the seven hidden-state probes on both lists (\cref{tab:gaindiff}), and on these two types adding the residual slightly lowers the hidden-state probes on the replication list, with a combined interval below zero for three of the seven on partial overlap (\cref{tab:bytypefull}). The residual also raises the detection of wrong-object errors by checks that need an extra pass, such as a shifted view or five sampled answers, in every model--check cell (\cref{tab:extraadd}). When the model writes a point instead of a box, an answer is right when the point lies in the annotated box, so only the object counts. With the read-out fitted to the model's own points on 968--988 unlabelled training items, the distance between the point read out and the point written serves as residual (\cref{sup:extra}); on the replication and held-out lists (post hoc) it raises the two-teacher probe's detection of all errors by 0.035 [0.015, 0.055] on average over the three models, by 0.070 and 0.020 on \qsmall{} and \qfour{}, the two models used to choose the weight, and by 0.015 on \qlarge{}. \qsmall{} was prompted for point coordinates; as \qfour{} answered some trial items with a box under that prompt, it was prompted for the point under a named JSON key, as was \qlarge{} (\cref{tab:pointlists}).

Among wrong answers, the residual also separates boxes on another object from the other errors with an AUROC of 0.726, averaged over the six models, whereas the probes trained for correctness or without labels and coordinate log-prob reach 0.457 to 0.579. Part of this is box size: a logistic classifier on the written box's geometry, trained on wrong training answers, reaches 0.668, and adding the residual raises it by 0.061 [0.031, 0.092] (\cref{tab:errtype}).

\input{tables/plugingain}

\subsection{SPRED against prior scores and supervised probes}
\label{sec:plug:detector}

\Cref{tab:spredrep} reports SPRED on the replication list, next to its two parts and to prior single-pass scores, and \cref{tab:detectho} on the held-out list. On the replication list the probe alone is the better part for all errors on six of the six models and the residual alone for wrong-object errors on four, and SPRED is above its probe on wrong-object errors on six. On all errors SPRED minus its probe is 0.007 [$-$0.003, 0.016], positive on four of the six models; both ends of the interval lie within 0.02, a bound read off the interval rather than set in advance. SPRED is ahead of coordinate log-prob on every model. Against the supervised linear probe, trained on 2,000 labelled RefCOCO validation items and applied unchanged to the test lists, the one-sided lower bound of the difference exceeds $-0.03$ on six of the six models on the replication list and four on the held-out list, where \qeight{} and \qllava{} fall short; the probe alone does so on four and four. With SPRED fitted on as many unlabelled items as the probe has labels, up to 10,000 for \qsmall{} and \qlarge{}, SPRED minus the linear probe trained on correctness at the largest size is 0.013 ($-$0.010 for \qsmall{}, 0.036 for \qlarge{}), with a one-sided lower bound of $-$0.007, which exceeds $-0.03$ (\cref{sup:tenk}). Of the other supervised probes compared by cross-validation, the strongest is also linear, trained on $1-\mathrm{IoU}$; SPRED minus it is 0.012 (bound $-$0.009). On all errors SPRED is ahead of our reimplementation of BICR at 2,000 labels, in which its blanked-image ranking and Brier terms did not raise the AUROC; with the residual added, BICR is not detectably different from SPRED on wrong-object errors and below it on all errors. New items and RefCOCO+ are in \cref{sup:natural}. %

On wrong-object errors, averaged over the six models, SPRED is not behind the supervised probe trained on correctness (0.037 on the replication list), and against one trained on wrong-object labels the interval includes zero (new items, held-out list). Handing on the top fifth leaves an error rate of 0.105 among the rest on average, close to the supervised linear probe and below coordinate log-prob; the probe alone gives about the same.

\input{tables/spredreplication}

\Cref{tab:ablationmain} removes each part of SPRED on the sets used when the weight was chosen (\qsmall, and the new items of \qfour) and on the new items of \qlarge, examined after it was fixed. The probe carries the overall ranking and the residual adds the wrong-object errors: with half weight it raises the AUROC for these errors over the probe alone by 0.049 to 0.063. The last two rows compare the two-teacher probe with single-teacher probes (more: \cref{tab:teachers}).

\input{tables/ablationmain}

\subsection{Few labels, a missing error type and other images}
\label{sec:plug:offideal}

\Cref{tab:offideal} collects settings where a supervised probe loses its advantage; entries are SPRED minus supervised probe, new items unless named. With 250 labelled items, SPRED fitted on the same items without labels has a higher mean than the supervised linear probe on all five models of the new items and on all six of the replication list; there it has a nested interval over images and subsets of items above zero on four (\qfour{} and \qeight{} excepted). Trained without one type of error, supervised probes fall below SPRED on that type in all ten model--type cells of the new items, also with that type removed from SPRED's unlabelled training items; on Objects365 images, with both transferred from RefCOCO, SPRED is ahead of the supervised linear probe on \qsmall{} and no difference is detected on the other two models (\cref{sup:domains}). %

%% file: tables/plugingain.tex
\begin{table}[!t]
\centering
\caption{Gain in AUROC on wrong-object errors when the residual is added with weight 0.5 (\cref{eq:combine}) to each single-pass score, replication list; mean over the models of the stored gain. $^{L}$label-free probe; $^{S}$supervised probe, 2,000 labels (W: trained on wrong-object labels, on all items or on wrong-object and correct items only); $^{T}$training-free or box-geometry score; $^\ast$four Qwen models. Bold: combined interval, stored gain $\pm 1.96\sqrt{\mathrm{SE}^2+\mathrm{SD}^2}$ with SE over images and SD over ten refits on redrawn training items, excludes zero. Other error types in \cref{tab:bytypefull}, the held-out list in \cref{tab:bytypeho}, per-model cells and the new items in \cref{sup:plugin}.}
\label{tab:plugingain}
\setlength{\tabcolsep}{1.5pt}
\begin{tabular}{@{}lc@{\hspace{0.5em}}lc@{\hspace{0.5em}}lc@{}}
\toprule
Score & Gain & Score & Gain & Score & Gain\\
\midrule
two-teacher$^{L}$ & \textbf{0.042} & pre-generation$^{S}$ & \textbf{0.043} & in-box attention$^{T\ast}$ & \textbf{0.089}\\
sampling teacher$^{L}$ & \textbf{0.036} & SAPLMA$^{S}$ & \textbf{0.049} & image attention$^{T\ast}$ & \textbf{0.175}\\
shift teacher$^{L}$ & \textbf{0.035} & linear, W$^{S}$ & \textbf{0.026} & Mahalanobis$^{T}$ & \textbf{0.120}\\
linear, correctness$^{S}$ & \textbf{0.037} & linear, W vs correct$^{S}$ & \textbf{0.026} & geometry, sup.$^{T}$ & \textbf{0.074}\\
linear, $1-\mathrm{IoU}$$^{S}$ & \textbf{0.029} & coord.\ log-prob$^{T}$ & \textbf{0.117} & geometry, teacher$^{T}$ & \textbf{0.072}\\
\bottomrule
\end{tabular}
\end{table}

%% file: tables/spredreplication.tex
\begin{table}[!t]
\centering
\caption{SPRED against prior single-pass scores on the replication list (1{,}200 RefCOCOg test items on 901 images; \cref{sec:setup}). Columns: the six models (3B, 7B: Qwen2.5-VL; Q3-4B, Q3-8B: Qwen3-VL; IVL: InternVL2.5-8B-MPO; LLaVA: LLaVA-1.5-7B). AUROC; wrong-object errors (IoU $<0.1$) are scored against correct answers. Linear probe: trained on the correctness of 2,000 labelled RefCOCO items (the accuracy probe of~\cite{kossen2024sep}). $^{\dagger}$Our reimplementation, trained on the same items. Probe alone, residual alone: the two parts of SPRED. Bound: one-sided 95\% lower bound of SPRED minus the linear probe (5th percentile of a bootstrap over images). Bold: highest AUROC in each column of each block; on IVL, SPRED and the probe alone agree to three decimals on all errors. Top fifth (post hoc): share of the wrong-object errors among the fifth of the answers with the highest SPRED score, and answers reviewed per wrong-object error found; means over the models: SPRED 0.607 and 6.28, its probe alone 0.505 and 7.59, random 0.200 and 18.37. Held-out list: \cref{tab:detectho}; further prior scores (SAPLMA, sampling-teacher probe, Mahalanobis distance, MTLA) on the new items: \cref{tab:detect}.}
\label{tab:spredrep}%
\setlength{\tabcolsep}{2.2pt}
\begin{tabular}{@{}lcccccc@{}}
\toprule
& 3B & 7B & Q3-4B & Q3-8B & IVL & LLaVA\\
\midrule
Items & 1{,}200 & 1{,}199 & 1{,}198 & 1{,}197 & 1{,}197 & 1{,}200\\
Errors & 176 & 152 & 149 & 132 & 127 & 457\\
\midrule
\multicolumn{7}{@{}l}{\textit{All errors (IoU $<0.5$)}}\\
Coordinate log-prob~\cite{ax2606.25760} (2026) & 0.633 & 0.665 & 0.616 & 0.615 & 0.612 & 0.658\\
Linear probe~\cite{kossen2024sep} (2024) & 0.778 & 0.769 & 0.765 & 0.780 & 0.793 & 0.819\\
BICR$^{\dagger}$~\cite{ax2605.10893} (2026) & 0.727 & 0.745 & 0.770 & 0.753 & 0.684 & 0.765\\
SPRED (ours) & 0.792 & \textbf{0.785} & \textbf{0.795} & \textbf{0.781} & \textbf{0.810} & 0.812\\
\quad probe alone & \textbf{0.794} & 0.765 & 0.782 & 0.764 & \textbf{0.810} & \textbf{0.820}\\
\quad residual alone & 0.658 & 0.699 & 0.708 & 0.698 & 0.691 & 0.717\\
Bound vs.\ linear probe & $-$0.011 & $-$0.017 & 0.008 & $-$0.024 & $-$0.008 & $-$0.021\\
\midrule
\multicolumn{7}{@{}l}{\textit{Wrong-object errors (IoU $<0.1$)}}\\
BICR$^{\dagger}$~\cite{ax2605.10893} (2026) & 0.787 & 0.774 & 0.819 & \textbf{0.826} & 0.639 & 0.832\\
SPRED (ours) & \textbf{0.863} & 0.815 & \textbf{0.831} & 0.815 & \textbf{0.757} & \textbf{0.894}\\
\quad probe alone & 0.828 & 0.739 & 0.779 & 0.771 & 0.732 & 0.877\\
\quad residual alone & 0.802 & \textbf{0.824} & 0.830 & 0.802 & 0.744 & 0.845\\
\quad SPRED $-$ linear probe & 0.075 & 0.020 & 0.049 & 0.019 & 0.033 & 0.024\\
SPRED, top fifth: share caught & 0.684 & 0.527 & 0.653 & 0.650 & 0.511 & 0.614\\
\quad answers reviewed per catch & 4.62 & 8.28 & 5.11 & 6.15 & 10.43 & 3.08\\
\bottomrule
\end{tabular}
\end{table}

%% file: tables/ablationmain.tex
\begin{table}[!t]
\centering
\caption{Ablation of SPRED. AUROC for all errors (All; IoU $<0.5$) against all other items, and for wrong-object errors (WO; IoU $<0.1$) against correct answers. \qsmall: six sets, and \qfour: new items, both used when the weight was chosen; \qlarge: new items, examined after it was fixed. Lower block: differences. Bold: highest in each column of the upper block. Intervals in \cref{tab:ablationdiff}, further variants in \cref{tab:ablation}.}
\label{tab:ablationmain}
\setlength{\tabcolsep}{3pt}
\begin{tabular}{@{}>{\raggedright\arraybackslash}p{3.3cm}cccccc@{}}
\toprule
 & \multicolumn{2}{c}{\qsmall} & \multicolumn{2}{c}{\qlarge} & \multicolumn{2}{c}{\qfour}\\
\cmidrule(lr){2-3}\cmidrule(lr){4-5}\cmidrule(lr){6-7}
Variant & All & WO & All & WO & All & WO\\
\midrule
Probe only & 0.813 & 0.829 & 0.833 & 0.818 & 0.831 & 0.805\\
Residual only & 0.740 & 0.875 & 0.725 & 0.833 & 0.773 & \textbf{0.867}\\
SPRED & \textbf{0.831} & \textbf{0.885} & \textbf{0.848} & \textbf{0.866} & \textbf{0.858} & \textbf{0.867}\\
\midrule
SPRED $-$ probe only & 0.018 & 0.056 & 0.016 & 0.049 & 0.027 & 0.063\\
SPRED $-$ residual only & 0.091 & 0.010 & 0.123 & 0.033 & 0.086 & $-$0.000\\
Probe: two $-$ shift teacher & 0.016 & 0.008 & 0.027 & $-$0.034 & 0.028 & $-$0.017\\
Probe: two $-$ sampling teacher & 0.013 & $-$0.019 & 0.031 & $-$0.040 & 0.007 & 0.022\\
\bottomrule
\end{tabular}
\end{table}

%% file: sections/measures.tex
\section{What the residual measures}
\label{sec:measure}

\subsection{It depends on the state of the item at hand}
\label{sec:measure:own}

The residual is not a prior over box positions: with the read-out of another item in place of the item's own, its AUROC for wrong-object errors falls from a mean of 0.808 over the six models to 0.545 to 0.626, each model's own value above all permuted ones (\cref{tab:readctl}). With the mean residual of the other five models on the same item in its place, the added term keeps the wrong-object gain (post hoc, \cref{tab:ownother}). Averaged over the seven hidden-state probes, the ratio of the others' gain to the own gain is 1.01 [0.77, 1.30] on the replication list and 0.89 on the held-out list. The residual of a single other model, added to the two-teacher probe, gains 0.032 against 0.042 with the own one. The signal is thus largely about the item; \cref{sec:discussion} returns to what the checked model's pass adds. On items that Ref-L4 re-checked, about a third of the wrong-object errors fall on annotations it found faulty (0.315 and 0.381 of model--item pairs on the new items and RefCOCO+; \cref{sup:pull}).

\subsection{Where the answer changes, the new box moves towards the read-out}
\label{sec:measure:second}

\begin{figure}[!t]
\centering
\includegraphics[width=\textwidth]{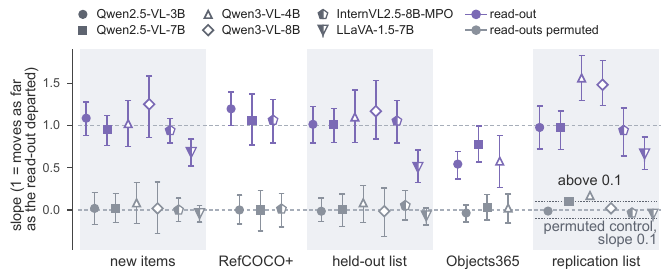}
\caption{The second answer moves with the read-out. Items whose answer changes under the content shift. Lavender: slope of the second box's displacement on the read-out's (95\% interval over images). Grey: permuted read-outs (mean and central 95\%; mean only on the replication list). Dashed: 0 and 1. Dotted: slope 0.1.}
\label{fig:second}
\end{figure}

Among items whose answer changes under the content shift (IoU of the two boxes below 0.5), the displacement of the second box centre, regressed on that of the read-out with the mean training box as a second regressor and no intercept, has a slope of 0.942 to 1.251 on the five Qwen and InternVL models across the new items, RefCOCO+ and the held-out list, near zero with permuted read-outs (\cref{fig:second}). On the replication list the slope is at least 0.6, with a lower end above 0.3 and below 0.1 for permuted read-outs, on five of the six models, as on the held-out list though not on the same models; these values were chosen with the held-out results in view (\cref{tab:replication}). The slope remains with the annotated centre as a third regressor and among items whose second answer is also wrong, so it is not a pull towards the annotated box (\cref{tab:secondans}). The read-out gives the direction, not the position: it is nearer the second box than the original one in only 0.293 to 0.553 of the changed items. %

\subsection{On wrong-object errors, the read-out lies towards the annotated object}
\label{sec:measure:pull}

\begin{figure}[!t]
\centering
\includegraphics[width=\textwidth]{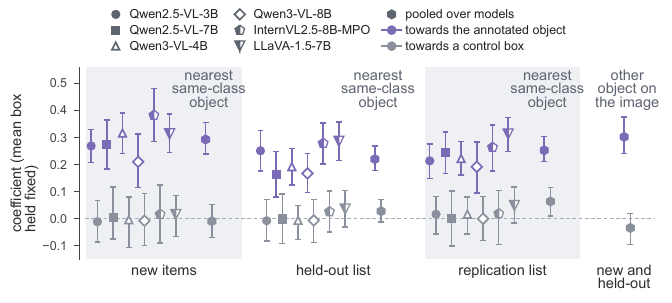}
\caption{The read-out leans towards the annotated object far more than towards other boxes. Wrong-object errors: coefficient of a box centre, relative to the written box, for the read-out centre; 95\% intervals over images; grey model bars: central 95\% over permutations. Per model: annotated object (lavender), another error's annotated box (grey). Hexagons, pooled: nearest same-class object, other object on the image.}
\label{fig:pull}
\end{figure}

Regressing the displacement of the read-out centre from the written one on that of the annotated centre and of the mean training centre, without intercept so that the second term absorbs the shrinkage of a ridge map, gives a coefficient on the annotated centre of 0.250 [0.212, 0.292] over the six models on the replication list; every per-model interval is above zero on both lists, and with the annotated centre of another wrong-object error in its place it is near zero (\cref{fig:pull}). On the replication list, 72.6\% of these errors lie on another annotated MS-COCO instance (IoU at least 0.5), and on them SPRED is ahead of its probe by 0.048, as on all wrong-object errors (post hoc). Referential ambiguity, a state mixing several candidates, would place the read-out between them whatever box is written; yet it is displaced far more towards the target than towards the nearest other object of the target's class (difference 0.189 [0.122, 0.269]), although the displacement towards that object is itself above zero (0.064, \cref{tab:replication}), and in correct answers no displacement towards that object is detected (\cref{tab:sameclass}); the data are also compatible with a mixture with unequal weights. Within strata of the number of target-class objects (post hoc), the residual's gain on wrong-object errors remains on the replication list: 0.040 pooled over the original strata, one of them empty, and 0.037 [0.020, 0.053] over strata redefined afterwards (\cref{sup:pull}). On TextVQA, several open multimodal models give the answer region more attention than the average same-size box in most layers, also when their answer is wrong~\cite{zhang2025mllmsknow}; the read-out's displacement is a related observation for boxes, from a fitted map.

\begin{figure}[!t]
\centering
\includegraphics[width=\textwidth]{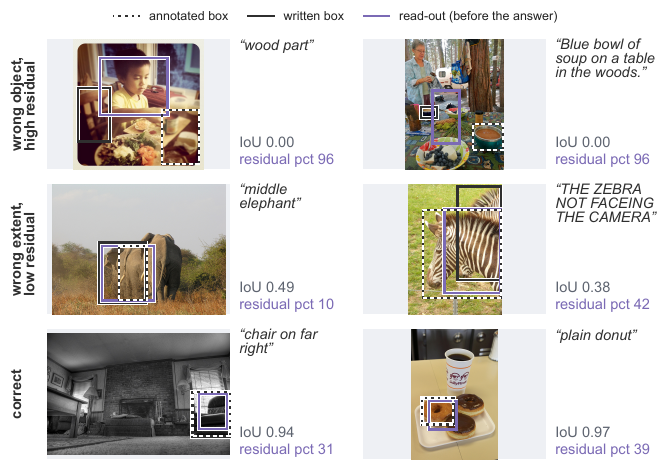}
\caption{What the residual sees and what it does not. \qsmall{}, new items; two items per row, drawn at random among photos whose licence allows reuse. Black: written box; lavender: read-out; white dashes: annotated box; beside each photo, IoU and residual percentile. Photographs: MS-COCO, CC BY 2.0, credited in the supplementary material.}
\label{fig:examples}
\end{figure}

\subsection{What it does not measure}
\label{sec:measure:not}

Alone the residual is weak on boxes of the right object with the wrong extent (\cref{fig:examples}): on the six sets of \qsmall{} its AUROC for these errors is 0.604, against 0.786 for the two-teacher probe, while for wrong-object errors it is the stronger of the two. Its gain for wrong-object errors also shrinks as the target narrows relative to the read-out error $\sigma$ (mean of the seven hidden-state probes, three models; \cref{sec:method:use,sup:domains}). We found this dependence post hoc, on Objects365 with in-domain read-outs, since RefCOCO-family sets have hardly any written boxes narrower than $2\sigma$.

%% file: sections/discussion.tex
\section{Discussion}
\label{sec:discussion}

\paragraph{Precise enough to detect, not to correct}
The read-out is coarse, with an error $\sigma$ of 0.069 to 0.081 of the image side on RefCOCO: enough to flag a box that departs from the state before writing, not to say where it should be, so the residual ranks answers for review rather than replacing the box. At a fixed review budget it changes which errors are found rather than how many (\cref{sec:plug:add}), so its value depends on the cost of errors: in the share of the total error cost caught with a fifth of the answers reviewed, SPRED breaks even against its probe when a wrong-object error costs about 0.86 times another error and at equal costs its gain is 0.006 in point estimate (replication list, post hoc; \cref{fig:reviewcost}). In counts (replication list, a fifth reviewed, post hoc), over six models SPRED catches 42 more wrong-object errors than its probe and 41 fewer others; on \qsmall{}, the example named before counting, 52 of its 76 against 39, and on \qlarge{} 1 more.

\paragraph{What re-asking cannot see}
The items whose box all seven views keep (\cref{tab:stable}) are selected by a failure of re-asking, which partly misses errors the model repeats (\cref{sec:intro}). On them SPRED leads the seven-view disagreement on each model, which shows where it adds most rather than how the two compare in general; it still reaches an AUROC of 0.748 to 0.870. On all new items, one pass of SPRED is ahead of the eight passes of re-asking (original and seven views) on the two Qwen3-VL models, by +0.078 and +0.070, and on the two Qwen2.5-VL models no difference is detected; the new items of \qfour{} were examined while the method was designed, and \qeight{} was run after it was fixed (\cref{sup:clean}). On the new items of \qsmall{}, \qlarge{} and \qfour{}, the disagreement of a frozen zero-shot open-vocabulary detector's top box with the written box adds no detected gain to the probe on wrong-object errors (0.001). The residual adds 0.057 [0.031, 0.083] more than that detector's disagreement and still 0.049 once it is added, positive on all three models (\cref{tab:external}).

\paragraph{Error scores and item difficulty}
Error scores can rank items by how hard they are for any model rather than by the checked model's state. On the new items that one of five models answers correctly and another wrongly, which hold difficulty fixed (\cref{sup:ctl}), SPRED picks the model that erred, counting errors of any type, in 0.634 of the cases, the residual alone in 0.599 and the supervised linear probe, trained on the correctness of 2{,}000 items, in 0.565, so on these items the order of the scores changes in favour of the residual in point estimates (\cref{sec:ctl:difficulty}); a signal shared by the two models reaches 0.5. Since the others' residual on the same item adds about as much to a score (\cref{sec:measure:own}), much of the signal belongs to the item. Not all of it: counting only the pairs in which the erring model placed its box on another object, the residual picks that model in 0.637 [0.551, 0.722] of 406 pairs on the new items (combined over images and refits) and in 0.608 [0.543, 0.674] of 559 on the replication list (exploratory; images only), so at fixed difficulty it still picks the erring model above chance. We read it as a second reading within the checked model's pass, not knowledge private to the model. %

\paragraph{Limitations}\label{sec:limits}
The residual addresses one error type, boxes on another object. Its gains were measured on RefCOCOg lists of MS-COCO images. Fitted without labels, SPRED reaches an AUROC of 0.692 to 0.762 on Objects365 and 0.752 and 0.784 on VRSBench remote-sensing images, but on annotated targets narrower than about twice the read-out error, a dependence found post hoc, SPRED minus its probe alone on all errors is $-$0.032 and $-$0.032 in AUROC with the read-out fitted on the domain; 0.599 to 0.665 and 0.543 to 0.590 of the written boxes are narrower than $2\sigma$ (\cref{sup:domains}). The residual's weight was fixed with labels on two of the models. A risk-guaranteed threshold needs target-domain labels (\cref{sup:ltt}). %

%% file: sections/conclusion.tex
\section{Conclusion}
\label{sec:conclusion}

We introduced the self-prediction residual: the disagreement between a grounded VLM's written box and the box that a linear map fitted to its own answers reads from its state before writing. Fitting it for a model or domain needs no labels and adds no pass at test time; the weight was fixed once with labels. Added with a fixed weight to fifteen single-pass scores on six models, the two-teacher probe and linear probes trained on 2,000 labelled items included, it raised their detection of wrong-object errors, with the mean gain over models above zero for every score on the replication list, scored after this measure was fixed, and, post hoc, on the held-out list. On all errors, SPRED minus the linear probe trained on 2,000 labelled RefCOCO items is 0.012 averaged over the six models on the replication list; single models range from $-$0.007 to 0.030 there and from $-$0.014 to 0.026 on the held-out list (\cref{tab:spredrep,tab:heldout}). Over all errors on the replication list, adding the residual to the two-teacher probe changes the AUROC, averaged over the models, by less than 0.02 (95\% interval); at a fixed review budget it changes which errors are found, not how many (post hoc). These results come from RefCOCOg lists of MS-COCO images. %

For models like the six tested, single-pass error detectors can add the residual, with the map fitted on unlabelled in-domain items, when wrong-object boxes are the errors to catch and under half the written boxes are narrower than about twice the read-out error, a post hoc condition for new domains and models to confirm.

A linear read-out on one layer limits the residual on small targets; a nonlinear or multi-layer one could help there. Wrong-extent errors need other internal signals.

%% file: sections/declarations.tex
\section*{CRediT authorship contribution statement}
\textbf{Enhao Cui:} Conceptualization, Methodology, Software, Formal analysis, Investigation, Writing -- original draft, Writing -- review \& editing, Project administration. \textbf{Junzhi Yang:} Software, Data curation, Investigation, Writing -- review \& editing. \textbf{Yaoshun Lin:} Resources, Validation, Visualization, Writing -- review \& editing. \textbf{Yuhan Feng:} Data curation, Investigation, Visualization, Writing -- review \& editing. \textbf{Zijian Chen:} Methodology, Formal analysis, Validation, Writing -- review \& editing.

\section*{Declaration of competing interest}
The authors declare that they have no known competing financial interests or personal relationships that could have appeared to influence the work reported in this paper.

\section*{Funding}
This research did not receive any specific grant from funding agencies in the public, commercial, or not-for-profit sectors.

\section*{Data availability}
All benchmarks and checkpoints used are public (sources and revisions in \cref{sup:repro}). The item lists, per-item answers, read-out boxes and scores of the replication, held-out, new-item, RefCOCO+, Objects365 and VRSBench sets and the point answers on the replication and held-out lists are in a supplementary data file with scripts that recompute the main reported numbers, including those of \cref{tab:lists,tab:plugingain,tab:spredrep,tab:ablationmain} and \cref{fig:plugin,fig:second,fig:pull}; the code and reproduction scripts are at \url{https://github.com/chen4study-hub/spred} (Apache-2.0).

\section*{Declaration of generative AI and AI-assisted technologies in the manuscript preparation process}
During the preparation of this work the authors used Claude in order to improve language and correct grammar. After using this tool, the authors reviewed and edited the content as needed and take full responsibility for the content of the publication.

%% file: sections/vitae.tex
\section*{Author biographies}
\begin{sloppypar}

\noindent\textbf{Enhao Cui} is currently pursuing the B.S. degree in computer science at Changzhou University, Changzhou, China. Enhao Cui's research interests include vision-language models, the interpretability of neural networks, and uncertainty quantification.

\noindent\textbf{Junzhi Yang} is currently pursuing the B.S. degree in computer science at Changzhou University, Changzhou, China.

\noindent\textbf{Yaoshun Lin} is currently pursuing the B.S. degree in computer science at Changzhou University, Changzhou, China.

\noindent\textbf{Yuhan Feng} is currently pursuing the B.S. degree in computer science at Changzhou University, Changzhou, China.

\noindent\textbf{Zijian Chen} is currently pursuing the B.S. degree in computer science at Changzhou University, Changzhou, China.
\end{sloppypar}

%% file: sections/supplement.tex
\section{Reproducibility}
\label{sup:repro}

\paragraph{Models and data}
\begin{sloppypar}
Qwen/Qwen2.5-VL-3B-Instruct (revision \texttt{66285546d2b8}), Qwen/Qwen2.5-VL-7B-Instruct (\texttt{cc594898137f}), Qwen/Qwen3-VL-4B-Instruct (\texttt{ebb281ec70b0}), Qwen/Qwen3-VL-8B-Instruct (\texttt{0c351dd01ed8}), OpenGVLab/InternVL2\_5-8B-MPO-hf (\texttt{543db189852e}) and llava-hf/llava-1.5-7b-hf (\texttt{b234b804b114}) from Hugging Face, and the external detector IDEA-Research/grounding-dino-tiny (\texttt{a2bb814dd30d}) of \cref{tab:external}; RefCOCO (lmms-lab-encoder/RefCOCO, \texttt{a5dff0b31947}), RefCOCOg (lmms-lab-encoder/RefCOCOg, \texttt{93af55545967}) and RefCOCO+ (lmms-lab/RefCOCOplus, \texttt{a283082593c4}); the Objects365 part of Ref-L4 (JierunChen/Ref-L4; original expressions, one per box, boxes covering at least 1\% of the image, training and test images disjoint). The read-out and the probes of the detector are fitted on the 2,000 training items below; the new error-detection items are drawn with a fixed seed from RefCOCO testB and RefCOCOg validation, excluding every image used before the method was designed; the detector's training items are an image-level split of RefCOCO validation items whose images are disjoint from all test items. %
\end{sloppypar}

\paragraph{Prompts, generation and hidden states}\label{sup:prompts} Each item is one user turn with the image followed by the text of \cref{tab:prompts}, formatted with the chat template of the model, applied unchanged (with its default system message where the template has one), and its generation prompt; nothing else is in the context. For the Qwen models the processor resizes the image to a multiple of the patch-merging unit (28 pixels for Qwen2.5-VL, 32 for Qwen3-VL) with between $256\times28^2$ and $1024\times28^2$ pixels in total; Qwen2.5-VL writes coordinates in pixels of the resized image, Qwen3-VL on a scale from 0 to 1000. \qivl{} uses the processor's tiling with at most six tiles plus the thumbnail. For \qllava{} the image is centred on a square canvas filled with the processor's mean colour, and the box is read on the square and mapped back to the image. Decoding is greedy with at most 64 new tokens (48 for \qllava), with weights in bf16. The sampling teacher draws one further answer per training item with temperature 0.7 and top-$p$ 0.9; the shift teacher repeats the greedy pass on the image padded on the left by a strip in its mean colour (two patch-merging units for the Qwen models, 64 pixels for the other two) and shifts the box back. In the Qwen models the strip moves the image content by two token positions in the rotary position code, whose role in discrete left--right answers is studied in~\cite{cui2026leftright}. The hidden states are taken from one forward pass over the prompt followed by the model's own greedy answer, which reproduces the states of the generation pass. Two positions are stored at every fourth layer of the language model and at its last layer (layer 0 is the embedding output): the last token of the generation prompt, before any answer token, and the mean over the positions of the coordinate digits of the answer. The read-out uses the first of the two only; its layer and ridge penalty are in \cref{tab:prompts}. %

\input{tables/prompts}

\paragraph{Environment and implementation}
Inference used Python 3.12.3, PyTorch 2.5.1 with CUDA 12.4 and transformers 4.57.6 on a single 32\,GB GPU, bf16 unless stated, greedy decoding. %

The supervised probes use the same features (the state at the last prompt token, before generation, and the mean state over the coordinate tokens), the same stored layers, the same penalties ($\lambda\in\{10^2,10^3,10^4,10^5\}$) and the same five-fold selection on the training items as the label-free probes, with the selection scored by the AUROC against the correctness labels; the probe on $1-\mathrm{IoU}$ is selected likewise. The perceptron uses the features selected for the linear probe, hidden layers of 256, 128 and 64 units, 50 epochs, learning rate $10^{-3}$, weight decay $10^{-4}$ and seed 0, without tuning. With the seed fixed, the perceptron's fit still depends on the number of CPU threads; all reported values, including those of the perceptrons in \cref{tab:equalinfo}, use four threads, and fits with one to eight threads change its AUROC on the new items by 0.009 to 0.052 depending on the model (mean 0.022). On \qlarge{} the linear probe on correctness selects the state before generation with the same layer and penalty as the pre-generation probe, so the two give identical scores there and count as one baseline. %

BICR reads, as in the original, the last layer at the last prompt token. On all errors it differs from the supervised linear probe that reads the same token with the layer chosen on the training items by $-$0.019 [$-$0.028, $-$0.010] on the held-out list and $-$0.016 [$-$0.024, $-$0.008] on the new items, below it on five of the six models and on all six, respectively; BICR differs from the same probe trained without the ranking and Brier terms by $-$0.001 and $-$0.002. Intervals resample images only.

\paragraph{Cost} Measured with batch size 1, the detector takes 0.997, 1.001 and 1.126 times the time of plain generation and peaks at 7.17, 15.69 and 8.47~GB of GPU memory, against 7.12, 15.59 and 8.40~GB for plain generation (\qsmall, \qlarge, \qfour); each further generation takes 1.010 to 1.012 times a plain one. Asked only above the 70\% training quantile of the detector, the second model brings the cost to 1.36 to 1.41 passes in all on average (\cref{sup:second}).

\input{tables/costs}

\paragraph{MTLA} As in the paper that proposed multi-token localized attention (MTLA)~\cite{ax2607.05978}, the score is the attention mass that the generated tokens of the box that contain digits give to the image tokens whose centres lie inside the predicted box, not normalized, averaged over heads, over layers 8 to 21 (6 to 16 of the 28 layers of \qlarge, the same relative depth) and over the four tokens, read in the generation pass; low attention is suspicious. There it detects hallucinated objects in COCO detection, objects that are predicted but absent. In referring-expression grounding most errors put the box on another real object or give it the wrong extent, so the attention inside the box can stay high. Other layer bands, the maximum over heads, normalization by the attention to the whole image or a layer chosen on labelled training items keep it between 0.536 and 0.638, and on wrong-object errors alone it reaches 0.631 to 0.721. The attention to the whole image stays near chance whichever direction is used (0.491 to 0.567 with the direction set on the training items). On its original task the same implementation reproduces the reported score: with \qeight, the prompt of~\cite{ax2607.05978} that lists every instance of the 80 COCO classes and one sampled answer for each of 200 COCO val2017 images (2{,}337 predicted boxes, 0.615 of them hallucinated), the definition of~\cite{ax2607.05978} (coordinate and label tokens, image tokens mapped by overlap) reaches an AUROC of 0.886, against 0.890 reported there; the centre-in-box variant used on referring expressions reaches 0.840, and mapping by overlap without the label tokens 0.887, so the difference comes from the centre rule. The attention to the whole image reaches 0.875 and coordinate log-prob 0.636 there. Part of the score on that task comes from the position of a prediction in the long answer list (11.7 boxes per image), where later boxes are more often hallucinated: the position alone reaches 0.834, and within strata of position MTLA reaches 0.771, still above its 0.547 to 0.609 on referring-expression errors, where a single box is written and a wrong box points at another object that is present. Three variants enter the tables, and they are different scores: \cref{tab:detect,tab:detectdiff} use the mapping of the original paper in a middle band of layers fixed per model family; \cref{tab:metrics2}, \cref{sup:ctl,fig:controlled} and the referring-expression numbers of this paragraph the centre-in-box rule above with layers 8 to 21; and the plug-in analyses (\cref{tab:plugingain,fig:plugin}, \cref{sup:plugin}), where the score is called in-box attention, the same centre-in-box attention averaged over all layers and heads instead of the band. The tables name them MTLA (original mapping), MTLA (layers 8--21) and in-box attention; on the items all three score, the rank correlation of in-box attention is 0.963 to 0.988 with the layer-band version and 0.699 to 0.849 with the original mapping. %

\paragraph{Scores and their sources}\label{sup:scores} All scores use the generation pass. Training-free: the mean log-probability of the coordinate tokens (coordinate log-prob)~\cite{kossen2024sep,ax2606.25760,ax2607.05978}; in-box attention, the attention of the coordinate tokens to the image tokens inside the written box averaged over all layers (a variant of MTLA~\cite{ax2607.05978}), and the same attention over the whole image, both defined for the four Qwen models; Chain-of-Embedding~\cite{wang2025coe}, LLM-Check~\cite{sriramanan2024llmcheck} and the box decoded by the model's own output head from an intermediate layer (the logit lens, the unfitted baseline of~\cite{belrose2023tunedlens}; applied to image representations, the output head also localizes objects~\cite{jiang2025vlinterp}). Label-free and fitted on the training items: the Mahalanobis distance of the hidden state~\cite{lee2018mahalanobis}, the probes with the sampling teacher alone (sampling-teacher probe, as in semantic-entropy probes~\cite{kossen2024sep,ax2604.19444}), with the shift teacher alone, and with both, and a ridge on the geometry of the written box fitted to the teacher. Supervised, with the gold boxes of the training items: a linear probe on correctness (the accuracy probe of~\cite{kossen2024sep}), a linear probe on $1-\mathrm{IoU}$, the three-layer perceptron of SAPLMA~\cite{azaria2023saplma}, a pre-generation probe~\cite{ax2603.05465}, two linear probes on wrong-object labels (all training items, and wrong-object and correct items only), and a ridge on box geometry. The nine probes on hidden states are the hidden-state probes; the seven hidden-state probes exclude the two trained on wrong-object labels. Chain-of-Embedding, LLM-Check and the logit-lens box are compared with the detector but are not among the fifteen scores to which the residual is added. Our reimplementation of BICR~\cite{ax2605.10893} replaces its Optuna search with a random search of 40 configurations in its published search space, selected by five-fold out-of-fold AUROC. %

\paragraph{Examples and photo credits} In \cref{fig:examples}, a high residual is at or above the 80th percentile among the new items and a low one at or below the median; wrong-object errors have IoU $<0.1$ and wrong-extent errors $0.3\le$ IoU $<0.5$. %

The photographs of \cref{fig:examples}, by row: ``Screw Irene, it's English Muffin Pizza Night.'' by eric molina (\url{https://www.flickr.com/photos/iamagenious/6086683055/}); ``Camp Breakfast'' by PunkToad (\url{https://www.flickr.com/photos/punktoad/9663077996/}); ``Elephant Study 1'' by U.S. Fish and Wildlife Service Headquarters (\url{https://www.flickr.com/photos/usfwshq/6987534275/}); ``DSC01455'' by davidhc (\url{https://www.flickr.com/photos/davidhc/500671215/}); ``Room with a Fireplace'' by Arturo Donate (\url{https://www.flickr.com/photos/arturodonate/3917206330/}); ``Donuts and Coffee'' by Steven Miller (\url{https://www.flickr.com/photos/aloha75/9492272170/}). The photos, from MS-COCO, are used under CC BY 2.0, resized and with boxes drawn.

\section{Paired differences and intervals}
\label{sup:diff}

\Cref{tab:detectdiff} gives the paired differences between our detector and every baseline of \cref{tab:detect} with their intervals. Coordinate log-prob is weak because its information sits almost entirely in the leading digit: the probability that the hundreds digit of a coordinate changes reaches 0.745, 0.753 and 0.710 on the new items, the units digit only 0.504, 0.496 and 0.535, and averaging over all digits dilutes the signal. In the digit-weighted coordinate uncertainty of \cref{tab:detect}, high digits count more: it is one minus the expected IoU between the written box and boxes in which each digit is redrawn from the single-digit tokens among its five most probable alternatives in the same pass, renormalized (200 draws); PACE instead weights per-digit entropies by place value~\cite{li2026pace}. Added to the digit-weighted uncertainty (post hoc), the residual raises its AUROC for wrong-object errors on the new items by 0.074 [0.057, 0.092], averaged over \qsmall, \qlarge{} and \qfour; over all errors the gain is 0.039 [0.026, 0.052].

In the main comparison, against the probe on $1-\mathrm{IoU}$ the detector is ahead on \qlarge{} and \qfour{} (by 0.049 and 0.027) and no difference is detected elsewhere; on RefCOCO+ for \qsmall{} that probe has the higher point estimate (0.834 against 0.830), the only single-pass baseline that does.

\input{tables/detectdiff}

\Cref{tab:boundsupervised} compares the detector with the supervised linear probe through the one-sided lower bound of their difference, and \cref{tab:surface} gives supervised baselines that see only surface features of the item. Pooled over the three sets examined only afterwards (the new items of \qlarge{} and both RefCOCO+ sets), with a joint bootstrap over images that keeps the two models paired on RefCOCO+, the detector is ahead of the correctness probe by 0.018. With the residual added to both, the two-teacher probe and the supervised probe show no detected difference on five sets (differences between $-$0.004 and 0.012), and ours is ahead by 0.030 on the new items of \qlarge{}.

\Cref{tab:labelsvsdetector} compares the label-free detector with supervised probes trained on as many labelled items (\cref{sec:plug:add}); with matched sizes up to 10,000 items (\cref{sup:tenk}), the probe of \qsmall{} is above SPRED at 10,000 ($-$0.010 on the replication list). %

\input{tables/labelsvsdetector}

\input{tables/boundsupervised}
\input{tables/surface}

The ablation table (\cref{tab:ablation}), its differences and the remaining statistics of the main text are given with their intervals in \cref{tab:ablationdiff,tab:otherdiff,tab:residadddiff}.

\input{tables/ablationdiff}
\input{tables/residadddiff}
\input{tables/risk80}
\input{tables/referral}

The two-teacher probe alone is level with the sampling-teacher probe on all six sets (differences $-$0.028 to 0.031), and a sampling-teacher probe trained on two sampled answers per item, the same teacher cost as ours, shows with the residual added no detected difference from the detector on both Qwen2.5-VL models and on \qivl{} (\cref{tab:teachers}).

\Cref{tab:metrics2} gives two further metrics on the new items (\cref{sec:plug:add}): the area under the precision--recall curve (AUPRC) and the share of correct answers flagged at 95\% recall of errors (FPR95). In paired bootstrap intervals over images, the detector is ahead of coordinate log-prob and of MTLA (layers 8--21) in AUROC, AUPRC and FPR95 on every model where they are defined, ahead of the Mahalanobis distance in AUROC and AUPRC on all five, and ahead of the sampling-teacher probe in AUPRC on \qsmall, \qlarge{} and \qfour{} (by 0.099, 0.132 and 0.056; 0.033 and 0.028 on \qeight{} and \qivl, intervals including zero). %

\input{tables/metrics2}
\input{tables/views}

\paragraph{Randomness on the training side}\label{sup:seeds} \Cref{tab:seeds} repeats the main comparisons with four further fold seeds for the cross-validation that selects layer and ridge strength of the probe, the residual map and the supervised linear probe. Every interval of the detector against the supervised linear probe and against coordinate log-prob on natural errors stays on the same side of zero when it also covers the seeds, and the detector stays ahead in the discordant cross-model cases of the new items and of the held-out list. The sampling teacher rests on one sampled answer per item. Drawing that answer twice more on all six models, and refitting the probe, the supervised probe and the read-out each time, moves the differences on natural errors by a standard deviation of 0.001 to 0.010; every interval nested over images and the three fits stays on the same side of zero, on \qllava{} [$-$0.018, 0.016] against the supervised linear probe and [0.096, 0.164] against coordinate log-prob on the new items. With 20 draws of the 250 labelled items of \cref{tab:labelsvsdetector}, the supervised probe trails the detector fitted on the same items by 0.020 to 0.093 on average (standard deviation over draws 0.032 to 0.047), with an interval over images and draws below zero on four of the five models, \qfour{} excepted.

\input{tables/seeds}

\Cref{tab:trainres} redraws the training items themselves. The difference of the detector to the supervised linear probe on natural errors has an interval including zero under both computations in all 16 model--set cells (new items and held-out list for the six models, RefCOCO+ for three, the six sets of \qsmall), the difference to coordinate log-prob an interval above zero in all 16, and the residual added to the supervised linear probe raises its AUROC for wrong-object errors with a nested interval above zero in 15 of the 16 cells and a combined one in 11. In the discordant cross-model cases the detector's own share stays above one half under both computations, and its difference to the supervised linear probe is positive in nine of the ten refits on the new items and in all ten on the held-out list; its nested interval covers zero on the new items and touches it on the held-out list, while the combined one excludes zero on the new items and includes it on the held-out list. On wrong-object errors averaged over the six models, the detector minus the supervised linear probe is positive in all 20 refits, with a nested interval of [0.006, 0.082] on the new items and [0.005, 0.076] on the held-out list and a combined one of [$-$0.013, 0.065] and [$-$0.000, 0.070]; with the sampling teacher alone the nested intervals are [$-$0.003, 0.064] and [$-$0.000, 0.059].

\input{tables/trainres}

\section{Thresholds with a risk guarantee, procedure and cost}
\label{sup:ltt}

\paragraph{The abstention rule}
\label{sup:ltt:rule}

A deployed detector needs a threshold, which turns the score into selective prediction~\cite{geifman2017selective}. Given $m$ labelled calibration items, we choose it by fixed-sequence testing in the Learn-then-Test framework~\cite{angelopoulos2021ltt}. The candidate thresholds are fixed before calibration as 41 quantiles of the detector score on the unlabelled training items, from keeping 20\% to keeping all items in steps of 2\%, ordered from the most conservative to the most permissive. For each candidate we test the null hypothesis that the error rate among kept items exceeds a target $\alpha$ with an exact binomial test on the calibration items, move to the next candidate while the null is rejected at level $\delta$, and stop at the first non-rejection. Because the hypotheses are tested in a fixed order, no multiplicity correction is needed, and with probability at least $1-\delta$ over the calibration sample the error rate among kept items in the population is at most $\alpha$. Unlike conformal risk control~\cite{angelopoulos2024crc}, which bounds an expected risk, this bounds the risk with high probability. The detector itself remains label-free; only the threshold uses the $m$ labels. %

\paragraph{Simulation}
\label{sup:ltt:simulation}

In a simulation, we apply the fixed-sequence threshold of \cref{sup:ltt:rule} with $\delta=0.1$ to 1,000 random splits by image into calibration and evaluation items. On the six sets of \qsmall{}, with 400 labelled calibration items and a target error rate of 10\%, a threshold is certified in 98\% of the splits and 81\% of the items are kept on average; the error rate of the selected threshold on the whole set exceeds the target in 2.1\% of the splits, within the guaranteed $\delta$. Across the test sets of \cref{tab:ltt} this share stays between 1.2\% and 7.1\%. The guarantee needs exchangeable calibration and deployment items: calibrated on RefCOCO and used on Objects365 with an 8\% target, the kept error rate exceeds the target in 68\% and 78\% of the splits on the Qwen2.5-VL models, against 2\% on \qfour. Without a guarantee, a threshold set on half of the images for 95\% correct adopted answers lets the detector adopt 0.663 to 0.847 of the other half, coordinate log-prob 0.078 to 0.363 and the supervised probe about as many as the detector, except 0.127 fewer on the new items of \qlarge; the adopted answers are then 0.946 to 0.949 correct. %

\section{Weight sensitivity, threshold simulation and training-set size}
\label{sup:weights}

\paragraph{Ablations}\label{sup:ablation} \Cref{tab:ablation} reports each variant under two scorings. The probe carries the overall ranking and the residual adds the wrong-object errors: with half weight it raises the AUROC for these errors over the probe alone by 0.049 to 0.063. Kept as fixed for \qlarge, \qeight{} and \qivl, the weight 0.5 gives the highest point estimate for all errors among the five weights tried on the new items; for wrong-object errors alone weight 1 is higher, at a cost on errors of extent. The read-out needs no separate training set: fitted out of fold on the deployment items alone, it changes the detector by $-$0.010 to 0.002 for all errors, and lowers it for wrong-object errors on the new items of \qsmall{} ($-$0.016). For wrong-object errors on the replication list the slope of the zROC is close to one on average, 0.95 for the residual, 1.01 for the probe and 0.97 for the detector, with single models from 0.76 to 1.33. Four further fold seeds and two redraws of the sampled teacher answer keep every interval of the detector against the supervised and coordinate log-prob baselines on its side of zero. With the training items redrawn ten times, the interval to the supervised linear probe includes zero in all 16 model--set cells and that to coordinate log-prob stays above zero (\cref{sup:seeds}). %

\input{tables/ablation}

\paragraph{The residual on overlap errors} After hidden-state probes with the residual were seen to detect partial-overlap and wrong-extent errors slightly worse on the replication list (\cref{sec:plug:add}), we checked whether a binormal model of the probe and the residual accounts for it. Only the new items and the held-out list were used. Within each error class the standardized probe and residual are correlated: on overlap errors 0.344 and 0.352 on average over the six models (0.25 to 0.59 for single models, highest on \qllava{}). A binormal model with this correlation and the measured AUROCs reproduces the increment of each of the fifteen scores when the residual is added: rank correlation 0.993 and 0.996 between predicted and measured increments, mean absolute difference 0.0025 and 0.0033. For the detector it is off by at most 0.0085 (partial-overlap errors of the held-out list), and assuming equal variances changes its prediction by at most 0.0016. This is consistent with the residual carrying little beyond the probe on these errors: without the part predicted by the probe, the residual detects overlap errors with an AUROC of 0.520 and 0.526, against 0.761 and 0.726 for wrong-object errors (\cref{tab:overlapbinormal}). Across the fifteen scores, the stronger a score is on overlap errors, the less it gains (rank correlation $-$0.964 and $-$0.971), in part by construction. Post hoc, the same model gives an AUROC-optimal weight for all errors of 0.50 on the new items and 0.48 on the held-out list, close to the fixed 0.5. %

\input{tables/overlapbinormal}

With labels, the AUROC-optimal weight of the residual for two normal scores of correlation $\rho$ is $w=(d_r-\rho d_p)/(d_p-\rho d_r)$, with $d=\sqrt2\,\Phi^{-1}(\mathrm{AUROC})$ of the probe ($p$) and the residual ($r$), expressed in the units of the detector; computed for each model, on the new items it is 0.43, 0.35, 0.60, 0.57 and 0.34 for the five models (mean 0.46; the post hoc value 0.50 above comes from the mean AUROCs and correlations of the six models), and the fixed 0.5 differs from it by at most 0.0010 AUROC and from the best weight on a grid of step 0.05 by at most 0.0006. This describes the weight; it cannot set it, since the formula needs labels. Standardizing the two parts with the mean and variance of the deployment items themselves, which needs no labels, instead of those of the fitting items changes the AUROC of the detector by at most 0.0016 on the new items and the held-out list and by at most 0.0064 on Objects365 and the remote-sensing images (16 model--set cells), equivalent to residual weights of 0.42 to 0.71. %

On wrong-object errors the detector leads the sampling-teacher probe by 0.037 on the six \qsmall{} sets and by 0.026 on these three sets pooled. On \qlarge{}, which was not used to choose the weight, $w=0.5$ gives 0.848 against 0.833 for the probe alone and 0.835 for equal weight. On wrong-object errors the lead of the detector over the sampling-teacher probe is 0.009 on the new items of \qlarge{} and 0.037 and 0.025 on RefCOCO+, 0.026 pooled by inverse variance (correlations between the per-set differences at most 0.13). %

\begin{figure}[!htbp]
\centering
\includegraphics[width=0.8\textwidth]{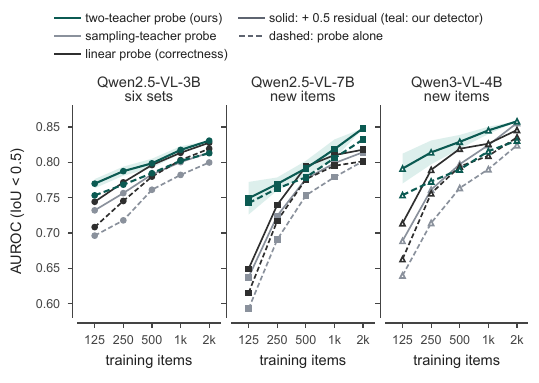}
\caption{AUROC against the number of unlabelled training items: \qsmall{} (six sets), \qlarge{} and \qfour{} (new items); mean over five random subsets, everything refitted on each; band: $\pm1$ SD of the detector. Teal: our probe; grey: sampling-teacher probe; black: linear probe on correctness. Solid: with half the residual; dashed: probe alone.}
\label{fig:curve}
\end{figure}

\Cref{tab:weights} shows the AUROC of every test set as a function of the residual weight, \cref{tab:ltt} the simulation of the fixed-sequence threshold of \cref{sup:ltt:simulation}, and \cref{fig:curve} the AUROC against the number of training items.

The AUROC still grows at 2,000 unlabelled training items (\cref{fig:curve}; matched sizes up to 10,000 in \cref{sup:tenk}; probe, read-out map and standardization refitted on each random subset; the points at 2,000 items are those of \cref{tab:detect}). With 250 items it reaches 0.788, 0.769 and 0.814 on \qsmall{}, \qlarge{} and \qfour{}, against 0.831, 0.848 and 0.858 with all 2{,}000. A supervised probe with the residual, trained on the gold boxes of the same items, is lower at small sizes, by 0.016 on \qsmall{} at 250 items and by 0.078 to 0.100 on \qlarge{} and \qfour{} at 125, and within 0.004 of it on \qsmall{} from 500 items on. %

\paragraph{Up to 10,000 training items}\label{sup:tenk} For \qsmall{} and \qlarge{} the training items were extended with referred objects from the RefCOCO train split, to 10,000 answered items (9,997 for \qlarge{}). SPRED was fitted on the first 2,000, the first 5,000 and all of them without their labels, and the supervised linear probe on correctness was trained on the same items with their labels, its feature, layer and penalty searched at each size (\cref{tab:tenk}). At the largest size SPRED minus the probe is 0.013 [$-$0.011, 0.037] on the replication list, with a one-sided lower bound of $-$0.007, and 0.001 [$-$0.025, 0.028] on the held-out list. SPRED at the largest size minus SPRED at 2,000 items is 0.025 [0.004, 0.045] on the replication list (0.020 for \qsmall{}, whose AUROC is the same at 5,000 and 10,000 items, and 0.029 for \qlarge{}) and 0.021 [$-$0.004, 0.045] on the held-out list. The gain comes with the number of items in one fit rather than with which items are added: fitted on all further items at once rather than on four blocks of 2,000 of them, SPRED is higher by 0.021 [0.006, 0.036], 0.013 [0.005, 0.022] and 0.008 [0.001, 0.015] on the replication list, the held-out list and the new items. A three-layer perceptron on correctness, its settings chosen among 24 by five-fold cross-validation on the labelled training items, stays below the linear probe at 10,000 labels: 0.787 and 0.748 against 0.822 and 0.778 on the replication list ($-$0.032 on average), $-$0.020 on the held-out list and $-$0.021 on the new items. Among this perceptron, a linear probe on the state before generation and a linear probe trained on $1-\mathrm{IoU}$, the same cross-validation picks the last for both models at every size; SPRED minus it is 0.012 [$-$0.013, 0.037] on the replication list, with a one-sided lower bound of $-$0.009, and 0.014 on the held-out list. %

As reported for text question answering~\cite{phillips2026entropy}, where combined uncertainty features also improve on single ones~\cite{xiong2024uqllm}, a label-free score adds to a supervised probe, and on the replication list it does so at 10,000 labels (\cref{tab:tenkadd}). Added after standardization to the probe on correctness, SPRED fitted on the same items raises the AUROC for all errors by 0.027 [0.012, 0.042] on the replication list (0.010 for \qsmall{}, 0.044 for \qlarge{}) and for wrong-object errors by 0.036 [0.017, 0.056], with the same sign on the held-out list and the new items. With the configuration of the probe searched again at 10,000 items, the gains for all errors are 0.022, 0.018 and 0.015 on the replication list, the held-out list and the new items, and those for wrong-object errors 0.044, 0.029 and 0.035. With the probe trained on $1-\mathrm{IoU}$ the gain for all errors is 0.027 [0.014, 0.040], and SPRED fitted on the original 2,000 unlabelled items adds 0.017 [0.003, 0.030] to the probe trained on 10,000 labels. In a post hoc comparison on wrong-object errors, a second supervised probe trained on the same labels adds $-$0.002, 0.004 and 0.006 on the three lists, and adding SPRED minus adding it is 0.039 [0.023, 0.054], 0.022 [0.005, 0.039] and 0.028 [0.015, 0.042]. With both probes searched again at 10,000 items the difference is 0.031 [0.003, 0.059], 0.035 [0.006, 0.064] and 0.026 [0.005, 0.047]; these intervals are wider because the second probe is also searched again in every refit, and for \qsmall{} both versions select the same configurations on the full training set, so the change in the estimates comes from \qlarge{}. Also post hoc, on wrong-object errors SPRED minus the probe on correctness is 0.059 [0.006, 0.111] on the replication list and 0.033 on the held-out list. %

\input{tables/tenk}
\input{tables/tenkadd}

\paragraph{BICR with 10,000 labels} For \qsmall{} and \qlarge{}, our reimplementation of BICR was also trained on the labels of the same 10,000 training items (9,997 for \qlarge{}), reading the last layer at the last prompt token as before, and SPRED was fitted on these items without their labels. On all errors of the replication list SPRED minus BICR is 0.040 [0.010, 0.071] on average over the two models (0.046 for \qsmall{}, 0.034 for \qlarge{}; AUROC 0.812 and 0.814 against 0.766 and 0.780), with a one-sided lower bound of 0.015; \cref{tab:bicrtenk} adds the held-out list, the new items and wrong-object errors.

\input{tables/bicrtenk}

\Cref{tab:sens} moves each setting of the detector on the new items of all five models, one at a time, from its label-free choice. Among the weights tried from 0 to 2, 0.5 gives the highest point estimate or one tied with it on every model; $w=0$ (the probe alone) gives 0.809 to 0.850 against 0.829 to 0.867. The layers and penalties chosen by out-of-fold criteria on the unlabelled training items are within 0.013 of the best value of each sweep; the probe penalty matters most, a penalty ten times off changing the AUROC by as much as $-$0.041 on \qlarge, which is why it is chosen, without labels, on the training items. %

\input{tables/weights}
\input{tables/sens}

The fixed-sequence threshold (\cref{sup:ltt:rule}) is insensitive to the step of its candidate grid: with 400 calibration items, $\delta=0.1$ and a target of 10\%, steps of 1, 2 and 5\% keep 79\%, 79\% and 77\% of the new items of \qsmall{} (\qlarge: 82\%, 82\% and 81\%; \qfour: 89\%, 89\% and 87\%), and starting the grid at keeping 40\% instead of 20\% keeps 79\%, 83\% and 89\%. Starting at 10\% is worse: because the test stops at its first non-rejection and keeping 10\% leaves about 40 calibration items, one error among them stops it, and at a target of 8\% a threshold is found in only 53\%, 21\% and 3\% of the splits. With $\delta=0.05$, 0.1 and 0.2 the detector keeps 54\%, 67\% and 74\% of the items of \qsmall{} at that target. %

\input{tables/ltt}

In the threshold simulation on the six sets of \qsmall{}, with 200 calibration items and a target of 10\%, the detector keeps 55\% of the items, the supervised correctness probe 51\%, the sampling-teacher probe 37\% and coordinate log-prob 7\%. %

\section{Natural errors on the new items and RefCOCO+}
\label{sup:natural}

\Cref{tab:detect} gives the detector and every single-pass score on the new items and RefCOCO+. With one pass and no labels, the detector reaches an AUROC of 0.831 on the six sets of \qsmall{} and 0.829 on its new items, 0.848 on the new items of \qlarge{}, 0.858 on those of \qfour{}, 0.837 on those of \qeight{}, and 0.830 and 0.822 on RefCOCO+. Handing the fifth of the answers with the highest detector scores on to review lowers the error rate among the rest from 0.110--0.146 to 0.050--0.079 on the new items of the five Qwen and InternVL models, against 0.048--0.086 when the supervised linear probe chooses them and 0.084--0.104 when coordinate log-prob does (\cref{tab:risk80,tab:referral}). Built with the sampling teacher alone, the detector differs from the supervised linear probe by $-$0.037 to 0.020, with no detected difference except on the held-out list of \qeight{}, where it is behind. On \qllava{} the detector (0.790) shows no detected difference from the supervised linear probe (0.001 [$-$0.016, 0.018]) and is ahead of coordinate log-prob by 0.132. On \qivl{}, whose language model is not from the Qwen family, it is ahead of coordinate log-prob by +0.259 over the new items and RefCOCO+, and its one-sided bound against the supervised correctness probe is $-$0.030, at the margin of 0.03.

\input{tables/detect}

\section{The residual added to other scores}
\label{sup:plugin}

A cell is one model with one score on one set; a score that is not defined for a model is not counted, and a score that coincides with another for a model (the pre-generation and the correctness probe of \qlarge) is counted once where distinct cells are counted. \Cref{tab:residadddiff} adds the residual to five probes on the RefCOCO-family sets, \cref{tab:residheld} to six base scores on the held-out list, model by model, \cref{tab:bytypenewitems} repeats \cref{tab:bytypefull} on the new items, \cref{tab:addsig} replaces the residual by other label-free signals, and \cref{tab:plugci} gives, for wrong-object errors, the stored gain of each of the fifteen scores with its combined interval, and the refit mean with its nested interval. At the level of single cells, counting the one fit shared by two scores of \qlarge{} once, the stored gain for wrong-object errors is positive in all 85 distinct cells on both lists; the combined interval is above zero in 79 cells on the new items and 59 on the held-out list, the nested interval in 85 and 76, and the combined interval is below zero in none. All scores are refitted here from the stored features with the stored selection, and all coincide with the scores of \cref{tab:detect}. %

Before the plug-in analyses were extended, the gain on six base scores was examined on its own and gave the same picture: positive in all 35 distinct cells of the held-out list, with intervals above zero on the new items, RefCOCO+ and the development items. The extension to fifteen scores and the reading of its intervals were fixed after the held-out list had been scored for the detector.

\begin{figure}[!htbp]
\centering
\includegraphics[width=\textwidth]{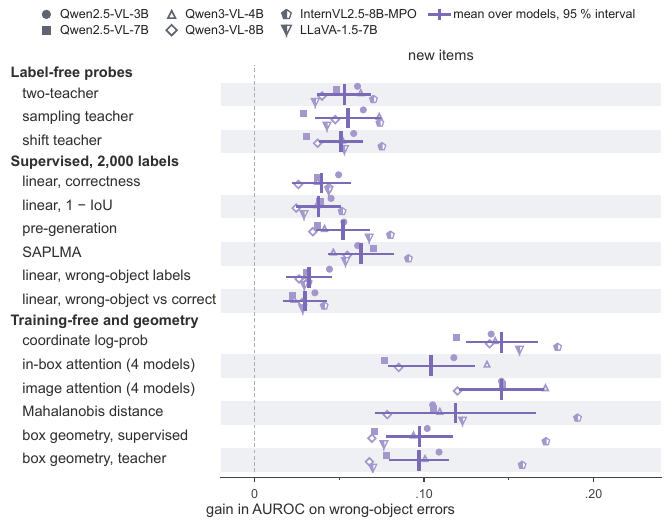}
\caption{Gain in AUROC on wrong-object errors of the new items when the residual is added with weight 0.5 to each of fifteen single-pass scores. Markers: models (attention scores: four Qwen models). Tick and line: mean over the models with its combined interval over images and ten training refits. Dashed: no gain.}
\label{fig:pluginnewitems}
\end{figure}

\input{tables/bytypefull}
\input{tables/bytypeho}
\input{tables/gaindiff}

\begin{figure}[!htbp]
\centering
\includegraphics[width=\textwidth]{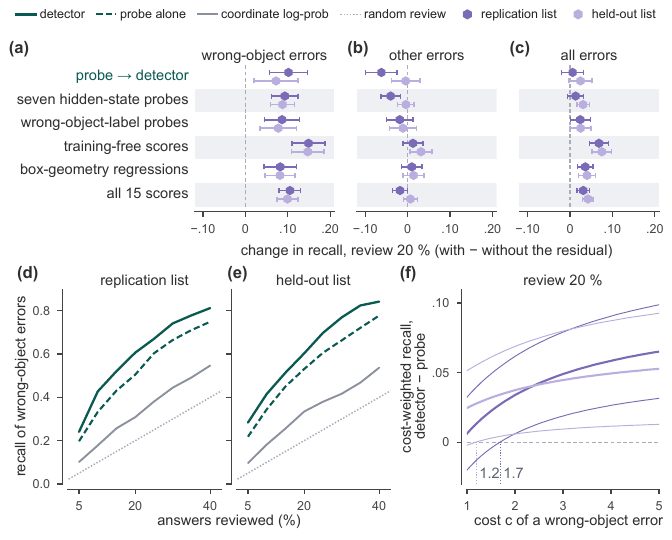}
\caption{Review budget and unequal error costs. (a)--(c) Reviewing the top-scoring 20\%: change in recall from adding the residual, for wrong-object, other and all errors; rows: score groups (means over their scores). (d), (e) Recall of wrong-object errors against the share of answers reviewed. (f) Reviewing 20\%, a wrong-object error costing $c$ and any other error 1: detector minus probe in cost-weighted recall; thin lines: interval; number: smallest $c$ with the interval above 0. Means over models with combined 95\% intervals; dark: replication list, light: held-out list. Dashed: no change.}
\label{fig:reviewcost}
\end{figure}
\input{tables/bytypenewitems}
\input{tables/plugci}
\input{tables/addsig}
\input{tables/equalinfo}
The gain on wrong-object errors also holds for our reimplementation of BICR and for ridge probes on the states of both positions trained with 2,000 labels, and a logistic model given the read-out box, the written box and their IoU is no better than the residual alone (\cref{tab:spredrepfull,tab:equalinfo}).

Boxes on another object tend to be smaller, and the area of the written box alone separates them from the other errors with an AUROC of 0.670 on the replication list and 0.636 on the held-out list (\cref{tab:errtype}). In object detection, errors are split into types such as localization and classification errors offline, against the ground truth~\cite{bolya2020tide}; errors of VLMs in knowledge-based question answering have been attributed to stages with supervised probes on visual and prompt states~\cite{ax2607.04683}; here the task is grounding, the residual needs no labels and works on each answer at test time, and the two kinds are boxes on another object and boxes that are off.

Post hoc and without correction for multiple comparisons, the residual alone is also above the classifier on box geometry and above box area, and a supervised probe trained on wrong-object labels reaches 0.654 (\cref{tab:errtype}).

\input{tables/errtype}

In terms of review, with the fifth of the answers returned, adding the residual raises the share of wrong-object errors caught by 0.116 [0.090, 0.142] on the new items and 0.100 [0.075, 0.124] on the held-out list (stored fit, combined interval; mean over the fifteen scores and six models), for the supervised linear probe from 0.627 to 0.738 and from 0.560 to 0.640; the share of all errors caught rises on average by 0.042 and 0.043.

\section{Further analyses}
\label{sup:extra}

\paragraph{Sampling baselines, extra passes, single teachers and imperfect labels} \Cref{tab:samplebase} compares the single-pass detector with sampling baselines on \qsmall{} and \qfour{} (\cref{sec:plug:add}); \cref{tab:extraadd} adds the residual to checks that need extra passes; \cref{tab:teachers} replaces the two consistency teachers by single ones (\cref{sec:method}); and \cref{tab:offidealci} gives the intervals behind \cref{tab:offideal}, with the SAPLMA probe also trained without one error type. When the detector is refitted without the same items, \qllava{} gives 0.068 [0.039, 0.098] without wrong-object labels and 0.012 [$-$0.008, 0.031] without extent labels. When the model writes a point instead of a box, an answer is right when the point lies in the annotated box, so only the object counts. With the read-out fitted to the model's own points and the distance between the point read out and the point written as residual, adding it to the two-teacher probe raises the detection of all errors by 0.031 on average over \qsmall{}, \qlarge{} and \qfour{} on the new items and RefCOCO+; the replication and held-out lists are in \cref{tab:pointlists}.

\paragraph{Few labels, a missing error type and other images} \Cref{tab:offideal} collects the settings of \cref{sec:plug:offideal}.

\input{tables/offideal}

\begin{sloppypar}
\paragraph{Further baselines}\label{sup:morebase} Single-pass scores that need no training, Chain-of-Embedding~\cite{wang2025coe} and LLM-Check on hidden states~\cite{sriramanan2024llmcheck}, are below the detector in all 29 comparisons with layers fixed in advance, on the five Qwen and InternVL models, after Holm correction. Scored against all other answers instead of the correct ones, wrong-object errors on the replication list give an AUROC of 0.805 for the detector, 0.758 for its probe and 0.798 for the residual, averaged over the six models, a lead of 0.047 [0.035, 0.060] of the detector over its probe. On natural errors the detector is also ahead of the surface-feature baselines and, in AUROC and FPR95, of the digit-weighted coordinate uncertainty, and it barely changes on items whose annotation Ref-L4 confirmed~\cite{chen2024refl4}. On natural errors, supervised baselines on surface features (expression length, image size, geometry of the written box) reach only 0.558 to 0.584 (\cref{tab:surface}); and on the new items of \qlarge{} the detector is ahead of all ten distinct single-pass baselines. On the new items of all five models it is also ahead of coordinate log-prob in AUPRC (0.388 to 0.522, chance level 0.110 to 0.146) and in FPR95; at 95\% recall of errors it still flags 0.491 to 0.591 of the correct answers (\cref{tab:metrics2}). Built with the sampling teacher alone, so that no part of it sees a shifted view, the detector stays ahead of coordinate log-prob in all 15 new, RefCOCO+ and held-out cells. Against the digit-weighted coordinate uncertainty (\qsmall, \qlarge{} and \qfour) it is ahead in AUROC, by 0.046 to 0.172, and in FPR95. On the items whose RefCOCO-family annotation was re-checked for Ref-L4~\cite{chen2024refl4}, 32.4\% of the errors of the five models fall on annotations the re-check found faulty; scored only on items whose annotation it confirmed, the detector and its difference to the supervised linear probe barely change (pooled 0.011 on all re-checked items, 0.015 on the confirmed ones). Against sampling baselines on \qsmall{} and \qfour, the single-pass detector is above EigenScore over ten sampled answers~\cite{chen2024inside} and self-consistency over one or two, and shows no detected difference from self-consistency over five; with one extra pass at a quarter of the pixel budget it is above the latter, and the detector plus five-answer self-consistency is above five-answer self-consistency on both models (\cref{tab:samplebase}). The residual also raises checks that need extra passes: added to a content-shifted or low-resolution second answer, to five sampled answers or to the box of another model, it raises their AUROC for wrong-object errors in point estimate in all 14 model--check cells, with an interval above zero in 13 (\cref{tab:extraadd}). With the residual weighted 0.25 or 1 instead of 0.5, the gain for wrong-object errors stays positive in all 14 cells, with an interval above zero in 14 and 13; for all errors it is positive in 14 and 13, with an interval above zero in 14 and 9. With the boundary of wrong-object errors at IoU $=0$, $<0.05$, $<0.1$ or $<0.2$, the gain of the residual over the probe alone is positive on all five models (0.019 to 0.090), with an interval above zero in 19 of 20 model--boundary cells. Added to the supervised linear probe, the detector raises it in point estimate on all five models (by 0.011 to 0.057; \cref{tab:detect}), and a reference that stacks all single-pass scores with labels of the test items, cross-fitted, gains on average only 0.014 over the detector. %
\end{sloppypar}

\input{tables/samplebase}
\input{tables/extraadd}
\input{tables/pointlists}
\input{tables/teachers}
\input{tables/offidealci}

\paragraph{A token-style read-out}\label{sup:tokenreadout} \Cref{tab:tokenreadout} compares the ridge read-out with a read-out of the kind a future lens would fit, softmax classifiers over binned coordinates trained on the model's own answers from the same state. Added to the two-teacher probe, both residuals raise the detection of wrong-object errors, and the difference between the two gains, averaged over the two models, has an interval that includes zero on both lists. Alone, the token-style residual is lower than the ridge residual by 0.051 [0.003, 0.099] on the replication list on average over \qsmall{} and \qlarge{} (on \qsmall{} alone the interval includes zero), and on the held-out list no difference is detected; it reads the model's own box slightly more closely, so the difference lies in what the disagreement detects, not in how well the box is read.

\input{tables/tokenreadout}

\paragraph{Attention heads as the reference}\label{sup:attnref} A localizer that needs no fitting can also be taken from attention heads, as in training-free grounding with a few heads~\cite{kang2025heads}. As the reference we use the mean attention centroid over the image of three heads at the last prompt token, and as the score its distance to the centre of the written box, added to the two-teacher probe with weight 0.5 and standardized on the training items. Only summary statistics of each head's attention were stored, among them its centroid, so this approximates the cited approach rather than reproducing its boxes. With the heads chosen on the unlabelled training items by attention mass and entropy, the residual gains more on wrong-object errors of the new items than this reference, by 0.085 [0.052, 0.118] on average over \qsmall{} and \qlarge{}, but these heads do not point at the model's own answers. Chosen instead, post hoc, as the three heads whose centroids lie nearest the model's own answers on the training items, the reference gains $-$0.003 against 0.055 for the residual, a difference of 0.058 [0.032, 0.085]. At the token before the first coordinate, where many heads already point near the answer, the difference is 0.045 [0.022, 0.068], and for \qlarge{}, \qfour{} and \qeight{} alone its interval includes zero. These comparisons concern a reference built from stored centroids; they do not show that attention maps locate the object worse than the fitted read-out.

\paragraph{Shared and stable errors}\label{sup:clean} Split by how many of the other four models also err on the item, the lead of the detector over the supervised linear probe lies in the errors of one model only (0.046 [0.011, 0.082]); on errors shared with others the interval includes zero. Of the errors in the detector's least suspicious half of the items, 74\%, 86\% and 78\% are shared with another model (\qsmall, \qlarge, \qfour). Errors whose box all seven views keep make up 0.21, 0.26 and 0.44 of the errors; consistency checks see them poorly, and on these items the detector still reaches 0.748, 0.790 and 0.870 (\cref{tab:stable}). The comparison with seven views was made on three models after their results were known, and on \qeight{}, run later, after the method was fixed; \cref{tab:stable} gives the share of stable errors, the AUROC of the seven views and the detector's lead on all items and on stable items. On the stable new items, adding the residual to the two-teacher probe raises the detection of wrong-object errors by 0.059 [0.026, 0.092] on average over the four models (7 to 18 such errors per model).

\input{tables/stable}

\begin{figure}[!htbp]
\centering
\includegraphics[width=\textwidth]{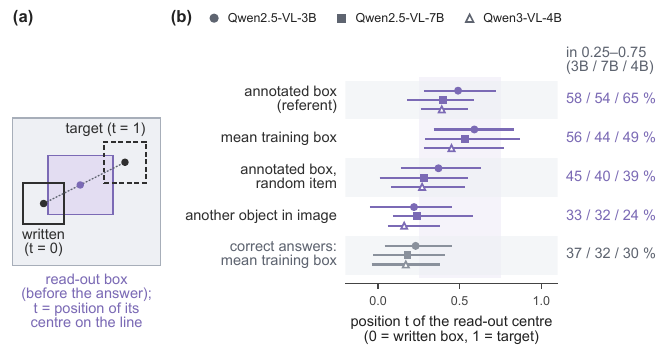}
\caption{Where the read-out lies for wrong-object errors (new items). (a) $t$: position of the read-out centre on the line from the written box ($t=0$) to a target box ($t=1$). (b) Median (marker) and interquartile range (bar) of $t$ towards four targets: the annotated box, the mean training box, the annotated box of a random other item and another annotated object of the same image; grey: correct answers towards the mean box; right: share of $t$ in 0.25--0.75.}
\label{fig:residual}
\end{figure}

\Cref{fig:residual} shows where the pre-generation read-out points for natural wrong-object errors.

\paragraph{Where the read-out points}\label{sup:pull} \emph{Towards the annotated object.} Over the six models, regressing the centre of the read-out minus that of the written box on the annotated centre minus the written one and on the mean training centre minus the written one, without intercept, gives a coefficient on the annotated centre of 0.210 to 0.382 on the new items and 0.164 to 0.286 on the held-out list, and near zero with the annotated centre of another wrong-object item in its place. The read-out of a wrong-object error is also larger than the written box (median area ratio 1.58 in \qsmall{}, against 1.09 for correct answers; \cref{fig:residual}). With the read-out of another item or the mean training box in place of its own, the residual loses most of its separation of wrong-object errors (\cref{tab:readctl,tab:readctlho,tab:readctlnewitems}).

\emph{On re-checked annotations.} Part of the annotations are themselves faulty. Ref-L4 re-checked the annotation of part of the new and RefCOCO+ items~\cite{chen2024refl4}, and on the items whose annotation it confirmed the displacement remains: 0.314 on the new items and 0.240 on RefCOCO+. On these items the residual still detects wrong-object errors, with a mean AUROC of 0.842 and 0.817, slightly below all re-checked items, and still raises the supervised linear probe, by 0.040 and 0.027. About a third of the wrong-object errors on re-checked items fall on annotations the re-check found faulty (0.315 and 0.381 of model--item pairs). Counted over distinct items the shares are 0.229 and 0.362, against 0.079 and 0.227 of all re-checked items, since several models often err on the same faulty item.

\emph{Other objects of the image.} By the MS-COCO instance annotations, the written box covers another annotated object in 81\% of the wrong-object errors of \qsmall{}, against 34\% of its errors of extent. On the 150 wrong-object errors whose image holds another object annotated for a different expression, that object receives a coefficient of $-$0.034 [$-$0.094, 0.019], against 0.303 for the annotated object, so the read-out does not drift towards just any annotated object.

\input{tables/readctl}
\input{tables/readctlho}
\input{tables/readctlnewitems}

\emph{Second answers.} Where the answer changes under the content shift, the new box moves with the read-out (\cref{tab:secondans}): on the new items and RefCOCO+ of the five models the slope is 0.942 to 1.251, near zero with the read-outs permuted, and it stays at 0.603 to 1.100 with the annotated centre added. The agreement is one of direction, not position: the read-out is closer to the second box than to the first in only 0.293 to 0.553 of these items. The slope is lower for \qllava{} and on Objects365. Every cell has a slope of at least 0.6 with a lower bound above 0.3 and a permuted slope below 0.1 in absolute value (values read off the new items of the four Qwen models; cells with at least 30 changed items) except \qllava{} on the held-out list and two of the three Objects365 cells.

\emph{Same-class objects.} On the items of the held-out list with a single further object of the target's class, the three observations of \cref{sec:measure} remain: the residual detects wrong-object errors with an AUROC of 0.859, the read-out is displaced towards the target by 0.243, and the second answer follows it with a slope of 0.918. The residual of correct answers grows with the number of same-class objects, from a mean standardized value of 0.06 to 0.26 with one to 0.22 to 0.42 with three or more (\cref{tab:sameclass}). Within strata of the number of objects of the target's class (post hoc; gain on wrong-object errors when the residual is added to the two-teacher probe, six models), the original strata, one, two and three or more objects including the target, leave the first empty, since no replication-list item has a single such object; the other two give 0.031 and 0.054, pooled 0.040 [0.025, 0.055] on the replication list and 0.045 [0.030, 0.060] on the held-out list. Defined after the empty stratum was seen, three strata of one, two and three or more other objects give 0.031, 0.052 and 0.055, pooled 0.037 [0.020, 0.053], on the replication list.

\input{tables/secondans}
\input{tables/sameclass}

Sampled second answers on the held-out list point the same way. With five answers sampled at temperature 1 on each item, we take the items whose sampled boxes disperse (mean $1-\mathrm{IoU}$ to the greedy box above 0.3) and regress the displacement of the mean sampled centre on that of the read-out as above: the slope is 0.429 [0.295, 0.573] for \qlarge{} (138 items) and 0.599 [0.497, 0.690] for \qllava{} (485 items), and $-$0.009 and 0.007 with the read-outs permuted. \qeight{} and \qivl{} rarely change their answer at this temperature (48 and 43 such items) and are not judged. Over all items of the four models, the residual and the dispersion of the sampled answers have a rank correlation of 0.397 to 0.562. %

\paragraph{The residual beyond box geometry} The residual also reflects the geometry of the written box: on the training items of \qsmall{}, the position and size of the written box explain $R^2=0.354$ of it. To remove this route, we fit the residual that box geometry predicts on the training items and split the test items into 12 bins of it, comparing errors and correct answers only within a bin. The gain of the detector over the probe on wrong-object errors stays at 0.043 to 0.057 on the new items and RefCOCO+ sets of the three models, against 0.049 to 0.063 without bins, and a trivial alternative that adds a small-box score to the probe reaches only 0.797, 0.805 and 0.822 overall on the new items, against 0.829, 0.848 and 0.858 for the detector (intervals in \cref{tab:otherdiff}). Correct answers also have a higher residual when the image holds more objects of the target's class (\cref{sec:measure:pull}). These observations are consistent with a pre-generation state that still carries information about the referent when the model writes a box on another object; they do not show whether that information takes the form of a competing box.

\paragraph{Other image domains and target size}\label{sup:domains} Trained without labels on Objects365 images~\cite{chen2024refl4}, the detector reaches 0.692 to 0.762, and the detector fitted on RefCOCO and applied unchanged does as well (differences $-$0.010 to +0.016); against the supervised linear probe transferred from RefCOCO, the detector fitted on RefCOCO is ahead on \qsmall{} with no detected difference on the other two models, and it is above the transferred SAPLMA probe on all three (0.066 to 0.092). With the maps fitted on RefCOCO and applied unchanged, the difference in wrong-object gain between written boxes at least $2\sigma$ wide and smaller ones is $-$0.007 on Objects365. On VRSBench remote-sensing images~\cite{li2024vrsbench}, fitted without labels on 1{,}000 and 998 training items, it reaches 0.752 and 0.784 and is ahead of coordinate log-prob by 0.083 and 0.127; against the transferred supervised probe it is ahead on \qfour{} with no detected difference on \qsmall{} (\cref{tab:o365,tab:offideal}). Measured against the read-out error and with the read-out fitted on Objects365, the gain for wrong-object errors averaged over the seven hidden-state probes and the three models rises with the annotated target from $-$0.008 below $\sigma$ and $-$0.003 at one to two $\sigma$ to 0.016 at two to three and 0.086 at three or more; by the written box it is 0.001, 0.014, 0.040 and 0.074; 0.543 and 0.590 of the remote-sensing boxes are narrower than $2\sigma$, and there the residual changes the AUROC by $-$0.007 and $-$0.002. With the read-out fitted on the domain, for annotated targets narrower than $2\sigma$ the detector is below its probe on all errors: the difference is $-$0.032 [$-$0.047, $-$0.016] on Objects365 (three models) and $-$0.032 [$-$0.043, $-$0.021] on remote-sensing images (two models). On Objects365 the weight 0.5 is below the best weight on the test items by 0.011, 0.001 and 0.005 (\qsmall, \qlarge, \qfour). %

\input{tables/o365}

Some ``errors'' are annotation problems, and the evaluation can be repeated on confirmed items. The Ref-L4 review~\cite{chen2024refl4} covers 794 of the new items and marks 63 as problematic. On the 731 items it confirms, the detector reaches 0.835 on \qsmall{} and 0.856 on \qlarge; on the flagged items it drops to 0.563 and 0.658, because many of their ``errors'' are not model errors. Among the correct answers in the detector's most suspicious 20\%, the review flags 16\% (\qsmall) and 17\% (\qlarge) of those it covers, against 2\% and 2\% in the least suspicious half. %

\paragraph{Optional second passes}\label{sup:second} Where more passes are affordable, they can be used alone or added to the detector. Used alone, consistency checks need several passes to match one pass of the detector (\cref{tab:views}; new items): the disagreement over seven views, eight passes in all, is 0.837, 0.834 and 0.780 on \qsmall, \qlarge{} and \qfour{}, against 0.829, 0.848 and 0.858 for the detector, and the mean disagreement over randomly drawn views reaches the detector at about seven views on \qsmall{} and eleven on \qlarge{}, whereas on \qfour{} seven views give 0.780. In the Qwen2.5-VL models a left shift of the content by two tokens is the strongest single check: on the new items it reaches 0.765 in \qsmall{} and 0.787 in \qlarge{} (\cref{tab:detect}), and in the comparison of perturbations it is ahead of the best other one of equal cost by 0.063 and 0.057. On top of the full detector (\qlarge{}, new items; intervals in \cref{tab:otherdiff}), asking \qfour{} for its box adds 0.028. One more pass of the same model adds little (a content shift 0.012, a resampled answer $-$0.014), so the detector already carries most of it. %

Asking a model of another family for its box is an existing second check~\cite{ax2609.37488}; with the detector as the gate, asked only on the 36.2\% to 40.9\% of the items above the 70\% training quantile (1.36 to 1.41 times the cost), it raises the AUROC by 0.018 to 0.030 on all five models, above the supervised probe on three. %

A frozen open-vocabulary detector used zero-shot as a second localizer, Grounding DINO in its tiny version~\cite{liu2024groundingdino}, is compared with the residual in \cref{tab:external}.

\input{tables/external}

Adding the standardized shift score to the single-teacher probe gives 0.076 and 0.066 over the shift check alone.

\section{Error detection with item difficulty held fixed}
\label{sup:ctl}

Natural errors are confounded with item difficulty: a score can rank errors well by recording only that an item is hard. This section compares the scores at fixed difficulty across models and gives a signal-detection prediction for that comparison (\cref{eq:twocomp}).

\subsection{What the controlled comparison measures}
\label{sec:ctl:estimand}

Comparing a model's own probe with probes on other models comes from text tasks~\cite{ashuach2026masked,ax2509.24988}, and so does separating item difficulty from the failures of a run~\cite{martinezplumed2019irt,ax2509.15339,ax2610.05572,sharma2026probeeval}. The comparison across models is within an item: on the 1{,}146 new items answered by all five models, each of the 1{,}224 cases in which one model of a pair is right and the other wrong counts as won when the erring model's score, as a percentile among its own model's items, is the higher one (ties count half); a score that records only the difficulty the two models share reaches 0.5. A higher value means better ranking among cases of matched difficulty, not that more errors are detected over all items.

For the comparison across models, the equal-variance binormal model of signal detection with a component shared by the models~\cite{green1966sdt,dorfman1992mrmc} predicts the share of discordant cases a score wins, with no fitted parameter. With $d_{\mathrm{s}}$ and $d_{\mathrm{c}}$ the AUROCs for a model's errors of its own score and of another model's, each as $\sqrt2\,\Phi^{-1}(\cdot)$, $\kappa$ the excess error rate of the other model where the first errs, and $\rho$ the correlation of the two scores where both are right, the predicted share is
\begin{equation}
\Phi\Bigl(\frac{d_{\mathrm{s}}-d_{\mathrm{c}}}{(1-\kappa)\sqrt{2(1-\rho)}}\Bigr).
\label{eq:twocomp}
\end{equation}
On the held-out list it gives the measured shares of the twelve scores within 0.013 on average (\cref{sup:heldout}).

\subsection{Across models}
\label{sec:ctl:models}

The detector picks the model that erred in 0.634 of the cases (0.619 when built with the sampling teacher alone), the supervised linear probe trained on the correctness of 2{,}000 items in 0.565 and coordinate log-prob in 0.520; the detector's share is higher than those of all four supervised probes, by 0.054 to 0.084, and model by model higher than that of the supervised linear probe on all five (by 0.044 to 0.114). Counting only the pairs in which the erring model placed its box on another object, the residual picks that model in 0.637 [0.551, 0.722] of 406 pairs and the detector in 0.640 [0.559, 0.721] (combined intervals over images and ten refits). On the held-out list the detector picks the model that erred in 0.647 of the 1{,}218 cases and the supervised linear probe in 0.601. With the training items redrawn (\cref{sec:setup}), the detector's share stays above one half (0.602 and 0.633 on average) and its difference to the supervised linear probe is positive in 19 of 20 refits, but its interval then covers zero on the new items and touches it on the held-out list.

\subsection{Item difficulty and supervised probes}
\label{sec:ctl:difficulty}

With the mean residual of the other five models on the same item in place of its own, a score gains about as much on wrong-object errors (\cref{tab:ownother,tab:ownothermodel}): over the seven hidden-state probes the ratio of the two gains is 1.01 [0.77, 1.30] on the replication list and 0.89 [0.67, 1.10] on the held-out list, the own gain minus the others' gain has a combined interval above zero for 0 of the fifteen scores on the replication list and 1 on the held-out list, and the others' residual raises all 15 scores on the replication list with an interval above zero (post hoc). The residual of a single other model, added to the two-teacher probe, gains 0.032 on the replication list against 0.042 with the own one; post hoc, the own gain minus that of a single other model is 0.009 [$-$0.001, 0.019] on the replication list, where no difference is detected, and 0.018 [0.008, 0.028] on the held-out list.

\input{tables/ownother}
\input{tables/ownothermodel}

Holding difficulty fixed changes the order of the scores: the supervised linear probe, second of twelve in mean AUROC on the natural errors (0.834), is eighth across models, and the residual, ninth there (0.747), is third. Another model's score finds a model's errors nearly as well as its own (detector 0.814 against 0.849), as on text tasks, where probes on another model's representations predict a model's correctness about as well as probes on its own~\cite{ashuach2026masked}. A probe distilled from the other models' label-free scores, which never sees this model's correctness, shows no detected difference from the supervised linear probe in four of the five models but is below both in the discordant cases (0.549, against 0.565 and 0.634; \cref{tab:dprobe}). At 0.634, the detector also ranks errors partly by the difficulty that the models share.

\begin{figure}[!htbp]
\centering
\includegraphics[width=0.8\textwidth]{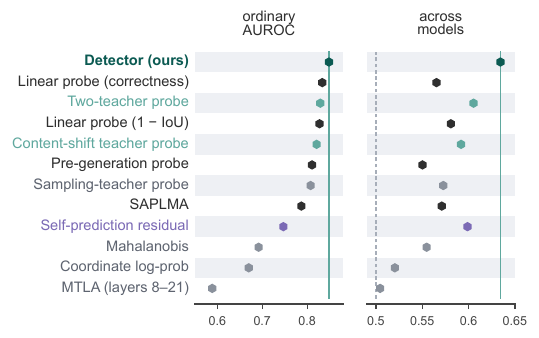}
\caption{Error detection with item difficulty held fixed. Rows: scores, sorted by AUROC on natural errors (mean over five models; MTLA, layers 8--21, four). Second column: share of discordant cross-model cases in which the wrong answer scores higher; dashed: 0.5. Definitions in \cref{sup:ctl}.}
\label{fig:controlled}
\end{figure}

\paragraph{Details of the comparison}
\Cref{tab:controlled} gives the comparison with its intervals. A share of 0.5 is what a score that reflects only item difficulty would give. With other error thresholds, without percentiles, without any one model or split by error type, the cross-model difference of the detector to the supervised linear probe stays positive, by 0.036 to 0.086 (\cref{tab:pairsrob}). With scores standardized by the training items instead of percentile ranks, the detector reaches 0.634 and the supervised linear probe 0.570; on the 1{,}028 cases between models of different families the detector leads that probe by 0.069. On Objects365, with the detector and the supervised probes fitted on RefCOCO and applied unchanged, the detector picks the model that erred in 0.617 of the 510 cases and the supervised linear probe in 0.540, a difference of 0.076 [0.021, 0.133]. MTLA in \cref{fig:controlled} is the version with layers 8 to 21 (\cref{sup:repro}).

\input{tables/controlled}
\input{tables/pairsrob}
\input{tables/dprobe}

\section{Held-out evaluation}
\label{sup:heldout}

The held-out list was drawn with a fixed seed from the test split of RefCOCOg (UMD partition) after removing the images of the training items, the new items and the RefCOCO+ items: 1{,}200 items on 983 images. 36 of these images (47 items) also occur, with other expressions, in three early development lists (RefCOCO testA and testB, RefCOCOg validation) that were examined while the method and its weight were chosen. Without these items, the detector minus the supervised linear probe for all errors changes from 0.010 to 0.012, the detector minus coordinate log-prob from 0.162 to 0.159, and the mean gain of the fifteen scores for wrong-object errors when the residual is added from 0.065 to 0.066. Each model answered it with its own code, prompt and shifted view as for the new items; the detectors, probes, read-out maps and reference scores were the ones fitted earlier on the training items and were applied unchanged. \Cref{tab:heldout,tab:heldoutpairs} give the results. Model by model the detector is above coordinate log-prob by 0.113 to 0.233, all intervals above zero, and its difference to the supervised linear probe runs from $-$0.014 to 0.026, all intervals including zero. Across the five models the detector picks the model that erred in 0.647 [0.607, 0.688] of the 1{,}218 discordant cases, the supervised linear probe in 0.601 [0.558, 0.645] and coordinate log-prob in 0.564, a difference to the supervised linear probe of 0.046 [0.002, 0.090]. \Cref{eq:twocomp} gives the share of the twelve scores within 0.013 [0.009, 0.024] on average (largest 0.037, content-shift check; $r=0.97$; mean predicted minus measured 0.008); for the detector it predicts 0.667 against 0.647 measured, for the supervised linear probe 0.601 against 0.601. With every input of \cref{eq:twocomp} and every pair weight taken from the new items instead, it predicts the shares measured on the held-out list within 0.015 on average (largest 0.035; $r=0.90$; detector 0.665 against 0.647). %

\input{tables/spredreplicationfull}
\input{tables/detectho}
\input{tables/heldout}
\input{tables/residheld}

\input{tables/heldoutpairs}

\input{sections/supplement_replication}

%% file: tables/prompts.tex
\begin{table}[!htbp]
\centering
\caption{Prompts and read-out settings. Top: the text of the user turn after the image; \texttt{\{ref\}} is the referring expression. Bottom: number of layers of the language model, and the layer and ridge penalty of the read-out, chosen without labels by five-fold cross-validation on the training items.}
\label{tab:prompts}
\small
\begin{tabular}{@{}>{\raggedright\arraybackslash}p{2.9cm}>{\raggedright\arraybackslash}p{8.0cm}@{}}
\toprule
Model & Prompt\\
\midrule
\qsmall, \qlarge, \qfour, \qeight & \texttt{Locate \textquotedbl\{ref\}\textquotedbl{} in the image, and output its bbox coordinates in JSON format.}\\
\qivl & \texttt{Please provide the bounding box coordinate of the region this sentence describes: \textless ref\textgreater\{ref\}\textless /ref\textgreater}\\
\qllava & \texttt{Please provide the bounding box coordinate of the region this sentence describes: \{ref\}.}\\
\bottomrule
\end{tabular}

\vspace{1em}
\begin{tabular}{@{}lccc@{}}
\toprule
Model & Layers & Read-out layer & Ridge penalty\\
\midrule
\qsmall & 36 & 28 & 1000\\
\qlarge & 28 & 24 & 1000\\
\qfour & 36 & 28 & 1000\\
\qeight & 36 & 24 & 1000\\
\qivl & 32 & 20 & 1000\\
\qllava & 32 & 20 & 1000\\ %
\bottomrule
\end{tabular}
\end{table}

%% file: tables/costs.tex
\begin{table}[!t]
\centering
\caption{What each check needs at test time. Labelled items: items with gold boxes of the task. Passes: forward passes of any model per item (\cref{sec:method:setup}); for the second model, asked only above the 70\% training quantile of SPRED, mean passes per item, range over five models (\cref{sup:second}). $^{a}$Beyond the weight fixed once with labels on development models (\cref{sec:method:combine}). SPRED takes 0.997 to 1.126 times the time of plain generation (three models, \cref{sup:repro}); the other checks were not timed under the same conditions.}
\label{tab:costs}%
\begin{tabular}{@{}lccc@{}}
\toprule
Check & \hq{Labelled}{items} & \hq{Other}{model} & Passes\\
\midrule
SPRED (ours) & none$^{a}$ & no & 1\\
Supervised linear probe~\cite{kossen2024sep} & 2,000 & no & 1\\
Re-asking, seven perturbed views & none & no & 8\\
SPRED, then a second model & none & yes & 1.36--1.41\\
Open-vocabulary detector~\cite{liu2024groundingdino} & none & yes & 2\\
\bottomrule
\end{tabular}
\end{table}

%% file: tables/detectdiff.tex
\begin{table}[!t]
\centering
\caption{AUROC of our detector minus that of each score (all errors), with 95\% paired-bootstrap intervals over images (10,000 resamples). Columns and dashes as in \cref{tab:detect}.}
\label{tab:detectdiff}
\scriptsize
\setlength{\tabcolsep}{2.5pt}
\resizebox{\textwidth}{!}{%
\begin{tabular}{lcccccc}
\toprule
 & \multicolumn{3}{c}{\qsmall} & \multicolumn{2}{c}{\qlarge} & \qfour\\
\cmidrule(lr){2-4}\cmidrule(lr){5-6}\cmidrule(lr){7-7}
Ours minus & Six sets$^{a}$ & New & RefCOCO+ & New & RefCOCO+ & New\\
 & 3{,}352 / 456 & 1{,}152 / 168 & 1{,}200 / 183 & 1{,}152 / 160 & 1{,}200 / 152 & 1{,}151 / 138\\
\midrule
Sampling-teacher probe & 0.031 [0.013, 0.049] & 0.049 [0.017, 0.082] & 0.026 [$-$0.001, 0.053] & 0.047 [0.012, 0.082] & 0.003 [$-$0.027, 0.032] & 0.034 [0.015, 0.052]\\
Linear probe, correctness & 0.011 [$-$0.004, 0.028] & 0.009 [$-$0.016, 0.035] & 0.007 [$-$0.018, 0.031] & 0.047 [0.014, 0.080] & 0.012 [$-$0.018, 0.042] & 0.023 [$-$0.003, 0.048]\\
Linear probe, $1-\mathrm{IoU}$ & 0.013 [$-$0.002, 0.029] & 0.015 [$-$0.010, 0.042] & $-$0.004 [$-$0.029, 0.020] & 0.049 [0.016, 0.082] & 0.011 [$-$0.020, 0.041] & 0.027 [0.003, 0.053]\\
SAPLMA & 0.035 [0.016, 0.054] & 0.039 [0.007, 0.071] & 0.021 [$-$0.012, 0.053] & 0.099 [0.060, 0.139] & 0.083 [0.041, 0.126] & 0.055 [0.021, 0.091]\\
Pre-generation probe & 0.015 [$-$0.002, 0.034] & 0.024 [$-$0.007, 0.056] & 0.020 [$-$0.009, 0.050] & --- & --- & 0.025 [$-$0.004, 0.054]\\
Box geometry & 0.205 [0.174, 0.236] & 0.236 [0.183, 0.290] & 0.201 [0.150, 0.249] & 0.247 [0.193, 0.300] & 0.209 [0.162, 0.256] & 0.266 [0.209, 0.325]\\
Mahalanobis & 0.094 [0.068, 0.118] & 0.109 [0.069, 0.150] & 0.132 [0.083, 0.181] & 0.214 [0.161, 0.267] & 0.089 [0.045, 0.133] & 0.188 [0.142, 0.235]\\
Coordinate log-prob & 0.151 [0.123, 0.180] & 0.109 [0.067, 0.153] & 0.152 [0.102, 0.201] & 0.134 [0.096, 0.175] & 0.188 [0.141, 0.236] & 0.182 [0.138, 0.227]\\
MTLA (original mapping) & --- & 0.223 [0.172, 0.276] & 0.167 [0.126, 0.208] & 0.248 [0.198, 0.299] & 0.167 [0.123, 0.213] & 0.265 [0.211, 0.322]\\
Sampling-teacher probe + residual & 0.017 [0.003, 0.031] & 0.026 [$-$0.000, 0.052] & 0.007 [$-$0.013, 0.025] & 0.034 [0.005, 0.064] & $-$0.006 [$-$0.032, 0.018] & 0.002 [$-$0.004, 0.009]\\
Linear probe (correctness) + residual & 0.003 [$-$0.010, 0.015] & 0.003 [$-$0.018, 0.023] & $-$0.004 [$-$0.024, 0.014] & 0.030 [0.004, 0.058] & $-$0.002 [$-$0.027, 0.022] & 0.012 [$-$0.009, 0.033]\\
Content-shift check (2 passes) & 0.083 [0.057, 0.110] & 0.064 [0.021, 0.107] & 0.080 [0.035, 0.125] & 0.062 [0.023, 0.100] & 0.104 [0.058, 0.152] & 0.168 [0.121, 0.216]\\
\bottomrule
\end{tabular}}
\end{table}

%% file: tables/labelsvsdetector.tex
\begin{table}[!htbp]
\centering
\caption{Labels against the label-free detector, new items. From the unlabelled training items, $N$ are drawn (five subsets per $N$; the largest is all training items, one subset). The supervised linear probe is trained on the correctness labels of these $N$ items; the detector uses the same $N$ items without labels, with configuration, read-out and standardization from them only. Entries: AUROC of the supervised probe minus that of the detector, mean over subsets, with 95\% bootstrap intervals over images in which the subset mean is taken in every resample; they do not include the variation from drawing other labelled items (between-subset standard deviation at $N=250$: 0.010 to 0.044). The detector fitted on all training items here uses its own cross-validation folds, so its column All does not match \cref{tab:detect} exactly.}
\label{tab:labelsvsdetector}
\footnotesize
\resizebox{\textwidth}{!}{%
\begin{tabular}{lcccc}
\toprule
 & $N=250$ & $N=500$ & $N=1{,}000$ & All\\
\midrule
\multicolumn{5}{l}{\emph{Detector fitted on the same $N$ items, without labels}}\\
\qsmall & $-$0.046 [$-$0.067, $-$0.024] & $-$0.015 [$-$0.034, 0.005] & $-$0.008 [$-$0.034, 0.018] & $-$0.009 [$-$0.035, 0.016]\\
\qlarge & $-$0.052 [$-$0.078, $-$0.027] & $-$0.015 [$-$0.042, 0.010] & $-$0.024 [$-$0.043, $-$0.005] & $-$0.040 [$-$0.073, $-$0.007]\\
\qfour & $-$0.058 [$-$0.079, $-$0.038] & $-$0.035 [$-$0.062, $-$0.008] & $-$0.036 [$-$0.060, $-$0.012] & $-$0.023 [$-$0.048, 0.003]\\
\qeight & $-$0.027 [$-$0.053, $-$0.001] & $-$0.035 [$-$0.063, $-$0.005] & $-$0.019 [$-$0.044, 0.005] & $-$0.002 [$-$0.026, 0.023]\\
\qivl & $-$0.095 [$-$0.120, $-$0.070] & $-$0.049 [$-$0.071, $-$0.028] & $-$0.032 [$-$0.055, $-$0.009] & 0.002 [$-$0.019, 0.023]\\
\midrule
\multicolumn{5}{l}{\emph{Detector fitted on all unlabelled training items}}\\
\qsmall & $-$0.089 [$-$0.114, $-$0.064] & $-$0.053 [$-$0.079, $-$0.028] & $-$0.024 [$-$0.050, 0.001] & $-$0.009 [$-$0.035, 0.016]\\
\qlarge & $-$0.128 [$-$0.158, $-$0.096] & $-$0.069 [$-$0.095, $-$0.042] & $-$0.049 [$-$0.076, $-$0.022] & $-$0.040 [$-$0.073, $-$0.007]\\
\qfour & $-$0.102 [$-$0.124, $-$0.081] & $-$0.064 [$-$0.090, $-$0.038] & $-$0.049 [$-$0.074, $-$0.024] & $-$0.023 [$-$0.048, 0.003]\\
\qeight & $-$0.076 [$-$0.104, $-$0.047] & $-$0.057 [$-$0.086, $-$0.028] & $-$0.027 [$-$0.053, $-$0.001] & $-$0.002 [$-$0.026, 0.023]\\
\qivl & $-$0.155 [$-$0.179, $-$0.130] & $-$0.088 [$-$0.110, $-$0.066] & $-$0.055 [$-$0.076, $-$0.034] & 0.002 [$-$0.019, 0.023]\\
\bottomrule
\end{tabular}}
\end{table}

%% file: tables/boundsupervised.tex
\begin{table}[!htbp]
\centering
\caption{The detector against the supervised linear probe on correctness (trained on the gold boxes of the 2{,}000 training items); on Objects365 the detector is trained there without labels and the probe is the one trained on RefCOCO labels. Difference: detector minus probe. One-sided bound: 5th percentile of the image-level paired bootstrap (10,000 resamples), the same resamples as for the two-sided 95\% interval. Within 0.03: one-sided bound above $-0.03$; for \qivl{} the bound rounds to $-0.030$ in both single-set columns and lies just above $-0.03$ on the new items and just below it on RefCOCO+.}
\label{tab:boundsupervised}
\footnotesize
\resizebox{\textwidth}{!}{%
\begin{tabular}{lcccccc}
\toprule
Model, items & Detector & Supervised & Difference & One-sided bound & 95\% interval & Within 0.03\\
\midrule
\qsmall, new & 0.829 & 0.820 & +0.009 & $-$0.012 & [$-$0.016, +0.035] & yes\\
\qsmall, RefCOCO+ & 0.830 & 0.823 & +0.007 & $-$0.014 & [$-$0.018, +0.031] & yes\\
\qsmall, both & 0.830 & 0.818 & +0.012 & $-$0.003 & [$-$0.006, +0.029] & yes\\
\qlarge, new & 0.848 & 0.802 & +0.047 & +0.019 & [+0.014, +0.080] & yes\\
\qlarge, RefCOCO+ & 0.822 & 0.810 & +0.012 & $-$0.014 & [$-$0.018, +0.042] & yes\\
\qlarge, both & 0.833 & 0.804 & +0.029 & +0.010 & [+0.007, +0.051] & yes\\
\qfour, new & 0.858 & 0.835 & +0.023 & +0.002 & [$-$0.003, +0.048] & yes\\
\qeight, new & 0.837 & 0.835 & +0.002 & $-$0.019 & [$-$0.023, +0.026] & yes\\
\qivl, new & 0.867 & 0.876 & $-$0.009 & $-$0.030 & [$-$0.034, +0.016] & yes\\
\qivl, RefCOCO+ & 0.881 & 0.892 & $-$0.012 & $-$0.030 & [$-$0.034, +0.011] & no\\
\qivl, both & 0.873 & 0.884 & $-$0.011 & $-$0.025 & [$-$0.028, +0.005] & yes\\
\midrule
\multicolumn{7}{l}{\emph{Objects365, against the supervised probe trained on RefCOCO labels}}\\
\qsmall & 0.758 & 0.731 & +0.026 & +0.003 & [$-$0.002, +0.055] & yes\\
\qlarge & 0.692 & 0.687 & +0.004 & $-$0.032 & [$-$0.039, +0.048] & no\\
\qfour & 0.762 & 0.714 & +0.048 & +0.020 & [+0.015, +0.081] & yes\\
\bottomrule
\end{tabular}}
\end{table}

%% file: tables/surface.tex
\begin{table}[!htbp]
\centering
\caption{Supervised baselines on surface features of the item, new items answered by all five models (1{,}146 items). Ridge regressions on correctness, fitted on each model's own labelled training items. Expression and image: number of words and characters, image area and aspect ratio. Written-box geometry: area, width, height, centre and its distance from the image centre, aspect ratio. Gold-box size needs the annotation and serves only as a reference. Brackets: 95\% paired-bootstrap intervals over images.}
\label{tab:surface}
\footnotesize
\resizebox{\textwidth}{!}{%
\begin{tabular}{lccccc}
\toprule
 & \qsmall & \qlarge & \qfour & \qeight & \qivl\\
\midrule
Detector & 0.828 & 0.850 & 0.859 & 0.838 & 0.869\\
Expression and image & 0.544 & 0.522 & 0.517 & 0.519 & 0.528\\
\quad + written-box geometry & 0.579 & 0.582 & 0.571 & 0.584 & 0.558\\
\quad + gold-box size (reference) & 0.590 & 0.581 & 0.601 & 0.602 & 0.576\\
Detector minus expression, image and written box & +0.249 [+0.200, +0.298] & +0.268 [+0.215, +0.318] & +0.288 [+0.234, +0.344] & +0.254 [+0.204, +0.304] & +0.311 [+0.257, +0.364]\\
\bottomrule
\end{tabular}}
\end{table}

%% file: tables/ablationdiff.tex
\begin{table}[!htbp]
\centering
\caption{Ablation differences of \cref{tab:ablation} with 95\% paired-bootstrap intervals over images (10,000 resamples). All errors against all other items; wrong-object errors against correct answers.}
\label{tab:ablationdiff}
\scriptsize
\setlength{\tabcolsep}{2pt}
\resizebox{\textwidth}{!}{%
\begin{tabular}{lcccccc}
\toprule
 & \multicolumn{2}{c}{\qsmall{} (six sets)} & \multicolumn{2}{c}{\qlarge{} (new)} & \multicolumn{2}{c}{\qfour{} (new)}\\
\cmidrule(lr){2-3}\cmidrule(lr){4-5}\cmidrule(lr){6-7}
Difference & All errors & Wrong object & All errors & Wrong object & All errors & Wrong object\\
\midrule
Ours $-$ probe only & 0.018 [0.009, 0.027] & 0.056 [0.043, 0.069] & 0.016 [0.002, 0.030] & 0.049 [0.022, 0.076] & 0.027 [0.011, 0.044] & 0.063 [0.041, 0.086]\\
Ours $-$ residual only & 0.091 [0.071, 0.111] & 0.010 [$-$0.011, 0.031] & 0.123 [0.088, 0.158] & 0.033 [$-$0.003, 0.071] & 0.086 [0.053, 0.119] & $-$0.000 [$-$0.037, 0.038]\\
Adding coordinate log-prob & $-$0.017 [$-$0.028, $-$0.006] & $-$0.054 [$-$0.071, $-$0.038] & $-$0.015 [$-$0.031, $-$0.001] & $-$0.030 [$-$0.052, $-$0.009] & $-$0.032 [$-$0.051, $-$0.014] & $-$0.048 [$-$0.075, $-$0.025]\\
Two teachers $-$ shift teacher & 0.016 [0.008, 0.024] & 0.008 [$-$0.004, 0.021] & 0.027 [$-$0.008, 0.064] & $-$0.034 [$-$0.084, 0.017] & 0.028 [0.005, 0.051] & $-$0.017 [$-$0.054, 0.019]\\
Two teachers $-$ sampling teacher & 0.013 [$-$0.006, 0.033] & $-$0.019 [$-$0.047, 0.009] & 0.031 [$-$0.007, 0.069] & $-$0.040 [$-$0.085, 0.003] & 0.007 [$-$0.002, 0.015] & 0.022 [0.009, 0.036]\\
\bottomrule
\end{tabular}}
\end{table}

\begin{table}[!htbp]
\centering
\caption{Other statistics with 95\% intervals. Read-out analysis (\cref{sec:measure}): regression coefficients of the read-out displacement of wrong-object errors on the directions to the gold box and to the mean training box, and the gain of the detector over the probe on wrong-object errors when items are compared only within 12 bins of the residual predicted from box geometry; paired bootstrap over images. Second passes (\cref{sup:second}): gain of each added pass over the detector, \qlarge{} new items, paired bootstrap over items.}
\label{tab:otherdiff}
\footnotesize
\resizebox{\textwidth}{!}{%
\begin{tabular}{lccc}
\toprule
 & \qsmall{} (six sets) & \qlarge{} (new) & \qfour{} (new)\\
\midrule
Coefficient towards the gold box & 0.268 [0.216, 0.318] & 0.275 [0.185, 0.367] & 0.317 [0.240, 0.390]\\
Coefficient towards the mean box & 0.368 [0.291, 0.440] & 0.248 [0.124, 0.358] & 0.194 [0.085, 0.299]\\
Ours $-$ probe, wrong object, 12 geometry bins (3B: new) & 0.052 [0.024, 0.082] & 0.043 [0.007, 0.079] & 0.043 [0.021, 0.065]\\
Same, RefCOCO+ & 0.048 [0.025, 0.072] & 0.057 [0.028, 0.088] & ---\\
\midrule
Added content shift & --- & 0.012 [$-$0.006, 0.030] & ---\\
Added resampled answer & --- & $-$0.014 [$-$0.032, 0.005] & ---\\
Added JPEG compression & --- & $-$0.001 [$-$0.018, 0.016] & ---\\
Added box of \qsmall{} & --- & 0.013 [$-$0.006, 0.035] & ---\\
Added box of \qfour{} & --- & 0.028 [0.008, 0.050] & ---\\
\bottomrule
\end{tabular}}
\end{table}

%% file: tables/residadddiff.tex
\begin{table}[!htbp]
\centering
\caption{The self-prediction residual added to five probes on the RefCOCO-family sets: AUROC of base $+\,0.5\times$ residual minus that of the base alone, with the same standardization, for wrong-object errors against correct answers (upper block) and for all errors (lower block), with 95\% paired-bootstrap intervals over images (10,000 resamples). Development: the four development sets of \qsmall{} (RefCOCO testA, testB and validation, RefCOCOg validation). $^{\ast}$Supervised with the gold boxes of the 2,000 training items. Dashes in the pre-generation rows of \qlarge: the same probe as the linear probe on correctness there.}
\label{tab:residadddiff}
\scriptsize
\setlength{\tabcolsep}{2pt}
\resizebox{\textwidth}{!}{%
\begin{tabular}{lcccccc}
\toprule
 & \multicolumn{3}{c}{\qsmall} & \multicolumn{2}{c}{\qlarge} & \qfour\\
\cmidrule(lr){2-4}\cmidrule(lr){5-6}\cmidrule(lr){7-7}
Base & Development & New & RefCOCO+ & New & RefCOCO+ & New\\
\midrule
\multicolumn{7}{l}{\emph{Wrong-object errors against correct answers}}\\
Two-teacher probe & 0.051 [0.037, 0.068] & 0.061 [0.039, 0.087] & 0.055 [0.031, 0.080] & 0.049 [0.023, 0.077] & 0.060 [0.033, 0.089] & 0.063 [0.042, 0.086]\\
Sampling-teacher probe & 0.036 [0.024, 0.051] & 0.064 [0.040, 0.089] & 0.039 [0.020, 0.059] & 0.029 [0.008, 0.051] & 0.030 [0.011, 0.050] & 0.074 [0.052, 0.098]\\
Linear probe, correctness & 0.054 [0.036, 0.074] & 0.050 [0.032, 0.069] & 0.036 [0.018, 0.053] & 0.037 [0.013, 0.062] & 0.037 [0.015, 0.060] & 0.037 [0.021, 0.055]\\
Pre-generation probe & 0.038 [0.026, 0.051] & 0.053 [0.032, 0.076] & 0.036 [0.018, 0.053] & --- & --- & 0.041 [0.023, 0.061]\\
SAPLMA & 0.056 [0.038, 0.075] & 0.061 [0.039, 0.086] & 0.035 [0.021, 0.052] & 0.070 [0.040, 0.105] & 0.091 [0.055, 0.130] & 0.046 [0.024, 0.073]\\
\midrule
\multicolumn{7}{l}{\emph{All errors}}\\
Two-teacher probe & 0.016 [0.006, 0.027] & 0.020 [0.004, 0.037] & 0.025 [0.008, 0.043] & 0.016 [0.001, 0.030] & 0.031 [0.011, 0.052] & 0.027 [0.012, 0.044]\\
Sampling-teacher probe & 0.010 [0.000, 0.019] & 0.023 [0.007, 0.039] & 0.020 [0.006, 0.035] & 0.013 [$-$0.001, 0.027] & 0.009 [$-$0.005, 0.024] & 0.032 [0.015, 0.048]\\
Linear probe, correctness & 0.010 [$-$0.001, 0.021] & 0.006 [$-$0.007, 0.020] & 0.012 [$-$0.002, 0.027] & 0.016 [0.003, 0.030] & 0.014 [$-$0.001, 0.031] & 0.010 [$-$0.002, 0.024]\\
Pre-generation probe & 0.008 [$-$0.002, 0.018] & 0.012 [$-$0.003, 0.025] & 0.018 [0.005, 0.031] & --- & --- & 0.016 [0.003, 0.030]\\
SAPLMA & 0.019 [0.009, 0.030] & 0.021 [0.007, 0.036] & 0.017 [0.004, 0.031] & 0.043 [0.024, 0.064] & 0.054 [0.030, 0.080] & 0.032 [0.013, 0.050]\\
\bottomrule
\end{tabular}}
\end{table}

%% file: tables/risk80.tex
\begin{table}[!htbp]
\centering
\caption{Error rate among the answers kept when the 20\% with the highest score are handed on (coverage 0.8), by the score that chooses them; All: error rate of all answers. Interval for the detector from a bootstrap over images (1,000 resamples). Held-out items: \cref{sup:heldout}.}
\label{tab:risk80}
\footnotesize
\begin{tabular}{@{}lcccc@{}}
\toprule
Model & All & Detector & Supervised linear & Coordinate log-prob\\
\midrule
\multicolumn{5}{@{}l}{\emph{New items}}\\
\qsmall{} & 0.146 & 0.079 {\scriptsize[0.059, 0.100]} & 0.079 & 0.104\\
\qlarge{} & 0.139 & 0.066 {\scriptsize[0.051, 0.085]} & 0.086 & 0.102\\
\qfour{} & 0.120 & 0.051 {\scriptsize[0.037, 0.068]} & 0.052 & 0.091\\
\qeight{} & 0.110 & 0.050 {\scriptsize[0.034, 0.066]} & 0.052 & 0.084\\
\qivl{} & 0.120 & 0.053 {\scriptsize[0.037, 0.071]} & 0.048 & 0.101\\
\qllava{} & 0.368 & 0.279 {\scriptsize[0.251, 0.310]} & 0.274 & 0.318\\
\midrule
\multicolumn{5}{@{}l}{\emph{Held-out items}}\\
\qsmall{} & 0.148 & 0.076 {\scriptsize[0.059, 0.094]} & 0.085 & 0.119\\
\qlarge{} & 0.127 & 0.081 {\scriptsize[0.063, 0.100]} & 0.083 & 0.104\\
\qfour{} & 0.117 & 0.062 {\scriptsize[0.048, 0.080]} & 0.061 & 0.092\\
\qeight{} & 0.108 & 0.050 {\scriptsize[0.035, 0.066]} & 0.053 & 0.080\\
\qivl{} & 0.118 & 0.064 {\scriptsize[0.047, 0.080]} & 0.065 & 0.097\\
\qllava{} & 0.400 & 0.311 {\scriptsize[0.279, 0.343]} & 0.302 & 0.339\\
\bottomrule
\end{tabular}
\end{table}

%% file: tables/referral.tex
\begin{table}[!htbp]
\centering
\caption{Handing on the answers with the highest scores, averaged over \qsmall, \qlarge, \qfour, \qeight{} and \qivl: precision (share of the handed-on answers that are wrong), recall of all errors and recall of wrong-object errors, for the detector (D), the supervised linear probe (S) and coordinate log-prob (L). Ties are broken by item order. Averaged over the five models, D minus L is above zero with a 95\% interval over images in every cell; D minus S has an interval including zero for precision and recall at every share.}
\label{tab:referral}
\footnotesize
\setlength{\tabcolsep}{3.5pt}
\begin{tabular}{@{}llccccccccc@{}}
\toprule
& & \multicolumn{3}{c}{Precision} & \multicolumn{3}{c}{Recall} & \multicolumn{3}{c}{Recall, wrong object}\\
\cmidrule(lr){3-5}\cmidrule(lr){6-8}\cmidrule(lr){9-11}
Set & Handed on & D & S & L & D & S & L & D & S & L\\
\midrule
New items & 10\% & 0.492 & 0.475 & 0.281 & 0.392 & 0.381 & 0.221 & 0.444 & 0.382 & 0.174\\
 & 20\% & 0.393 & 0.380 & 0.248 & 0.625 & 0.605 & 0.389 & 0.665 & 0.625 & 0.321\\
 & 30\% & 0.327 & 0.313 & 0.215 & 0.775 & 0.746 & 0.505 & 0.813 & 0.776 & 0.437\\
\addlinespace
Held-out list & 10\% & 0.482 & 0.447 & 0.273 & 0.390 & 0.363 & 0.221 & 0.421 & 0.376 & 0.184\\
 & 20\% & 0.352 & 0.340 & 0.225 & 0.570 & 0.553 & 0.365 & 0.598 & 0.563 & 0.340\\
 & 30\% & 0.296 & 0.283 & 0.196 & 0.717 & 0.688 & 0.474 & 0.770 & 0.681 & 0.416\\
\bottomrule
\end{tabular}
\end{table}

%% file: tables/metrics2.tex
\begin{table}[!htbp]
\centering
\caption{Second metrics on the new items. AUPRC: average precision for errors (IoU $<0.5$), whose chance level is the error rate of the model. FPR95: share of correct answers flagged at the threshold that catches 95\% of the errors (lower is better). MTLA needs the coordinate-digit attention of a Qwen model (n/a for \qivl). Digit-weighted: one minus the expected IoU over the stored per-digit alternatives of the same pass (\cref{tab:detect}); the alternatives were stored for three models only (n/a).}
\label{tab:metrics2}
\footnotesize
\setlength{\tabcolsep}{3pt}
\begin{tabular}{@{}>{\raggedright\arraybackslash}p{3.3cm}ccccc@{}}
\toprule
 & \hqsmall & \hqlarge & \hqfour & \hqeight & \hqivl\\
\midrule
Error rate (chance AUPRC) & 0.146 & 0.139 & 0.120 & 0.110 & 0.120\\
\midrule
\multicolumn{6}{l}{AUPRC (higher is better)}\\
Detector (ours) & 0.475 & 0.522 & 0.436 & 0.388 & 0.497\\
Sampling-teacher probe & 0.376 & 0.390 & 0.380 & 0.355 & 0.469\\
Content shift (two passes) & 0.433 & 0.438 & 0.280 & 0.236 & 0.383\\
Mahalanobis distance & 0.326 & 0.217 & 0.223 & 0.301 & 0.189\\
Coordinate log-prob & 0.348 & 0.291 & 0.216 & 0.204 & 0.197\\
MTLA (layers 8--21) & 0.255 & 0.264 & 0.218 & 0.250 & n/a\\
Digit-weighted coordinate uncertainty & 0.408 & 0.391 & 0.212 & n/a & n/a\\
\midrule
\multicolumn{6}{l}{FPR95 (lower is better)}\\
Detector (ours) & 0.591 & 0.531 & 0.491 & 0.542 & 0.497\\
Sampling-teacher probe & 0.686 & 0.571 & 0.573 & 0.606 & 0.632\\
Content shift (two passes) & 0.921 & 0.814 & 0.873 & 0.866 & 0.822\\
Mahalanobis distance & 0.839 & 0.840 & 0.815 & 0.642 & 0.854\\
Coordinate log-prob & 0.834 & 0.826 & 0.766 & 0.855 & 0.919\\
MTLA (layers 8--21) & 0.933 & 0.955 & 0.958 & 0.917 & n/a\\
Digit-weighted coordinate uncertainty & 0.760 & 0.769 & 0.732 & n/a & n/a\\
\bottomrule
\end{tabular}
\end{table}

%% file: tables/views.tex
\begin{table}[!htbp]
\centering
\caption{One pass of the detector against repeated consistency checks on the new items. Seven views: the disagreement of the box with its versions under seven views (content shifts, resampled answers, JPEG compression, noise), eight passes in all. Mean over $k$ views: the mean disagreement over $k$ views drawn at random from eleven (\qsmall{}, \qlarge{}) or seven (\qfour{}), $k+1$ passes, averaged over up to 30 draws. Brackets: 95\% paired-bootstrap interval over images.}
\label{tab:views}
\footnotesize
\resizebox{\textwidth}{!}{%
\begin{tabular}{lccc}
\toprule
Score (forward passes) & \qsmall & \qlarge & \qfour\\
\midrule
Detector (1) & 0.829 & 0.848 & 0.858\\
Seven views (8) & 0.837 & 0.834 & 0.780\\
Detector minus seven views & $-$0.008 [$-$0.040, 0.023] & 0.014 [$-$0.017, 0.046] & 0.078 [0.033, 0.120]\\
\midrule
Mean over $k=1$ view (2) & 0.700 & 0.721 & 0.680\\
Mean over $k=3$ views (4) & 0.787 & 0.796 & 0.742\\
Mean over $k=7$ views (8) & 0.830 & 0.833 & 0.780\\
Mean over $k=11$ views (12) & 0.842 & 0.847 & ---\\
\bottomrule
\end{tabular}}
\end{table}

%% file: tables/seeds.tex
\begin{table}[!htbp]
\centering
\caption{Randomness on the training side: the detector minus the supervised linear probe, refitted with the stored fold seed and four further ones, each of which reselects layer and ridge strength by cross-validation. Upper part: AUROC on natural errors (IoU $<0.5$); lower part: share of the discordant cross-model cases won (\cref{sec:ctl:models}). Difference: stored run; SD: standard deviation over the five seeds; interval: 95\% interval over images and seeds, nested. Where a seed selects the same layer and strength, the refit is identical, since the ridge fit is deterministic given the folds.}
\label{tab:seeds}
\footnotesize
\setlength{\tabcolsep}{5.5pt}
\begin{tabular}{@{}llccc@{}}
\toprule
Set & Model & Difference & \hq{SD over}{seeds} & \hq{Interval over}{images and seeds}\\
\midrule
New items & \qsmall{} & 0.009 & 0.000 & [$-$0.016, 0.035]\\
 & \qlarge{} & 0.047 & 0.010 & [0.008, 0.077]\\
 & \qfour{} & 0.023 & 0.000 & [$-$0.002, 0.048]\\
 & \qeight{} & 0.002 & 0.017 & [$-$0.022, 0.060]\\
 & \qivl{} & $-$0.009 & 0.003 & [$-$0.026, 0.019]\\
 & \qllava{} & 0.001 & 0.002 & [$-$0.017, 0.017]\\
\addlinespace
RefCOCO+ & \qsmall{} & 0.007 & 0.000 & [$-$0.018, 0.031]\\
 & \qlarge{} & 0.012 & 0.009 & [$-$0.018, 0.046]\\
 & \qivl{} & $-$0.012 & 0.005 & [$-$0.026, 0.021]\\
\addlinespace
Held-out list & \qsmall{} & 0.022 & 0.000 & [$-$0.009, 0.051]\\
 & \qlarge{} & 0.026 & 0.005 & [$-$0.008, 0.068]\\
 & \qfour{} & 0.014 & 0.000 & [$-$0.017, 0.046]\\
 & \qeight{} & $-$0.010 & 0.011 & [$-$0.037, 0.037]\\
 & \qivl{} & 0.022 & 0.002 & [$-$0.001, 0.053]\\
 & \qllava{} & $-$0.014 & 0.001 & [$-$0.032, 0.004]\\
\midrule
Across models, new items & five models & 0.069 & 0.015 & [0.003, 0.104]\\
Across models, held-out list & five models & 0.046 & 0.006 & [0.007, 0.098]\\
\bottomrule
\end{tabular}
\end{table}

%% file: tables/trainres.tex
\begin{sidewaystable}
\centering
\caption{Training items redrawn: each of ten refits per model draws the 2{,}000 training items with replacement (copies of an item stay in one fold) and selects and fits the probes and the read-out again; for \qsmall{} and \qlarge{} it also redraws which of the two stored sampled answers enters the sampling teacher. Upper part: AUROC on natural errors, detector minus supervised linear probe (all errors) and the increment of the supervised linear probe when the residual is added (wrong-object errors). Lower part: discordant cross-model cases. Mean: mean over the refits; nested: 95\% interval over images, each draw paired with one random refit; combined: stored value $\pm 1.96$ times the root of the summed image variance and refit variance. A redrawn training set holds about 63\% distinct items and both probes are somewhat weaker than in the stored run, so the refits approximate the variation due to the sampling of training items rather than estimate it exactly.}
\label{tab:trainres}
\footnotesize
\setlength{\tabcolsep}{3pt}
\begin{tabular}{@{}llcccccc@{}}
\toprule
& & \multicolumn{3}{c}{Detector $-$ supervised} & \multicolumn{3}{c}{Residual added to supervised}\\
\cmidrule(lr){3-5}\cmidrule(lr){6-8}
Set & Model & Mean & Nested & Combined & Mean & Nested & Combined\\
\midrule
New items & \qsmall{} & 0.023 & {\scriptsize[$-$0.014, 0.062]} & {\scriptsize[$-$0.025, 0.043]} & 0.053 & {\scriptsize[0.029, 0.082]} & {\scriptsize[0.026, 0.073]}\\
 & \qlarge{} & 0.043 & {\scriptsize[$-$0.004, 0.104]} & {\scriptsize[$-$0.006, 0.099]} & 0.041 & {\scriptsize[0.013, 0.078]} & {\scriptsize[0.005, 0.069]}\\
 & \qfour{} & 0.029 & {\scriptsize[$-$0.006, 0.072]} & {\scriptsize[$-$0.011, 0.056]} & 0.041 & {\scriptsize[0.016, 0.078]} & {\scriptsize[0.009, 0.065]}\\
 & \qeight{} & 0.020 & {\scriptsize[$-$0.020, 0.089]} & {\scriptsize[$-$0.044, 0.048]} & 0.038 & {\scriptsize[0.013, 0.071]} & {\scriptsize[0.001, 0.051]}\\
 & \qivl{} & 0.013 & {\scriptsize[$-$0.035, 0.064]} & {\scriptsize[$-$0.058, 0.040]} & 0.060 & {\scriptsize[0.028, 0.100]} & {\scriptsize[0.009, 0.079]}\\
 & \qllava{} & 0.012 & {\scriptsize[$-$0.014, 0.053]} & {\scriptsize[$-$0.030, 0.032]} & 0.049 & {\scriptsize[0.025, 0.077]} & {\scriptsize[0.017, 0.070]}\\
\addlinespace
RefCOCO+ & \qsmall{} & 0.025 & {\scriptsize[$-$0.007, 0.063]} & {\scriptsize[$-$0.025, 0.039]} & 0.039 & {\scriptsize[0.018, 0.061]} & {\scriptsize[0.015, 0.057]}\\
 & \qlarge{} & 0.016 & {\scriptsize[$-$0.036, 0.107]} & {\scriptsize[$-$0.055, 0.079]} & 0.043 & {\scriptsize[0.014, 0.087]} & {\scriptsize[$-$0.000, 0.074]}\\
 & \qivl{} & 0.009 & {\scriptsize[$-$0.039, 0.074]} & {\scriptsize[$-$0.067, 0.044]} & 0.040 & {\scriptsize[0.007, 0.083]} & {\scriptsize[$-$0.007, 0.067]}\\
\addlinespace
Development & \qsmall{} & 0.013 & {\scriptsize[$-$0.012, 0.040]} & {\scriptsize[$-$0.013, 0.038]} & 0.042 & {\scriptsize[0.020, 0.067]} & {\scriptsize[0.027, 0.082]}\\
\addlinespace
Held-out list & \qsmall{} & 0.051 & {\scriptsize[$-$0.018, 0.114]} & {\scriptsize[$-$0.050, 0.093]} & 0.064 & {\scriptsize[0.038, 0.097]} & {\scriptsize[0.025, 0.089]}\\
 & \qlarge{} & 0.034 & {\scriptsize[$-$0.014, 0.105]} & {\scriptsize[$-$0.036, 0.089]} & 0.039 & {\scriptsize[0.002, 0.085]} & {\scriptsize[$-$0.009, 0.074]}\\
 & \qfour{} & 0.015 & {\scriptsize[$-$0.032, 0.056]} & {\scriptsize[$-$0.027, 0.054]} & 0.041 & {\scriptsize[0.011, 0.071]} & {\scriptsize[0.018, 0.074]}\\
 & \qeight{} & 0.008 & {\scriptsize[$-$0.037, 0.065]} & {\scriptsize[$-$0.061, 0.040]} & 0.022 & {\scriptsize[$-$0.011, 0.058]} & {\scriptsize[$-$0.027, 0.042]}\\
 & \qivl{} & 0.034 & {\scriptsize[$-$0.014, 0.091]} & {\scriptsize[$-$0.031, 0.076]} & 0.062 & {\scriptsize[0.026, 0.099]} & {\scriptsize[0.016, 0.095]}\\
 & \qllava{} & $-$0.001 & {\scriptsize[$-$0.024, 0.030]} & {\scriptsize[$-$0.041, 0.013]} & 0.029 & {\scriptsize[0.007, 0.053]} & {\scriptsize[$-$0.001, 0.047]}\\
\midrule
& & \multicolumn{3}{c}{Detector $-$ supervised} & \multicolumn{3}{c}{Detector's share}\\
\cmidrule(lr){3-5}\cmidrule(lr){6-8}
Across models & New items & 0.027 & {\scriptsize[$-$0.035, 0.084]} & {\scriptsize[0.011, 0.127]} & 0.602 & {\scriptsize[0.558, 0.647]} & {\scriptsize[0.589, 0.680]}\\
Across models & Held-out list & 0.053 & {\scriptsize[$-$0.005, 0.111]} & {\scriptsize[$-$0.011, 0.103]} & 0.633 & {\scriptsize[0.585, 0.680]} & {\scriptsize[0.600, 0.693]}\\
\bottomrule
\end{tabular}
\end{sidewaystable}

%% file: tables/ablation.tex
\begin{table}[!t]
\centering
\caption{Ablations under two scorings: all errors (All; IoU $<0.5$) against all other items, and wrong-object errors (IoU $<0.1$) against correct answers. \qsmall: six sets, and \qfour: new items, both used when the weight was chosen; \qlarge: new items, examined after it was fixed. Lower block: differences. Bold: highest in each column of the upper block.}
\label{tab:ablation}
\scriptsize
\setlength{\tabcolsep}{2pt}
\begin{tabular}{lcccccc}
\toprule
 & \multicolumn{2}{c}{\qsmall} & \multicolumn{2}{c}{\qlarge} & \multicolumn{2}{c}{\qfour}\\
\cmidrule(lr){2-3}\cmidrule(lr){4-5}\cmidrule(lr){6-7}
Variant & All & Wrong object & All & Wrong object & All & Wrong object\\
\midrule
Probe only & 0.813 & 0.829 & 0.833 & 0.818 & 0.831 & 0.805\\
Residual only & 0.740 & 0.875 & 0.725 & 0.833 & 0.773 & \textbf{0.867}\\
Probe + 0.5 $\times$ residual (ours) & \textbf{0.831} & \textbf{0.885} & \textbf{0.848} & \textbf{0.866} & \textbf{0.858} & \textbf{0.867}\\
\midrule
Ours $-$ probe only & 0.018 & 0.056 & 0.016 & 0.049 & 0.027 & 0.063\\
Ours $-$ residual only & 0.091 & 0.010 & 0.123 & 0.033 & 0.086 & $-$0.000\\
Adding coordinate log-prob & $-$0.017 & $-$0.054 & $-$0.015 & $-$0.030 & $-$0.032 & $-$0.048\\
Two teachers $-$ shift teacher (probe) & 0.016 & 0.008 & 0.027 & $-$0.034 & 0.028 & $-$0.017\\
Two teachers $-$ sampling teacher (probe) & 0.013 & $-$0.019 & 0.031 & $-$0.040 & 0.007 & 0.022\\
\bottomrule
\end{tabular}
\end{table}

%% file: tables/overlapbinormal.tex
\begin{table}[!htbp]
\centering
\caption{The residual on overlap errors (partial overlap and wrong extent together) under a binormal model, new items and held-out list, stored fit, means over the six models. Within-class correlation: of the standardized probe and residual, correct answers pooled with the overlap errors. Remainder: the residual minus the part predicted by the probe (slope fitted on the training items). Increments of the fifteen scores when the residual is added: predicted by the binormal model with the measured correlation and AUROCs against measured. Optimal weight: computed after the results.}
\label{tab:overlapbinormal}
\footnotesize
\begin{tabular}{@{}lcc@{}}
\toprule
& New items & Held-out list\\
\midrule
Within-class correlation, overlap errors & 0.344 & 0.352\\
Remainder, AUROC on overlap errors & 0.520 & 0.526\\
Remainder, AUROC on wrong-object errors & 0.761 & 0.726\\
Increments of 15 scores, predicted against measured: rank correlation & 0.993 & 0.996\\
\quad mean absolute difference & 0.0025 & 0.0033\\
Increment against AUROC on overlap errors: rank correlation & $-$0.964 & $-$0.971\\
Optimal weight for all errors (post hoc) & 0.50 & 0.48\\
\bottomrule
\end{tabular}
\end{table}

%% file: tables/tenk.tex
\begin{table}[!htbp]
\centering
\caption{SPRED and the supervised linear probe trained on correctness with matched numbers of training items: AUROC for all errors. SPRED is fitted on the items without their labels, the probe with them, its feature, layer and penalty searched at each size. $\Delta$: SPRED minus the probe. Last row: mean over the two models at the largest size, 95\% interval combining image resampling and ten refits on resampled training items, and the one-sided 95\% lower bound.}
\label{tab:tenk}%
\setlength{\tabcolsep}{4pt}
\begin{tabular}{@{}llcccccc@{}}
\toprule
& & \multicolumn{3}{c}{Replication list} & \multicolumn{3}{c}{Held-out list}\\
\cmidrule(lr){3-5}\cmidrule(l){6-8}
Model & Items & SPRED & Probe & $\Delta$ & SPRED & Probe & $\Delta$\\
\midrule
\qsmall{} & 2,000 & 0.792 & 0.778 & 0.014 & 0.838 & 0.816 & 0.022\\
 & 5,000 & 0.812 & 0.806 & 0.006 & 0.850 & 0.826 & 0.024\\
 & 10,000 & 0.812 & 0.822 & $-$0.010 & 0.850 & 0.838 & 0.012\\ \addlinespace
\qlarge{} & 2,000 & 0.785 & 0.769 & 0.016 & 0.765 & 0.739 & 0.026\\
 & 5,000 & 0.809 & 0.768 & 0.040 & 0.782 & 0.762 & 0.020\\
 & 9,997 & 0.814 & 0.778 & 0.036 & 0.794 & 0.803 & $-$0.009\\
\midrule
\multicolumn{2}{@{}l}{Mean $\Delta$, largest size} & \multicolumn{3}{c}{\hq{0.013 [$-$0.011, 0.037]}{bound $-$0.007}} & \multicolumn{3}{c}{\hq{0.001 [$-$0.025, 0.028]}{bound $-$0.021}}\\
\bottomrule
\end{tabular}
\end{table}

%% file: tables/tenkadd.tex
\begin{table}[!htbp]
\centering
\caption{SPRED with supervised linear probes trained on 10,000 labelled items (9,997 for \qlarge{}): change in AUROC. Each cell gives the mean over the two models, the 95\% interval combining image resampling and ten refits where shown, and in parentheses \qsmall{} and \qlarge{}. Scores are added after standardization, SPRED fitted on the same items without their labels. Probe configuration chosen at 2,000 items and kept, or searched again at 10,000. Second probe: the supervised probe with the highest cross-validated AUROC among those reading another feature or layer, on the same labels. $^{\dagger}$Post hoc.}
\label{tab:tenkadd}%
\setlength{\tabcolsep}{3pt}
\begin{tabular}{@{}>{\raggedright\arraybackslash}p{3.6cm}ccc@{}}
\toprule
& Replication list & Held-out list & New items\\
\midrule
\multicolumn{4}{@{}l}{\emph{All errors}}\\
SPRED added to the probe on correctness & \hqt{0.027}{\relax[0.012, 0.042]}{\relax(0.010, 0.044)} & \hqtt{0.031}{\relax(0.023, 0.039)} & \hqtt{0.031}{\relax(0.019, 0.043)}\\ \addlinespace
\quad configuration searched again & \hqt{0.022}{\relax[0.009, 0.034]}{\relax(0.010, 0.033)} & \hqt{0.018}{\relax[0.000, 0.035]}{\relax(0.023, 0.012)} & \hqt{0.015}{\relax[$-$0.002, 0.032]}{\relax(0.019, 0.011)}\\ \addlinespace
SPRED added to the probe on $1-\mathrm{IoU}$ & \hqt{0.027}{\relax[0.014, 0.040]}{\relax(0.013, 0.041)} & \hqtt{0.028}{\relax(0.021, 0.034)} & \hqtt{0.028}{\relax(0.014, 0.042)}\\ \addlinespace
SPRED from 2,000 unlabelled items added to the probe on correctness & \hqt{0.017}{\relax[0.003, 0.030]}{\relax(0.001, 0.032)} & \hqtt{0.022}{\relax(0.017, 0.027)} & \hqtt{0.030}{\relax(0.015, 0.045)}\\ \addlinespace
\multicolumn{4}{@{}l}{\emph{Wrong-object errors}}\\
SPRED added to the probe on correctness & \hqt{0.036}{\relax[0.017, 0.056]}{\relax(0.025, 0.048)} & \hqtt{0.026}{\relax(0.042, 0.011)} & \hqtt{0.035}{\relax(0.033, 0.036)}\\ \addlinespace
\quad configuration searched again & \hqt{0.044}{\relax[0.021, 0.067]}{\relax(0.025, 0.063)} & \hqtt{0.029}{\relax(0.042, 0.016)} & \hqtt{0.035}{\relax(0.033, 0.037)}\\ \addlinespace
Adding SPRED minus adding a second probe$^{\dagger}$ & \hqt{0.039}{\relax[0.023, 0.054]}{\relax(0.030, 0.047)} & \hqt{0.022}{\relax[0.005, 0.039]}{\relax(0.043, 0.001)} & \hqt{0.028}{\relax[0.015, 0.042]}{\relax(0.031, 0.025)}\\ \addlinespace
\quad both probes searched again$^{\dagger}$ & \hqt{0.031}{\relax[0.003, 0.059]}{\relax(0.030, 0.032)} & \hqt{0.035}{\relax[0.006, 0.064]}{\relax(0.043, 0.027)} & \hqt{0.026}{\relax[0.005, 0.047]}{\relax(0.031, 0.020)}\\ \addlinespace
SPRED minus the probe on correctness$^{\dagger}$ & \hqt{0.059}{\relax[0.006, 0.111]}{\relax(0.021, 0.097)} & \hqtt{0.033}{\relax(0.059, 0.007)} & \hqtt{0.044}{\relax(0.040, 0.048)}\\ \addlinespace
\bottomrule
\end{tabular}
\end{table}

%% file: tables/bicrtenk.tex
\begin{table}[!htbp]
\centering
\caption{SPRED against BICR trained on the labels of 10,000 training items (9,997 for \qlarge{}), \qsmall{} and \qlarge{}: SPRED minus BICR in AUROC, SPRED fitted on the same items without their labels. Each cell gives the mean over the two models and the 95\% interval combining image resampling with ten refits on the same redrawn items for both scores. BICR's configuration was chosen among 40 drawn at random from the published search space.}
\label{tab:bicrtenk}%
\setlength{\tabcolsep}{4pt}
\begin{tabular}{@{}lccc@{}}
\toprule
& Replication list & Held-out list & New items\\
\midrule
All errors & \hqtt{0.040}{\relax[0.010, 0.071]} & \hqtt{0.042}{\relax[0.011, 0.072]} & \hqtt{0.029}{\relax[0.003, 0.055]}\\ \addlinespace
Wrong-object errors & \hqtt{0.043}{\relax[0.008, 0.079]} & \hqtt{0.018}{\relax[$-$0.022, 0.057]} & \hqtt{0.041}{\relax[0.008, 0.075]}\\
\bottomrule
\end{tabular}
\end{table}

%% file: tables/weights.tex
\begin{table}[!htbp]
\centering
\caption{Sensitivity to the weight $w$ of the residual in $z(g)+w\,z(u)$: AUROC for all errors. $w=0$ is the probe alone; $w=1$ the equal-weight sum. The weight 0.5 was chosen while examining the six sets of \qsmall{} (which include its new items) and the new items of \qfour{}.}
\label{tab:weights}
\scriptsize
\setlength{\tabcolsep}{3pt}
\begin{tabular}{lrcccccc}
\toprule
Test set & Items & $w=0$ & 0.25 & 0.5 & 0.75 & 1 & 2\\
\midrule
\qsmall{}, six sets & 3{,}352 & 0.813 & 0.828 & 0.831 & 0.828 & 0.822 & 0.802\\
\qsmall{}, new items & 1{,}152 & 0.809 & 0.826 & 0.829 & 0.826 & 0.820 & 0.797\\
\qsmall{}, RefCOCO+ & 1{,}200 & 0.805 & 0.825 & 0.830 & 0.828 & 0.825 & 0.810\\
\qlarge{}, new items & 1{,}152 & 0.833 & 0.847 & 0.848 & 0.843 & 0.835 & 0.806\\
\qlarge{}, RefCOCO+ & 1{,}200 & 0.791 & 0.815 & 0.822 & 0.821 & 0.816 & 0.798\\
\qfour{}, new items & 1{,}151 & 0.831 & 0.851 & 0.858 & 0.858 & 0.854 & 0.836\\
\bottomrule
\end{tabular}
\end{table}

%% file: tables/sens.tex
\begin{table}[!htbp]
\centering
\caption{Sensitivity of the detector on the new items (AUROC for all errors). One setting at a time is moved from its value chosen without labels on the training items (\cref{sec:method}), the others kept; for the weight $w$ the chosen value is 0.5. Best minus chosen: largest gain over the stored layers or over penalties $10^1$ to $10^6$. Ten times smaller or larger: the larger of the two drops.}
\label{tab:sens}
\footnotesize
\setlength{\tabcolsep}{2.5pt}
\begin{tabular}{@{}>{\raggedright\arraybackslash}p{3.3cm}ccccc@{}}
\toprule
 & \hqsmall & \hqlarge & \hqfour & \hqeight & \hqivl\\
\midrule
Detector at the label-free choice & 0.829 & 0.848 & 0.858 & 0.837 & 0.867\\
\midrule
$w=0$ (probe alone) & 0.809 & 0.833 & 0.831 & 0.820 & 0.850\\
$w=0.25$ & 0.826 & 0.847 & 0.851 & 0.833 & 0.863\\
$w=1$ & 0.820 & 0.835 & 0.854 & 0.831 & 0.858\\
$w=2$ & 0.797 & 0.806 & 0.836 & 0.816 & 0.835\\
\midrule
Probe layer: best minus chosen & 0.000 & 0.000 & 0.000 & 0.013 & 0.007\\
Probe penalty: best minus chosen & 0.003 & 0.000 & 0.000 & 0.003 & 0.000\\
\quad penalty ten times smaller or larger & $-$0.024 & $-$0.041 & $-$0.021 & $-$0.009 & $-$0.014\\
Read-out layer: best minus chosen & 0.001 & 0.000 & 0.000 & 0.002 & 0.000\\
Read-out penalty: best minus chosen & 0.000 & 0.000 & 0.000 & 0.000 & 0.000\\
\quad penalty ten times smaller or larger & $-$0.006 & $-$0.010 & $-$0.008 & $-$0.009 & $-$0.009\\
\bottomrule
\end{tabular}
\end{table}

%% file: tables/ltt.tex
\begin{table}[!htbp]
\centering
\caption{Simulation of thresholds by fixed-sequence testing ($\delta=0.1$): 41 candidate thresholds fixed in advance as quantiles of the detector score on the unlabelled training items, 1,000 random splits by image into calibration and evaluation items. Found: share of splits in which a threshold is certified; kept: mean share of items kept (0 when none is certified); error: mean error rate of the kept items; exceeded: share of splits in which the error rate of the selected threshold on the whole test set exceeds $\alpha$, which the guarantee bounds by $\delta$.}
\label{tab:ltt}
\footnotesize
\setlength{\tabcolsep}{3pt}
\resizebox{\textwidth}{!}{%
\begin{tabular}{lllcccc}
\toprule
Test set & $\alpha$ & $m$ & Found & Kept & Error & Exceeded\\
\midrule
\qsmall{}, six sets & 8\% & 200 & 68\% & 38\% & 4.1\% & 3.7\%\\
\qsmall{}, six sets & 8\% & 400 & 91\% & 64\% & 5.6\% & 6.1\%\\
\qsmall{}, six sets & 10\% & 200 & 79\% & 55\% & 6.0\% & 3.1\%\\
\qsmall{}, six sets & 10\% & 400 & 98\% & 81\% & 7.5\% & 2.1\%\\
\qlarge{}, new items & 10\% & 200 & 62\% & 41\% & 5.8\% & 3.5\%\\
\qlarge{}, new items & 10\% & 400 & 100\% & 82\% & 7.4\% & 2.6\%\\
\qlarge{}, RefCOCO+ & 10\% & 200 & 100\% & 72\% & 6.1\% & 2.7\%\\
\qlarge{}, RefCOCO+ & 10\% & 400 & 100\% & 82\% & 7.5\% & 1.2\%\\
\qfour{}, new items & 10\% & 200 & 85\% & 71\% & 6.6\% & 7.1\%\\
\qfour{}, new items & 10\% & 400 & 100\% & 89\% & 7.5\% & 4.0\%\\
\bottomrule
\end{tabular}}
\end{table}

%% file: tables/detect.tex
\begin{table}[!t]
\centering
\caption{AUROC for localization errors (IoU $<0.5$). Second header row: items / errors. $^{a}$Examined while the detector was designed (\cref{sec:setup}); the six sets include the new items. $^{d}$The same score as the sampling-teacher probe + 0.5 $\times$ residual further down. All scores use one forward pass of one model except the last two rows (the content-shift check, two passes; the box of another model, a second model). $^{\ast}$Supervised with the gold boxes of the 2,000 training items. $^{\ddagger}$Computed separately on the new items. Bold: highest AUROC in each column among the scores of one pass of one model, the linear probe combined with the detector excepted. Dashes in the \qeight{}, \qivl{} and \qllava{} columns: not computed for these models; \qllava: paired differences with intervals in the text; in the pre-generation row of \qlarge: identical to the correctness probe. $^{b}$With the mapping of the original paper, recomputed in a middle band of layers fixed per model family in advance; not defined for \qivl. $^{c}$Disagreement with the box of \qeight{} (of \qlarge{} for \qeight), standardized and added with equal weight~\cite{ax2609.37488}. ---: not computed for that set.}
\label{tab:detect}
\scriptsize
\setlength{\tabcolsep}{2.5pt}
\resizebox{\textwidth}{!}{%
\begin{tabular}{lccccccccc}
\toprule
 & \multicolumn{3}{c}{\qsmall} & \multicolumn{2}{c}{\qlarge} & \qfour & \qeight & \qivl & \qllava\\
\cmidrule(lr){2-4}\cmidrule(lr){5-6}\cmidrule(lr){7-7}\cmidrule(lr){8-8}\cmidrule(lr){9-9}\cmidrule(lr){10-10}
Score & Six sets$^{a}$ & New$^{a}$ & RefCOCO+ & New & RefCOCO+ & New$^{a}$ & New & New & New\\
 & 3{,}352 / 456 & 1{,}152 / 168 & 1{,}200 / 183 & 1{,}152 / 160 & 1{,}200 / 152 & 1{,}151 / 138 & 1{,}149 / 126 & 1{,}152 / 138 & 1{,}152 / 424\\
\midrule
Ours: probe + 0.5 $\times$ residual & \textbf{0.831} & \textbf{0.829} & 0.830 & \textbf{0.848} & 0.822 & \textbf{0.858} & 0.837 & 0.867 & 0.790\\
\quad two-teacher probe only & 0.813 & 0.809 & 0.805 & 0.833 & 0.791 & 0.831 & 0.820 & 0.850 & 0.780\\
\quad sampling teacher only in the probe$^{d}$ & 0.814 & 0.803 & 0.823 & 0.814 & 0.828 & 0.856 & 0.825 & 0.858 & 0.797\\
\quad self-prediction residual only & 0.740 & 0.729 & 0.763 & 0.725 & 0.752 & 0.773 & 0.760 & 0.749 & 0.729\\
\midrule
Coordinate log-prob & 0.680 & 0.719 & 0.677 & 0.715 & 0.634 & 0.676 & 0.649 & 0.587 & 0.658\\
MTLA (original mapping)$^{b}$~\cite{ax2607.05978} & --- & 0.606 & 0.663 & 0.600 & 0.654 & 0.593 & 0.636 & --- & ---\\
Whole-image attention & 0.463 & 0.474 & 0.433 & 0.439 & 0.423 & 0.496 & 0.502 & --- & ---\\
Mahalanobis~\cite{lee2018mahalanobis} & 0.737 & 0.720 & 0.698 & 0.635 & 0.732 & 0.670 & 0.784 & 0.649 & 0.670\\
Digit-weighted coordinate uncertainty$^{\ddagger}$ (cf.~\cite{li2026pace}) & --- & 0.782 & --- & 0.779 & --- & 0.687 & --- & --- & ---\\
Sampling-teacher probe~\cite{kossen2024sep,ax2604.19444} & 0.800 & 0.780 & 0.803 & 0.802 & 0.819 & 0.824 & 0.791 & 0.836 & 0.786\\
\midrule
Linear probe, correctness$^{\ast}$~\cite{kossen2024sep} & 0.819 & 0.820 & 0.823 & 0.802 & 0.810 & 0.835 & 0.835 & 0.876 & 0.789\\
Linear probe, $1-\mathrm{IoU}^{\ast}$ & 0.818 & 0.814 & \textbf{0.834} & 0.800 & 0.811 & 0.831 & 0.821 & 0.870 & 0.794\\
SAPLMA$^{\ast}$~\cite{azaria2023saplma} & 0.796 & 0.790 & 0.809 & 0.749 & 0.739 & 0.803 & 0.784 & 0.809 & 0.752\\
Pre-generation probe$^{\ast}$~\cite{ax2603.05465} & 0.815 & 0.805 & 0.809 & --- & --- & 0.833 & 0.823 & 0.792 & 0.747\\
Box geometry$^{\ast}$ & 0.626 & 0.593 & 0.629 & 0.601 & 0.613 & 0.592 & 0.619 & 0.582 & 0.693\\
\midrule
Sampling-teacher probe + 0.5 $\times$ residual & 0.814 & 0.803 & 0.823 & 0.814 & \textbf{0.828} & 0.856 & 0.825 & 0.858 & \textbf{0.797}\\
Linear probe (correctness)$^{\ast}$ + 0.5 $\times$ residual & 0.828 & 0.826 & \textbf{0.834} & 0.818 & 0.824 & 0.846 & \textbf{0.842} & \textbf{0.886} & \textbf{0.797}\\
Linear probe (correctness)$^{\ast}$ + detector & --- & 0.840 & --- & 0.858 & --- & 0.864 & 0.850 & 0.887 & ---\\
\midrule
Content-shift check (2 passes) & 0.748 & 0.765 & 0.750 & 0.787 & 0.718 & 0.690 & 0.691 & 0.778 & 0.708\\
Detector + box of another model$^{c}$ (2 models) & --- & 0.875 & --- & 0.877 & --- & 0.876 & 0.869 & 0.906 & ---\\
\bottomrule
\end{tabular}}
\end{table}

%% file: tables/bytypefull.tex
\begin{table}[!htbp]
\centering
\caption{Gain in AUROC when the residual is added with weight 0.5 (\cref{eq:combine}) to each single-pass score, by error type, replication list (all error types for the scores of \cref{tab:plugingain}). W, P, E: wrong-object, partial-overlap and wrong-extent errors; All: all errors. Each type is scored against the correct answers; all errors against all other answers. Cell: mean over the models of the stored gain. Bold: combined interval, stored gain $\pm 1.96\sqrt{\mathrm{SE}^2+\mathrm{SD}^2}$ with SE over images and SD over ten refits on training items redrawn with replacement, excludes zero. Last rows: mean AUROC without and with the residual. The held-out list in \cref{tab:bytypeho}; per-model cells with both kinds of interval, and the new items, in \cref{sup:plugin}.}
\label{tab:bytypefull}
\setlength{\tabcolsep}{3pt}
\begin{tabular}{@{}lcccc@{}}
\toprule
Score & W & P & E & All\\
\midrule
\multicolumn{5}{@{}l}{\emph{Label-free probes}}\\
\quad two-teacher (detector base) & \textbf{0.042} & \textbf{$-$0.018} & $-$0.016 & 0.007\\
\quad sampling teacher & \textbf{0.036} & $-$0.011 & $-$0.011 & 0.008\\
\quad shift teacher & \textbf{0.035} & $-$0.016 & $-$0.010 & 0.006\\
\multicolumn{5}{@{}l}{\emph{Supervised, 2,000 labels}}\\
\quad linear, correctness & \textbf{0.037} & \textbf{$-$0.017} & $-$0.018 & 0.005\\
\quad linear, $1-\mathrm{IoU}$ & \textbf{0.029} & \textbf{$-$0.019} & \textbf{$-$0.020} & 0.000\\
\quad pre-generation & \textbf{0.043} & $-$0.008 & 0.001 & \textbf{0.015}\\
\quad SAPLMA perceptron & \textbf{0.049} & $-$0.002 & $-$0.007 & \textbf{0.017}\\
\quad linear, wrong-object labels & \textbf{0.026} & $-$0.003 & 0.007 & \textbf{0.012}\\
\quad \quad same, W and correct items only & \textbf{0.026} & $-$0.008 & 0.002 & 0.009\\
\multicolumn{5}{@{}l}{\emph{Training-free and geometry}}\\
\quad coordinate log-prob & \textbf{0.117} & \textbf{0.036} & \textbf{0.021} & \textbf{0.063}\\
\quad in-box attention (4 models) & \textbf{0.089} & \textbf{0.043} & \textbf{0.029} & \textbf{0.058}\\
\quad whole-image attention (4 models) & \textbf{0.175} & \textbf{0.063} & \textbf{0.045} & \textbf{0.106}\\
\quad Mahalanobis distance & \textbf{0.120} & 0.015 & 0.013 & \textbf{0.056}\\
\quad box geometry, supervised & \textbf{0.074} & \textbf{0.032} & \textbf{0.019} & \textbf{0.044}\\
\quad box geometry, teacher & \textbf{0.072} & \textbf{0.024} & \textbf{0.015} & \textbf{0.040}\\
\midrule
Residual alone & 0.808 & 0.642 & 0.602 & 0.695\\
Two-teacher probe alone & 0.788 & 0.824 & 0.768 & 0.789\\
Detector (probe + residual) & 0.829 & 0.806 & 0.752 & 0.796\\
\bottomrule
\end{tabular}
\end{table}

%% file: tables/bytypeho.tex
\begin{table}[!t]
\centering
\caption{Gain in AUROC when the residual is added with weight 0.5 (\cref{eq:combine}) to each single-pass score, by error type, held-out list. Each type is scored against the correct answers; all errors against all other answers. Cell: mean over the models of the stored gain; in parentheses: models with a positive gain of models where the score is defined. Bold: combined interval, stored gain $\pm 1.96\sqrt{\mathrm{SE}^2+\mathrm{SD}^2}$ with SE over images and SD over ten refits on training items redrawn with replacement, excludes zero. Last rows: mean AUROC without and with the residual. The replication list in \cref{tab:bytypefull}; per-model cells with both kinds of interval, and the new items, in \cref{sup:plugin}.}
\label{tab:bytypeho}
\scriptsize
\setlength{\tabcolsep}{4pt}
\begin{tabular}{@{}lcccc@{}}
\toprule
Score & Wrong object & Partial overlap & Wrong extent & All errors\\
\midrule
\multicolumn{5}{@{}l}{\emph{Label-free probes}}\\
\quad two-teacher (base of the detector) & \textbf{0.044} (6/6) & $-$0.008 (2/6) & $-$0.012 (0/6) & \textbf{0.012} (5/6)\\
\quad sampling teacher & \textbf{0.050} (6/6) & $-$0.001 (4/6) & $-$0.005 (1/6) & \textbf{0.019} (6/6)\\
\quad shift teacher & \textbf{0.037} (6/6) & 0.006 (4/6) & $-$0.010 (1/6) & \textbf{0.013} (5/6)\\
\multicolumn{5}{@{}l}{\emph{Supervised, 2,000 labels}}\\
\quad linear, correctness & \textbf{0.037} (6/6) & $-$0.012 (2/6) & $-$0.014 (1/6) & 0.007 (4/6)\\
\quad linear, $1-\mathrm{IoU}$ & \textbf{0.034} (6/6) & $-$0.010 (2/6) & $-$0.012 (0/6) & 0.007 (4/6)\\
\quad pre-generation & \textbf{0.046} (6/6) & 0.004 (4/6) & $-$0.004 (4/6) & \textbf{0.018} (6/6)\\
\quad SAPLMA perceptron & \textbf{0.050} (6/6) & 0.007 (3/6) & $-$0.004 (2/6) & \textbf{0.021} (6/6)\\
\quad linear, wrong-object labels & \textbf{0.029} (6/6) & 0.008 (4/6) & 0.009 (4/6) & \textbf{0.017} (6/6)\\
\quad \quad same, wrong-object and correct items & \textbf{0.029} (6/6) & 0.005 (4/6) & 0.006 (4/6) & \textbf{0.015} (6/6)\\
\multicolumn{5}{@{}l}{\emph{Training-free and geometry}}\\
\quad coordinate log-prob & \textbf{0.131} (6/6) & \textbf{0.047} (6/6) & \textbf{0.035} (6/6) & \textbf{0.075} (6/6)\\
\quad in-box attention (4 models) & \textbf{0.106} (4/4) & \textbf{0.070} (4/4) & \textbf{0.044} (4/4) & \textbf{0.076} (4/4)\\
\quad whole-image attention (4 models) & \textbf{0.173} (4/4) & \textbf{0.089} (4/4) & \textbf{0.086} (4/4) & \textbf{0.122} (4/4)\\
\quad Mahalanobis distance & \textbf{0.104} (6/6) & 0.041 (5/6) & 0.015 (6/6) & \textbf{0.057} (6/6)\\
\quad box geometry, supervised & \textbf{0.081} (6/6) & \textbf{0.044} (6/6) & \textbf{0.018} (6/6) & \textbf{0.050} (6/6)\\
\quad box geometry, teacher & \textbf{0.077} (6/6) & \textbf{0.036} (6/6) & \textbf{0.016} (5/6) & \textbf{0.046} (6/6)\\
\midrule
Residual alone & 0.820 & 0.692 & 0.637 & 0.723\\
Two-teacher probe alone & 0.799 & 0.826 & 0.781 & 0.797\\
Detector (probe + residual) & 0.843 & 0.818 & 0.769 & 0.808\\
\bottomrule
\end{tabular}
\end{table}

%% file: tables/gaindiff.tex
\begin{table}[!htbp]
\centering
\caption{Gain on wrong-object errors minus gain on the other two error types when the residual is added with weight 0.5 to each single-pass score. W, P, E: gain in AUROC on wrong-object, partial-overlap and wrong-extent errors, each scored against the correct answers. The difference is taken within each image resample and each refit, then averaged over the models (post hoc); in small type, the combined interval over images and ten refits on redrawn training items. Seven probes: the label-free and supervised probes of the first two groups without the two trained on wrong-object labels.}
\label{tab:gaindiff}
\scriptsize
\setlength{\tabcolsep}{1.5pt}
\begin{tabular}{@{}lcccc@{}}
\toprule
& \multicolumn{2}{c}{Replication list} & \multicolumn{2}{c}{Held-out list}\\
\cmidrule(lr){2-3}\cmidrule(l){4-5}
Score & W $-$ P & W $-$ E & W $-$ P & W $-$ E\\
\midrule
\multicolumn{5}{@{}l}{\emph{Label-free probes}}\\
\quad two-teacher (base of the detector) & 0.060 {\tiny[0.040, 0.080]} & 0.058 {\tiny[0.034, 0.081]} & 0.053 {\tiny[0.030, 0.075]} & 0.056 {\tiny[0.034, 0.078]}\\
\quad sampling teacher & 0.047 {\tiny[0.027, 0.066]} & 0.047 {\tiny[0.023, 0.072]} & 0.050 {\tiny[0.026, 0.075]} & 0.055 {\tiny[0.031, 0.078]}\\
\quad shift teacher & 0.051 {\tiny[0.029, 0.074]} & 0.045 {\tiny[0.022, 0.069]} & 0.031 {\tiny[0.008, 0.054]} & 0.047 {\tiny[0.026, 0.067]}\\
\multicolumn{5}{@{}l}{\emph{Supervised, 2,000 labels}}\\
\quad linear, correctness & 0.054 {\tiny[0.034, 0.074]} & 0.055 {\tiny[0.033, 0.076]} & 0.049 {\tiny[0.027, 0.070]} & 0.051 {\tiny[0.030, 0.072]}\\
\quad linear, $1-\mathrm{IoU}$ & 0.048 {\tiny[0.029, 0.067]} & 0.049 {\tiny[0.026, 0.072]} & 0.044 {\tiny[0.024, 0.064]} & 0.046 {\tiny[0.024, 0.069]}\\
\quad pre-generation & 0.051 {\tiny[0.030, 0.071]} & 0.042 {\tiny[0.019, 0.065]} & 0.042 {\tiny[0.021, 0.063]} & 0.050 {\tiny[0.029, 0.071]}\\
\quad SAPLMA perceptron & 0.051 {\tiny[0.027, 0.074]} & 0.055 {\tiny[0.027, 0.084]} & 0.042 {\tiny[0.019, 0.066]} & 0.053 {\tiny[0.030, 0.076]}\\
\quad linear, wrong-object labels & 0.029 {\tiny[0.006, 0.051]} & 0.019 {\tiny[$-$0.005, 0.042]} & 0.022 {\tiny[0.000, 0.043]} & 0.021 {\tiny[$-$0.001, 0.043]}\\
\quad \quad same, wrong-object and correct items & 0.034 {\tiny[0.017, 0.052]} & 0.024 {\tiny[0.005, 0.044]} & 0.024 {\tiny[0.002, 0.045]} & 0.023 {\tiny[0.002, 0.044]}\\
\multicolumn{5}{@{}l}{\emph{Training-free and geometry}}\\
\quad coordinate log-prob & 0.081 {\tiny[0.052, 0.110]} & 0.096 {\tiny[0.070, 0.122]} & 0.084 {\tiny[0.055, 0.112]} & 0.096 {\tiny[0.068, 0.124]}\\
\quad in-box attention (4 models) & 0.046 {\tiny[0.018, 0.074]} & 0.060 {\tiny[0.031, 0.089]} & 0.037 {\tiny[0.007, 0.067]} & 0.062 {\tiny[0.031, 0.094]}\\
\quad whole-image attention (4 models) & 0.112 {\tiny[0.071, 0.154]} & 0.130 {\tiny[0.091, 0.169]} & 0.083 {\tiny[0.042, 0.125]} & 0.086 {\tiny[0.045, 0.128]}\\
\quad Mahalanobis distance & 0.105 {\tiny[0.074, 0.136]} & 0.107 {\tiny[0.075, 0.139]} & 0.063 {\tiny[0.033, 0.093]} & 0.090 {\tiny[0.053, 0.127]}\\
\quad box geometry, supervised & 0.042 {\tiny[0.016, 0.068]} & 0.055 {\tiny[0.031, 0.079]} & 0.037 {\tiny[0.015, 0.059]} & 0.063 {\tiny[0.042, 0.083]}\\
\quad box geometry, teacher & 0.048 {\tiny[0.025, 0.071]} & 0.057 {\tiny[0.036, 0.078]} & 0.041 {\tiny[0.020, 0.062]} & 0.061 {\tiny[0.040, 0.082]}\\
\midrule
Mean, seven probes & 0.052 {\tiny[0.034, 0.069]} & 0.050 {\tiny[0.030, 0.070]} & 0.044 {\tiny[0.025, 0.064]} & 0.051 {\tiny[0.032, 0.070]}\\
Mean, fifteen scores & 0.055 {\tiny[0.038, 0.073]} & 0.058 {\tiny[0.039, 0.076]} & 0.046 {\tiny[0.028, 0.064]} & 0.056 {\tiny[0.038, 0.075]}\\
Interval above zero (of 15) & 15 & 14 & 15 & 14\\
\bottomrule
\end{tabular}
\end{table}

%% file: tables/bytypenewitems.tex
\begin{table}[!htbp]
\centering
\caption{Gain in AUROC when the residual is added with weight 0.5 (\cref{eq:combine}) to each single-pass score, by error type, new items (552 RefCOCO testB and 600 RefCOCOg validation items). Each type is scored against the correct answers; all errors against all other answers. Cell: mean over the models of the stored gain; small: models with a positive gain of models where the score is defined. Bold: combined interval, stored gain $\pm 1.96\sqrt{\mathrm{SE}^2+\mathrm{SD}^2}$ with SE over images and SD over ten refits on training items redrawn with replacement, excludes zero. Last rows: mean AUROC without and with the residual.}
\label{tab:bytypenewitems}
\scriptsize
\setlength{\tabcolsep}{4pt}
\begin{tabular}{@{}lcccc@{}}
\toprule
Score & Wrong object & Partial overlap & Wrong extent & All errors\\
\midrule
\multicolumn{5}{@{}l}{\emph{Label-free probes}}\\
\quad two-teacher (base of the detector) & \textbf{0.053} {\tiny6/6} & 0.009 {\tiny4/6} & $-$0.006 {\tiny0/6} & \textbf{0.018} {\tiny6/6}\\
\quad sampling teacher & \textbf{0.055} {\tiny6/6} & \textbf{0.017} {\tiny6/6} & $-$0.001 {\tiny3/6} & \textbf{0.022} {\tiny6/6}\\
\quad shift teacher & \textbf{0.051} {\tiny6/6} & 0.013 {\tiny5/6} & $-$0.000 {\tiny3/6} & \textbf{0.021} {\tiny6/6}\\
\multicolumn{5}{@{}l}{\emph{Supervised, 2,000 labels}}\\
\quad linear, correctness & \textbf{0.040} {\tiny6/6} & 0.004 {\tiny4/6} & $-$0.010 {\tiny1/6} & 0.010 {\tiny6/6}\\
\quad linear, $1-\mathrm{IoU}$ & \textbf{0.038} {\tiny6/6} & 0.005 {\tiny4/6} & $-$0.006 {\tiny1/6} & \textbf{0.011} {\tiny6/6}\\
\quad pre-generation & \textbf{0.052} {\tiny6/6} & 0.016 {\tiny5/6} & $-$0.005 {\tiny2/6} & \textbf{0.019} {\tiny6/6}\\
\quad SAPLMA perceptron & \textbf{0.063} {\tiny6/6} & \textbf{0.028} {\tiny6/6} & 0.007 {\tiny4/6} & \textbf{0.031} {\tiny6/6}\\
\quad linear, wrong-object labels & \textbf{0.032} {\tiny6/6} & \textbf{0.026} {\tiny6/6} & 0.011 {\tiny4/6} & \textbf{0.022} {\tiny6/6}\\
\quad \quad same, wrong-object and correct items & \textbf{0.030} {\tiny6/6} & \textbf{0.018} {\tiny6/6} & 0.004 {\tiny4/6} & \textbf{0.016} {\tiny6/6}\\
\multicolumn{5}{@{}l}{\emph{Training-free and geometry}}\\
\quad coordinate log-prob & \textbf{0.146} {\tiny6/6} & \textbf{0.063} {\tiny6/6} & \textbf{0.038} {\tiny6/6} & \textbf{0.080} {\tiny6/6}\\
\quad in-box attention (4 models) & \textbf{0.104} {\tiny4/4} & \textbf{0.078} {\tiny4/4} & \textbf{0.046} {\tiny4/4} & \textbf{0.074} {\tiny4/4}\\
\quad whole-image attention (4 models) & \textbf{0.146} {\tiny4/4} & \textbf{0.111} {\tiny4/4} & \textbf{0.074} {\tiny4/4} & \textbf{0.108} {\tiny4/4}\\
\quad Mahalanobis distance & \textbf{0.119} {\tiny6/6} & \textbf{0.059} {\tiny6/6} & 0.023 {\tiny4/6} & \textbf{0.064} {\tiny6/6}\\
\quad box geometry, supervised & \textbf{0.097} {\tiny6/6} & \textbf{0.062} {\tiny6/6} & \textbf{0.036} {\tiny6/6} & \textbf{0.063} {\tiny6/6}\\
\quad box geometry, teacher & \textbf{0.097} {\tiny6/6} & \textbf{0.055} {\tiny6/6} & \textbf{0.036} {\tiny6/6} & \textbf{0.061} {\tiny6/6}\\
\midrule
Residual alone & 0.863 & 0.753 & 0.649 & 0.744\\
Two-teacher probe alone & 0.825 & 0.861 & 0.802 & 0.820\\
Detector (probe + residual) & 0.878 & 0.870 & 0.796 & 0.838\\
\bottomrule
\end{tabular}
\end{table}

%% file: tables/plugci.tex
\begin{sidewaystable}
\centering
\caption{Gain for wrong-object errors when the residual is added with weight 0.5, mean over the models, with two intervals that cover randomness on the training side. Stored: the fit used throughout, with the combined interval, stored value $\pm 1.96\sqrt{\mathrm{SE}^2+\mathrm{SD}^2}$ (SE over images, SD over ten refits on training items drawn with replacement, base and residual refitted together). Refit mean: mean over the ten refits, with the nested interval, 2.5th and 97.5th percentiles over image resamples each paired with one refit. About 63\% of a redrawn training set are distinct items, and the refit mean is mostly slightly above the stored value. In \qlarge{} the selected correctness probe reads the pre-generation state, so these two scores are one fit and their cell is counted once.}
\label{tab:plugci}
\scriptsize
\setlength{\tabcolsep}{3pt}
\begin{tabular}{@{}lcccccc@{}}
\toprule
& \multicolumn{3}{c}{New items} & \multicolumn{3}{c}{Held-out list}\\
\cmidrule(lr){2-4}\cmidrule(lr){5-7}
Score & Stored & Combined & Refit mean, nested & Stored & Combined & Refit mean, nested\\
\midrule
two-teacher probe & 0.053 & [0.037, 0.069] & 0.057 [0.042, 0.075] & 0.044 & [0.030, 0.059] & 0.047 [0.033, 0.063]\\
sampling-teacher probe & 0.055 & [0.036, 0.075] & 0.064 [0.046, 0.085] & 0.050 & [0.031, 0.069] & 0.055 [0.038, 0.076]\\
shift-teacher probe & 0.051 & [0.038, 0.064] & 0.057 [0.043, 0.072] & 0.037 & [0.021, 0.052] & 0.043 [0.027, 0.060]\\
linear, correctness & 0.040 & [0.022, 0.057] & 0.047 [0.032, 0.067] & 0.037 & [0.020, 0.054] & 0.043 [0.026, 0.058]\\
linear, $1-\mathrm{IoU}$ & 0.038 & [0.024, 0.051] & 0.043 [0.030, 0.059] & 0.034 & [0.019, 0.049] & 0.038 [0.024, 0.055]\\
pre-generation & 0.052 & [0.036, 0.068] & 0.053 [0.039, 0.071] & 0.046 & [0.030, 0.062] & 0.048 [0.032, 0.063]\\
SAPLMA perceptron & 0.063 & [0.043, 0.082] & 0.070 [0.051, 0.090] & 0.050 & [0.032, 0.067] & 0.062 [0.043, 0.079]\\
linear, wrong-object labels & 0.032 & [0.018, 0.046] & 0.045 [0.032, 0.060] & 0.029 & [0.012, 0.047] & 0.042 [0.026, 0.060]\\
\quad same, wrong-object and correct items & 0.030 & [0.017, 0.043] & 0.041 [0.028, 0.055] & 0.029 & [0.014, 0.044] & 0.038 [0.023, 0.054]\\
coordinate log-prob & 0.146 & [0.125, 0.167] & 0.140 [0.119, 0.162] & 0.131 & [0.110, 0.152] & 0.125 [0.105, 0.145]\\
in-box attention (4 models) & 0.104 & [0.079, 0.130] & 0.099 [0.076, 0.125] & 0.106 & [0.084, 0.128] & 0.100 [0.080, 0.123]\\
whole-image attention (4 models) & 0.146 & [0.121, 0.171] & 0.141 [0.115, 0.165] & 0.173 & [0.146, 0.199] & 0.164 [0.140, 0.189]\\
Mahalanobis distance & 0.119 & [0.071, 0.166] & 0.133 [0.093, 0.177] & 0.104 & [0.063, 0.145] & 0.113 [0.077, 0.148]\\
box geometry, supervised & 0.097 & [0.077, 0.117] & 0.095 [0.073, 0.114] & 0.081 & [0.064, 0.098] & 0.078 [0.062, 0.094]\\
box geometry, teacher & 0.097 & [0.079, 0.115] & 0.096 [0.078, 0.113] & 0.077 & [0.061, 0.093] & 0.076 [0.061, 0.093]\\
\midrule
Distinct cells with a positive stored gain & \multicolumn{3}{c}{85} & \multicolumn{3}{c}{85}\\
\quad combined interval above zero & \multicolumn{3}{c}{79} & \multicolumn{3}{c}{59}\\
\quad nested interval above zero & \multicolumn{3}{c}{85} & \multicolumn{3}{c}{76}\\
\bottomrule
\end{tabular}
\end{sidewaystable}

%% file: tables/addsig.tex
\begin{table}[!htbp]
\centering
\caption{The residual against other label-free signals of the same pass, added under the same rule (\cref{eq:combine}) to the fifteen single-pass scores; wrong-object errors against correct answers. Gain with the residual minus gain with the alternative signal, averaged over scores and models (stored fit, combined interval as in \cref{tab:plugci}); a score is never added to itself, and each comparison uses the models on which both signals exist. Scores: number of scores whose six-model mean difference has a combined interval above zero, of 14. With the weight 0.25 or 1 for both signals the sign of every difference is unchanged.}
\label{tab:addsig}
\footnotesize
\setlength{\tabcolsep}{3pt}
\begin{tabular}{@{}lcccc@{}}
\toprule
& \multicolumn{2}{c}{New items} & \multicolumn{2}{c}{Held-out list}\\
\cmidrule(lr){2-3}\cmidrule(lr){4-5}
Alternative signal & Difference & Scores & Difference & Scores\\
\midrule
coordinate log-prob & 0.065 [0.048, 0.082] & 14 & 0.060 [0.041, 0.078] & 14\\
Mahalanobis distance & 0.057 [0.042, 0.072] & 14 & 0.052 [0.035, 0.069] & 14\\
in-box attention (4 models) & 0.045 [0.030, 0.060] & 14 & 0.057 [0.036, 0.078] & 14\\
attention on the image (4 models) & 0.075 [0.049, 0.102] & 14 & 0.086 [0.063, 0.109] & 14\\
box geometry, teacher & 0.059 [0.041, 0.077] & 14 & 0.057 [0.038, 0.076] & 14\\
two-teacher probe & 0.024 [0.010, 0.039] & 10 & 0.022 [0.007, 0.037] & 9\\
\bottomrule
\end{tabular}
\end{table}

%% file: tables/equalinfo.tex
\begin{table}[!htbp]
\centering
\caption{Supervised probes given more of the information the residual uses, trained on the 2,000 labelled training items, wrong-object errors, mean over the six models. Two positions: the states of the last prompt token and of the coordinate tokens of one layer, concatenated. Boxes and IoU: the read-out box (out of fold on the training items), the written box and their IoU, nine inputs. Perceptron: one hidden layer. Gain: AUROC change when the residual is added with weight 0.5. Detector minus probe with the one-sided 95\% lower bound in parentheses. Stored fit. Ridge and logistic rows: combined intervals and bounds over images (2,000 resamples) and ten training refits; perceptron rows: images only.}
\label{tab:equalinfo}
\scriptsize
\setlength{\tabcolsep}{2.5pt}
\resizebox{\textwidth}{!}{%
\begin{tabular}{@{}lccccc@{}}
\toprule
Probe & AUROC, held-out / new & Gain, held-out & Gain, new items & Detector $-$ probe, held-out & Detector $-$ probe, new items\\
\midrule
two positions, ridge, correctness & 0.821 / 0.853 & 0.032 {\tiny[0.015, 0.049]} & 0.039 {\tiny[0.025, 0.053]} & 0.022 ($-$0.012) & 0.025 (0.003)\\
two positions, ridge, wrong-object labels & 0.820 / 0.860 & 0.031 {\tiny[0.018, 0.045]} & 0.034 {\tiny[0.022, 0.045]} & 0.023 ($-$0.007) & 0.017 ($-$0.003)\\
two positions, perceptron, correctness & 0.788 / 0.805 & 0.054 {\tiny[0.039, 0.069]} & 0.074 {\tiny[0.056, 0.090]} & 0.055 (0.032) & 0.073 (0.054)\\
two positions, perceptron, wrong-object labels & 0.749 / 0.797 & 0.063 {\tiny[0.048, 0.079]} & 0.067 {\tiny[0.051, 0.084]} & 0.094 (0.064) & 0.080 (0.050)\\
boxes and IoU, logistic, correctness & 0.820 / 0.849 & 0.005 {\tiny[$-$0.001, 0.011]} & 0.011 {\tiny[0.004, 0.017]} & 0.023 (0.001) & 0.029 (0.010)\\
boxes and IoU, logistic, wrong-object labels & 0.819 / 0.853 & 0.003 {\tiny[$-$0.003, 0.010]} & 0.007 {\tiny[0.000, 0.013]} & 0.024 (0.000) & 0.024 (0.002)\\
boxes and IoU, perceptron, correctness & 0.768 / 0.795 & 0.048 {\tiny[0.036, 0.061]} & 0.058 {\tiny[0.040, 0.078]} & 0.075 (0.051) & 0.082 (0.054)\\
boxes and IoU, perceptron, wrong-object labels & 0.805 / 0.825 & 0.028 {\tiny[0.016, 0.041]} & 0.041 {\tiny[0.026, 0.057]} & 0.038 (0.013) & 0.053 (0.029)\\
\midrule
Residual alone & 0.820 / 0.863 & --- & --- & --- & ---\\
\bottomrule
\end{tabular}}
\end{table}

%% file: tables/errtype.tex
\begin{table}[!htbp]
\centering
\caption{Telling the kind of error among wrong answers: AUROC for boxes on another object (IoU $<0.1$) against the other errors ($0.1\le$ IoU $<0.5$), mean over the six models, with 95\% intervals over images (10,000 resamples). Area: smaller written box scores higher. Geometry classifier: logistic regression on eleven features of the written box, trained on the wrong answers among the training items with the error type as label; with the residual: the same with the residual as a twelfth input. Last row: the difference of the two classifiers with the combined interval (images and ten refits on redrawn training items).}
\label{tab:errtype}
\footnotesize
\begin{tabular}{@{}lccc@{}}
\toprule
Score & Replication list & Held-out list & New items\\
\midrule
Self-prediction residual & 0.726 {\tiny[0.681, 0.771]} & 0.703 {\tiny[0.657, 0.750]} & 0.737 {\tiny[0.692, 0.780]}\\
Detector & 0.597 {\tiny[0.548, 0.645]} & 0.580 {\tiny[0.529, 0.631]} & 0.590 {\tiny[0.539, 0.643]}\\
Two-teacher probe & 0.505 {\tiny[0.456, 0.554]} & 0.507 {\tiny[0.458, 0.559]} & 0.503 {\tiny[0.448, 0.559]}\\
Shift-teacher probe & 0.533 {\tiny[0.487, 0.578]} & 0.560 {\tiny[0.511, 0.610]} & 0.539 {\tiny[0.487, 0.590]}\\
Supervised linear probe, correctness & 0.528 {\tiny[0.475, 0.581]} & 0.532 {\tiny[0.478, 0.586]} & 0.542 {\tiny[0.487, 0.595]}\\
Pre-generation probe & 0.579 {\tiny[0.527, 0.631]} & 0.538 {\tiny[0.485, 0.591]} & 0.560 {\tiny[0.502, 0.619]}\\
Supervised probe, wrong-object labels & 0.654 {\tiny[0.608, 0.701]} & 0.643 {\tiny[0.597, 0.690]} & 0.682 {\tiny[0.635, 0.727]}\\
Coordinate log-prob & 0.457 {\tiny[0.402, 0.514]} & 0.444 {\tiny[0.393, 0.495]} & 0.438 {\tiny[0.387, 0.489]}\\
Area of the written box & 0.670 {\tiny[0.613, 0.725]} & 0.636 {\tiny[0.582, 0.687]} & 0.684 {\tiny[0.627, 0.735]}\\
Geometry classifier & 0.668 {\tiny[0.616, 0.719]} & 0.650 {\tiny[0.599, 0.699]} & 0.695 {\tiny[0.643, 0.743]}\\
Geometry classifier with the residual & 0.730 {\tiny[0.682, 0.776]} & 0.703 {\tiny[0.656, 0.749]} & 0.751 {\tiny[0.708, 0.792]}\\
\midrule
With the residual minus without & 0.061 {\tiny[0.031, 0.092]} & 0.054 {\tiny[0.025, 0.083]} & 0.056 {\tiny[0.024, 0.089]}\\
\bottomrule
\end{tabular}
\end{table}

%% file: tables/offideal.tex
\begin{table}[!t]
\centering
\caption{Away from ideal labels and images: AUROC of SPRED minus that of a supervised probe, new items unless named. Bold: 95\% interval excludes zero. n/a: not run. $^{b}$Supervised linear probe trained on the correctness of 250 items; the detector uses the same items without labels; mean over 20 draws of the 250 items, with the interval over images and draws. No wrong-object (extent) labels: training items with that error type removed before the supervised probe is refitted, scored on that type against correct answers. Extent: $0.1\le\mathrm{IoU}<0.5$. In \qlarge{} the selected linear probe reads the pre-generation state, so those two rows coincide. $^{d}$The same items are also removed from the detector's unlabelled training items, and both sides are refitted on what remains; in the other rows the detector uses all training items. $^{c}$Detector and supervised linear probe fitted on RefCOCO and applied unchanged.}
\label{tab:offideal}
\footnotesize
\setlength{\tabcolsep}{3pt}
\begin{tabular}{@{}lccccc@{}}
\toprule
Condition (supervised probe) & 3B & 7B & Q3-4B & Q3-8B & InternVL\\
\midrule
Labels for 250 items only$^{b}$ & \textbf{0.059} & \textbf{0.042} & 0.020 & \textbf{0.049} & \textbf{0.093}\\
No wrong-object labels, linear, correctness$^{d}$ & \textbf{0.101} & \textbf{0.101} & \textbf{0.122} & \textbf{0.085} & \textbf{0.101}\\
No wrong-object labels, linear, $1-\mathrm{IoU}$ & \textbf{0.100} & \textbf{0.072} & \textbf{0.109} & \textbf{0.079} & \textbf{0.107}\\
No wrong-object labels, pre-generation & \textbf{0.094} & \textbf{0.083} & \textbf{0.112} & \textbf{0.084} & \textbf{0.125}\\
No extent labels, linear, correctness$^{d}$ & \textbf{0.058} & \textbf{0.102} & \textbf{0.080} & \textbf{0.051} & \textbf{0.131}\\
No extent labels, linear, $1-\mathrm{IoU}$ & 0.015 & \textbf{0.077} & \textbf{0.047} & \textbf{0.037} & 0.004\\
No extent labels, pre-generation & \textbf{0.054} & \textbf{0.111} & \textbf{0.083} & \textbf{0.059} & \textbf{0.115}\\
Objects365, transferred$^{c}$ & \textbf{0.037} & 0.004 & 0.032 & n/a & n/a\\
Remote sensing, transferred$^{c}$ & $-$0.007 & n/a & \textbf{0.043} & n/a & n/a\\
\bottomrule
\end{tabular}
\end{table}

%% file: tables/samplebase.tex
\begin{table}[!htbp]
\centering
\caption{Sampling baselines, new items, all errors. Extra: generations beyond the answer itself. EigenScore: log-determinant of the covariance of the sentence embeddings of the sampled answers (temperature 0.5, top-$p$ 0.99, top-$k$ 5); self-consistency: one minus the mean IoU between the greedy box and the sampled boxes (temperature 1). Columns: AUROC; detector minus the score (upper block) or the combination minus five-answer self-consistency (lower block), with 95\% intervals (bootstrap over images). Combinations add the standardized scores with equal weight.}
\label{tab:samplebase}
\footnotesize
\setlength{\tabcolsep}{2.5pt}
\resizebox{\textwidth}{!}{%
\begin{tabular}{@{}lccccc@{}}
\toprule
 & & \multicolumn{2}{c}{\qsmall} & \multicolumn{2}{c}{\qfour}\\
\cmidrule(lr){3-4}\cmidrule(lr){5-6}
Score & Extra & AUROC & Difference & AUROC & Difference\\
\midrule
Detector & 0 & 0.829 [0.797, 0.859] & --- & 0.858 [0.827, 0.886] & ---\\
EigenScore, 10 answers~\cite{chen2024inside} & 10 & 0.619 & 0.210 [0.164, 0.256] & 0.709 & 0.149 [0.105, 0.195]\\
Self-consistency, 1 sampled answer & 1 & 0.726 & 0.103 [0.056, 0.150] & 0.739 & 0.119 [0.078, 0.164]\\
Self-consistency, 2 sampled answers & 2 & 0.767 & 0.062 [0.023, 0.101] & 0.773 & 0.085 [0.049, 0.123]\\
Self-consistency, 5 sampled answers & 5 & 0.818 & 0.010 [$-$0.019, 0.040] & 0.830 & 0.028 [$-$0.006, 0.062]\\
Pairwise agreement, 5 sampled answers & 5 & 0.806 & 0.023 [$-$0.008, 0.054] & 0.827 & 0.031 [$-$0.003, 0.065]\\
\midrule
Detector + second answer at a quarter of the pixels & 1 & 0.851 & 0.032 [0.003, 0.062] & 0.867 & 0.037 [0.004, 0.071]\\
Detector + self-consistency, 5 answers & 5 & 0.846 & 0.028 [0.007, 0.049] & 0.877 & 0.047 [0.022, 0.073]\\
\bottomrule
\end{tabular}}
\end{table}

%% file: tables/extraadd.tex
\begin{table}[!htbp]
\centering
\caption{The self-prediction residual added to checks that need extra passes, new items: AUROC of the check alone $\to$ check $+\,0.5\times$ residual, and the difference with 95\% interval (bootstrap over images). Each check scores $1-\mathrm{IoU}$ between the greedy box and the second box (1 when that box is missing), standardized on the test items; the residual is standardized on the unlabelled training items. $^{a}$\qeight{} for the other models, \qlarge{} for \qeight. Five sampled answers: mean over the first five answers sampled at temperature 1. Over the 14 model--check cells the residual raises the AUROC for wrong-object errors in 14 (interval above zero in 13) and for all errors in 14 (12).}
\label{tab:extraadd}
\footnotesize
\begin{tabular}{@{}>{\raggedright\arraybackslash}p{4.3cm}cc@{}}
\toprule
Model, check & \hq{Check $\to$}{check + residual} & Difference\\
\midrule
\multicolumn{3}{@{}l}{\emph{Wrong-object errors against correct answers}}\\
\qsmall, Content-shifted second answer & 0.804 $\to$ 0.891 & 0.087 [0.047, 0.133]\\
\qsmall, Box of another model$^{a}$ & 0.908 $\to$ 0.943 & 0.034 [0.008, 0.063]\\
\qsmall, Second answer at a quarter of the pixels & 0.786 $\to$ 0.881 & 0.094 [0.059, 0.136]\\
\qsmall, Five sampled answers & 0.853 $\to$ 0.898 & 0.045 [0.023, 0.071]\\
\qlarge, Content-shifted second answer & 0.819 $\to$ 0.860 & 0.040 [$-$0.001, 0.083]\\
\qlarge, Box of another model$^{a}$ & 0.891 $\to$ 0.928 & 0.037 [0.008, 0.070]\\
\qfour, Content-shifted second answer & 0.710 $\to$ 0.861 & 0.150 [0.092, 0.216]\\
\qfour, Box of another model$^{a}$ & 0.854 $\to$ 0.916 & 0.062 [0.032, 0.097]\\
\qfour, Second answer at a quarter of the pixels & 0.768 $\to$ 0.864 & 0.095 [0.056, 0.141]\\
\qfour, Five sampled answers & 0.854 $\to$ 0.897 & 0.042 [0.022, 0.066]\\
\qeight, Content-shifted second answer & 0.751 $\to$ 0.867 & 0.116 [0.059, 0.181]\\
\qeight, Box of another model$^{a}$ & 0.863 $\to$ 0.918 & 0.055 [0.022, 0.098]\\
\qivl, Content-shifted second answer & 0.734 $\to$ 0.862 & 0.128 [0.072, 0.185]\\
\qivl, Box of another model$^{a}$ & 0.810 $\to$ 0.894 & 0.084 [0.046, 0.126]\\
\midrule
\multicolumn{3}{@{}l}{\emph{All errors}}\\
\qsmall, Content-shifted second answer & 0.765 $\to$ 0.814 & 0.050 [0.021, 0.079]\\
\qsmall, Box of another model$^{a}$ & 0.861 $\to$ 0.883 & 0.022 [0.001, 0.043]\\
\qsmall, Second answer at a quarter of the pixels & 0.770 $\to$ 0.812 & 0.041 [0.016, 0.068]\\
\qsmall, Five sampled answers & 0.818 $\to$ 0.830 & 0.012 [$-$0.004, 0.028]\\
\qlarge, Content-shifted second answer & 0.787 $\to$ 0.808 & 0.021 [$-$0.002, 0.045]\\
\qlarge, Box of another model$^{a}$ & 0.835 $\to$ 0.869 & 0.034 [0.016, 0.054]\\
\qfour, Content-shifted second answer & 0.690 $\to$ 0.792 & 0.102 [0.066, 0.140]\\
\qfour, Box of another model$^{a}$ & 0.800 $\to$ 0.847 & 0.048 [0.022, 0.074]\\
\qfour, Second answer at a quarter of the pixels & 0.718 $\to$ 0.809 & 0.091 [0.060, 0.125]\\
\qfour, Five sampled answers & 0.830 $\to$ 0.860 & 0.029 [0.010, 0.050]\\
\qeight, Content-shifted second answer & 0.691 $\to$ 0.789 & 0.098 [0.055, 0.143]\\
\qeight, Box of another model$^{a}$ & 0.792 $\to$ 0.841 & 0.049 [0.025, 0.075]\\
\qivl, Content-shifted second answer & 0.778 $\to$ 0.834 & 0.057 [0.025, 0.090]\\
\qivl, Box of another model$^{a}$ & 0.868 $\to$ 0.891 & 0.022 [0.001, 0.044]\\
\bottomrule
\end{tabular}
\end{table}

%% file: tables/pointlists.tex
\begin{table}[!htbp]
\centering
\caption{Point outputs on the replication and held-out lists, on which no point answer had been generated before: gain in AUROC for all errors when the residual is added with weight 0.5 to each score. \qsmall{}, \qlarge{} and \qfour{} write one point; an answer is an error when the point lies outside the annotated box, and the residual is the distance between the point read out by the map fitted to the model's own points and the point written. Prompts, with the expression in place of \emph{expression}: \qsmall{}, ``Locate `\emph{expression}' in the image, and output its point coordinates in JSON format.''; \qlarge{} and \qfour{}, ``Point to `\emph{expression}' in the image, and output the point in JSON format with the key `point\_2d'.'' The read-out and the label-free probes are fitted on unlabelled training items with point answers, the supervised probes with their labels, and nothing is fitted on the two lists. Items with a point answer and errors: 2{,}389 and 213 (\qsmall), 2{,}285 and 124 (\qlarge), 2{,}369 and 144 (\qfour); the residual alone reaches an AUROC of 0.681. Two-teacher probe, both lists together, per model: 0.070, 0.015 and 0.020 on \qsmall, \qlarge{} and \qfour. Mean over the three models with a combined interval over images and ten refits on redrawn training items.}
\label{tab:pointlists}
\footnotesize
\begin{tabular}{@{}>{\raggedright\arraybackslash}p{2.9cm}ccc@{}}
\toprule
Score & \hq{Both}{lists} & \hq{Replication}{list} & \hq{Held-out}{list}\\
\midrule
Two-teacher probe & 0.035 [0.015, 0.055] & 0.039 [0.015, 0.063] & 0.032 [0.007, 0.058]\\
Sampling-teacher probe & 0.047 [0.027, 0.068] & 0.053 & 0.043\\
Supervised linear probe, correctness & 0.012 [$-$0.007, 0.031] & 0.010 & 0.014\\
SAPLMA perceptron, correctness & 0.035 [0.010, 0.060] & 0.030 & 0.041\\
Coordinate log-prob & 0.080 [0.062, 0.098] & 0.087 & 0.071\\
\bottomrule
\end{tabular}
\end{table}

%% file: tables/teachers.tex
\begin{table}[!htbp]
\centering
\caption{Single consistency teachers, each with the residual added as in the detector: AUROC for all errors, and in parentheses the detector minus the variant with 95\% interval (bootstrap over images). Sampling: one sampled answer; two samples: two sampled answers, the teacher cost of the detector; shift: the content-shifted answer alone. ---: two-sample teacher not trained for that model. Over the 22 cells the detector is ahead in 4 and behind in 0.}
\label{tab:teachers}
\footnotesize
\setlength{\tabcolsep}{2.5pt}
\resizebox{\textwidth}{!}{%
\begin{tabular}{@{}lcccc@{}}
\toprule
Model, set & Detector & Sampling teacher & Two samples & Shift teacher\\
\midrule
\qsmall, new & 0.829 & 0.803 (0.026 [$-$0.000, 0.052]) & 0.827 (0.002 [$-$0.011, 0.015]) & 0.825 (0.004 [$-$0.006, 0.014])\\
\qsmall, RefCOCO+ & 0.830 & 0.823 (0.007 [$-$0.013, 0.025]) & 0.835 ($-$0.005 [$-$0.016, 0.007]) & 0.819 (0.011 [0.002, 0.021])\\
\qlarge, new & 0.848 & 0.814 (0.034 [0.005, 0.064]) & 0.846 (0.002 [$-$0.015, 0.019]) & 0.822 (0.027 [$-$0.002, 0.056])\\
\qlarge, RefCOCO+ & 0.822 & 0.828 ($-$0.006 [$-$0.032, 0.018]) & 0.809 (0.013 [$-$0.004, 0.030]) & 0.827 ($-$0.005 [$-$0.031, 0.020])\\
\qfour, new & 0.858 & 0.856 (0.002 [$-$0.004, 0.009]) & --- & 0.835 (0.023 [0.007, 0.040])\\
\qeight, new & 0.837 & 0.825 (0.012 [$-$0.013, 0.036]) & --- & 0.854 ($-$0.017 [$-$0.039, 0.005])\\
\qivl, new & 0.867 & 0.858 (0.009 [0.001, 0.017]) & 0.849 (0.017 [$-$0.002, 0.038]) & 0.869 ($-$0.003 [$-$0.014, 0.007])\\
\qivl, RefCOCO+ & 0.881 & 0.882 ($-$0.001 [$-$0.009, 0.007]) & 0.882 ($-$0.002 [$-$0.020, 0.017]) & 0.885 ($-$0.005 [$-$0.015, 0.005])\\
\bottomrule
\end{tabular}}
\end{table}

%% file: tables/offidealci.tex
\begin{table}[!htbp]
\centering
\caption{Supervised probes trained without one error type (\cref{tab:offideal}), new items; here the detector uses all training items: AUROC of the supervised probe on that type with all labels $\to$ without them; detector minus the probe without them, with 95\% interval (bootstrap over images, 10,000 resamples). Lower block: detector minus the transferred supervised linear probe, AUROC for all errors on Objects365 (1{,}000, 996 and 999 items) and on remote-sensing images (1{,}000 and 997 items). For 250 labels see \cref{tab:labelsvsdetector}. In \qlarge{} the linear probe on correctness reads the pre-generation state, so those two rows coincide.}
\label{tab:offidealci}
\tiny
\setlength{\tabcolsep}{2pt}
\resizebox{\textwidth}{!}{%
\begin{tabular}{@{}lccccc@{}}
\toprule
Condition & \qsmall & \qlarge & \qfour & \qeight & \qivl\\
\midrule
No wrong-object labels, linear, correctness & 0.814$\to$0.758; 0.110 [0.065, 0.156] & 0.849$\to$0.784; 0.083 [0.018, 0.150] & 0.838$\to$0.753; 0.115 [0.065, 0.166] & 0.867$\to$0.804; 0.069 [0.023, 0.123] & 0.858$\to$0.779; 0.082 [0.038, 0.128]\\
No wrong-object labels, linear, $1-\mathrm{IoU}$ & 0.832$\to$0.768; 0.100 [0.055, 0.149] & 0.844$\to$0.794; 0.072 [0.016, 0.129] & 0.842$\to$0.758; 0.109 [0.067, 0.153] & 0.871$\to$0.794; 0.079 [0.025, 0.138] & 0.830$\to$0.755; 0.107 [0.060, 0.153]\\
No wrong-object labels, pre-generation & 0.826$\to$0.774; 0.094 [0.045, 0.145] & 0.849$\to$0.784; 0.083 [0.018, 0.150] & 0.849$\to$0.756; 0.112 [0.055, 0.172] & 0.859$\to$0.788; 0.084 [0.019, 0.152] & 0.780$\to$0.736; 0.125 [0.045, 0.204]\\
No wrong-object labels, SAPLMA & 0.812$\to$0.757; 0.110 [0.054, 0.170] & 0.796$\to$0.661; 0.205 [0.121, 0.297] & 0.840$\to$0.717; 0.150 [0.078, 0.228] & 0.806$\to$0.718; 0.155 [0.090, 0.227] & 0.762$\to$0.665; 0.197 [0.132, 0.262]\\
No extent labels, linear, correctness & 0.823$\to$0.742; 0.063 [0.023, 0.105] & 0.779$\to$0.728; 0.111 [0.063, 0.161] & 0.834$\to$0.776; 0.077 [0.036, 0.119] & 0.818$\to$0.771; 0.046 [0.007, 0.085] & 0.885$\to$0.761; 0.108 [0.060, 0.158]\\
No extent labels, linear, $1-\mathrm{IoU}$ & 0.802$\to$0.790; 0.015 [$-$0.022, 0.052] & 0.778$\to$0.763; 0.077 [0.032, 0.123] & 0.824$\to$0.806; 0.047 [0.012, 0.086] & 0.794$\to$0.780; 0.037 [0.002, 0.074] & 0.891$\to$0.865; 0.004 [$-$0.023, 0.031]\\
No extent labels, pre-generation & 0.792$\to$0.751; 0.054 [0.010, 0.100] & 0.779$\to$0.728; 0.111 [0.063, 0.161] & 0.824$\to$0.770; 0.083 [0.046, 0.122] & 0.802$\to$0.758; 0.059 [0.007, 0.113] & 0.798$\to$0.755; 0.115 [0.071, 0.158]\\
No extent labels, SAPLMA & 0.777$\to$0.668; 0.137 [0.084, 0.194] & 0.726$\to$0.670; 0.169 [0.114, 0.224] & 0.781$\to$0.701; 0.152 [0.095, 0.209] & 0.771$\to$0.678; 0.139 [0.076, 0.203] & 0.834$\to$0.683; 0.187 [0.115, 0.262]\\
\midrule
Objects365, transferred & 0.037 [0.012, 0.063] & 0.004 [$-$0.038, 0.046] & 0.032 [0.000, 0.064] & n/a & n/a\\
Remote sensing, transferred & $-$0.007 [$-$0.028, 0.014] & n/a & 0.043 [0.023, 0.063] & n/a & n/a\\
\bottomrule
\end{tabular}}
\end{table}

%% file: tables/tokenreadout.tex
\begin{sidewaystable}
\centering
\caption{The ridge read-out and a token-style read-out, \qsmall{} and \qlarge{}. Both read a box from the last prompt-token state and are fitted to the model's own answers on the same 2{,}000 unlabelled training items; the token-style read-out puts each of the four relative coordinates into $K$ equal bins and fits one softmax classifier per coordinate by cross-entropy. Its layer (any stored layer), bin count $K\in\{20,50\}$, decoding (most probable bin or expected bin centre) and penalty $\lambda\in\{10,100,1000,10000\}$ were fixed in advance and chosen, without labels, by the mean IoU with the model's own box in five-fold cross-validation on the training items: for both models the layer of the ridge read-out, $K=50$, expected bin centre and $\lambda=10$, the weakest penalty offered. AUROC for wrong-object errors against correct answers; gain: AUROC of $z(\text{two-teacher probe})+0.5\,z(\text{residual})$ minus that of the probe. Per model: stored fit with 95\% interval over images; mean: stored fits with combined interval over images and ten refits on redrawn training items, the token-style refits keeping the chosen configuration.}
\label{tab:tokenreadout}
\small
\begin{tabular}{@{}lccc@{}}
\toprule
 & \hqsmall & \hqlarge & Mean\\
\midrule
\multicolumn{4}{@{}l}{\emph{Replication list} (wrong-object errors: 76, 55)}\\
Ridge residual & 0.802 & 0.824 & 0.813\\
Token-style residual & 0.771 & 0.753 & 0.762\\
Ridge minus token-style & 0.031 [$-$0.014, 0.075] & 0.071 [0.008, 0.138] & 0.051 [0.003, 0.099]\\
Gain on the probe, ridge & 0.035 & 0.076 & 0.056 [0.030, 0.081]\\
Gain on the probe, token-style & 0.027 & 0.049 & 0.038 [0.012, 0.064]\\
Difference of the gains & 0.008 [$-$0.006, 0.022] & 0.027 [0.001, 0.053] & 0.017 [$-$0.001, 0.036]\\
\midrule
\multicolumn{4}{@{}l}{\emph{Held-out list} (wrong-object errors: 68, 52)}\\
Ridge residual & 0.865 & 0.770 & 0.818\\
Token-style residual & 0.845 & 0.770 & 0.807\\
Ridge minus token-style & 0.020 [$-$0.014, 0.055] & 0.001 [$-$0.054, 0.060] & 0.010 [$-$0.028, 0.049]\\
Gain on the probe, ridge & 0.063 & 0.063 & 0.063 [0.034, 0.092]\\
Gain on the probe, token-style & 0.059 & 0.063 & 0.061 [0.029, 0.093]\\
Difference of the gains & 0.004 [$-$0.009, 0.017] & 0.000 [$-$0.024, 0.026] & 0.002 [$-$0.015, 0.019]\\
\midrule
\multicolumn{4}{@{}l}{\emph{Mean IoU with the model's own box} (training items, five-fold)}\\
Ridge / token-style & 0.517 / 0.534 & 0.510 / 0.519 & ---\\
\bottomrule
\end{tabular}
\end{sidewaystable}

%% file: tables/stable.tex
\begin{table}[!htbp]
\centering
\caption{Stable errors and repeated consistency checks on the new items. An item is stable when the box of each of the seven views (eight passes) has IoU at least 0.5 with the original box; SPRED uses one pass. Stable: share of errors that are stable. Seven views: AUROC of the disagreement over the seven views. Brackets: 95\% paired-bootstrap intervals over images (10,000 resamples).}
\label{tab:stable}
\setlength{\tabcolsep}{2.5pt}
\begin{tabular}{@{}lcccc@{}}
\toprule
Model & Stable & \hq{Seven}{views} & \hq{SPRED $-$ seven views,}{all new items} & \hq{SPRED $-$ seven views,}{stable items}\\
\midrule
\qsmall{} & 0.208 & 0.837 & \hq{$-$0.008}{\relax[$-$0.039, +0.023]} & \hq{+0.098}{\relax[+0.008, +0.188]}\\ \addlinespace
\qlarge{} & 0.263 & 0.834 & \hq{+0.014}{\relax[$-$0.018, +0.047]} & \hq{+0.117}{\relax[+0.028, +0.211]}\\ \addlinespace
\qfour{} & 0.442 & 0.780 & \hq{+0.078}{\relax[+0.035, +0.120]} & \hq{+0.209}{\relax[+0.141, +0.280]}\\ \addlinespace
\qeight{} & 0.476 & 0.766 & \hq{+0.070}{\relax[+0.030, +0.112]} & \hq{+0.151}{\relax[+0.099, +0.207]}\\
\bottomrule
\end{tabular}
\end{table}

%% file: tables/readctl.tex
\begin{table}[!t]
\centering
\caption{The residual with its read-out replaced, replication list: AUROC of $1-\mathrm{IoU}$(read-out, written box) for wrong-object errors against correct answers, with the item's own read-out, the read-out of another item of the same set (mean over 100 random permutations) and the mean training box in place of the read-out. The held-out list in \cref{tab:readctlho}.}
\label{tab:readctl}
\setlength{\tabcolsep}{4pt}
\begin{tabular}{@{}lccc@{}}
\toprule
Model & Own & Another item's & Mean training box\\
\midrule
\qsmall{} & 0.802 & 0.626 & 0.679\\
\qlarge{} & 0.824 & 0.565 & 0.610\\
\qfour{} & 0.830 & 0.603 & 0.625\\
\qeight{} & 0.802 & 0.567 & 0.580\\
\qivl{} & 0.744 & 0.545 & 0.586\\
\qllava{} & 0.845 & 0.620 & 0.672\\
\bottomrule
\end{tabular}
\end{table}

%% file: tables/readctlho.tex
\begin{table}[!t]
\centering
\caption{The residual with its read-out replaced, held-out list: AUROC of $1-\mathrm{IoU}$(read-out, written box) for wrong-object errors against correct answers, with the item's own read-out, the read-out of another item of the same set (mean over 100 random permutations) and the mean training box in place of the read-out.}
\label{tab:readctlho}
\footnotesize
\begin{tabular}{@{}lccc@{}}
\toprule
Model & Own read-out & Another item's read-out & Mean training box\\
\midrule
\qsmall{} & 0.865 & 0.647 & 0.691\\
\qlarge{} & 0.770 & 0.560 & 0.580\\
\qfour{} & 0.839 & 0.602 & 0.629\\
\qeight{} & 0.780 & 0.574 & 0.617\\
\qivl{} & 0.841 & 0.610 & 0.680\\
\qllava{} & 0.826 & 0.622 & 0.694\\
\bottomrule
\end{tabular}
\end{table}

%% file: tables/readctlnewitems.tex
\begin{table}[!htbp]
\centering
\caption{The residual with its read-out replaced, new items: AUROC of $1-\mathrm{IoU}$(read-out, written box) for wrong-object errors against correct answers, with the item's own read-out, the read-out of another item of the same set (mean over 100 random permutations) and the mean training box in place of the read-out.}
\label{tab:readctlnewitems}
\footnotesize
\begin{tabular}{@{}lccc@{}}
\toprule
Model & Own read-out & Another item's read-out & Mean training box\\
\midrule
\qsmall{} & 0.870 & 0.612 & 0.651\\
\qlarge{} & 0.833 & 0.606 & 0.623\\
\qfour{} & 0.867 & 0.591 & 0.621\\
\qeight{} & 0.833 & 0.580 & 0.589\\
\qivl{} & 0.867 & 0.579 & 0.668\\
\qllava{} & 0.907 & 0.624 & 0.699\\
\bottomrule
\end{tabular}
\end{table}

%% file: tables/secondans.tex
\begin{sidewaystable}
\centering
\caption{The read-out and the second answer. Items whose answer changes under the content shift (IoU of the two boxes below 0.5; the second answer is mapped back to the original view). Slope: coefficient of the read-out centre minus the first box centre in a regression without intercept of the second box centre minus the first on it and on the mean training box centre minus the first ($x$ and $y$ stacked), with a 95\% interval over images; permuted: the same with read-outs permuted among these items (mean of 200 permutations); with annotated: the annotated centre minus the first added as a third regressor; second also wrong: items whose second answer is wrong as well (number in brackets; n/a: fewer than 30 items). Closer: share of the changed items whose read-out is closer to the second box than to the first by IoU, over all and over those with a large residual.}
\label{tab:secondans}
\scriptsize
\setlength{\tabcolsep}{2.5pt}
\begin{tabular}{@{}llcccccc@{}}
\toprule
Set & Model & Changed & Slope & Permuted & With annotated & Second also wrong & Closer: all / large\\
\midrule
New items & \qsmall{} & 132 & 1.086 {\tiny[0.881, 1.276]} & 0.018 & 0.812 {\tiny[0.548, 1.049]} & 0.988 {\tiny[0.697, 1.283]} (83) & 0.447 / 0.585\\
 & \qlarge{} & 109 & 0.947 {\tiny[0.765, 1.117]} & 0.016 & 0.640 {\tiny[0.441, 0.861]} & 0.941 {\tiny[0.666, 1.209]} (65) & 0.440 / 0.673\\
 & \qfour{} & 45 & 1.022 {\tiny[0.754, 1.293]} & 0.083 & 0.918 {\tiny[0.626, 1.201]} & 1.006 {\tiny[0.659, 1.356]} (32) & 0.511 / 0.789\\
 & \qeight{} & 41 & 1.251 {\tiny[0.859, 1.583]} & 0.017 & 1.100 {\tiny[0.673, 1.440]} & 1.358 {\tiny[0.876, 1.740]} (32) & 0.293 / 0.524\\
 & \qivl{} & 103 & 0.942 {\tiny[0.792, 1.084]} & $-$0.001 & 0.603 {\tiny[0.346, 0.840]} & 0.883 {\tiny[0.672, 1.095]} (54) & 0.553 / 0.740\\
 & \qllava{} & 249 & 0.678 {\tiny[0.518, 0.838]} & $-$0.046 & 0.544 {\tiny[0.343, 0.724]} & 0.611 {\tiny[0.395, 0.814]} (188) & 0.321 / 0.420\\
\addlinespace
RefCOCO+ & \qsmall{} & 144 & 1.197 {\tiny[0.999, 1.394]} & $-$0.001 & 0.937 {\tiny[0.701, 1.160]} & 1.013 {\tiny[0.764, 1.252]} (90) & 0.424 / 0.662\\
 & \qlarge{} & 93 & 1.063 {\tiny[0.767, 1.373]} & $-$0.001 & 0.772 {\tiny[0.454, 1.125]} & 1.034 {\tiny[0.595, 1.446]} (59) & 0.419 / 0.528\\
 & \qivl{} & 95 & 1.063 {\tiny[0.790, 1.308]} & 0.010 & 0.710 {\tiny[0.393, 1.037]} & 0.990 {\tiny[0.604, 1.425]} (54) & 0.474 / 0.660\\
\addlinespace
Held-out list & \qsmall{} & 138 & 1.013 {\tiny[0.790, 1.223]} & $-$0.016 & 0.639 {\tiny[0.389, 0.896]} & 0.915 {\tiny[0.598, 1.232]} (85) & 0.457 / 0.613\\
 & \qlarge{} & 109 & 1.006 {\tiny[0.798, 1.201]} & 0.004 & 0.727 {\tiny[0.478, 0.955]} & 0.939 {\tiny[0.562, 1.230]} (65) & 0.468 / 0.677\\
 & \qfour{} & 53 & 1.100 {\tiny[0.796, 1.417]} & 0.085 & 0.781 {\tiny[0.363, 1.206]} & 0.767 {\tiny[0.414, 1.118]} (36) & 0.434 / 0.588\\
 & \qeight{} & 39 & 1.169 {\tiny[0.839, 1.531]} & $-$0.016 & 1.026 {\tiny[0.555, 1.406]} & n/a (24) & 0.538 / 0.737\\
 & \qivl{} & 90 & 1.054 {\tiny[0.791, 1.292]} & 0.052 & 0.830 {\tiny[0.593, 1.057]} & 0.807 {\tiny[0.481, 1.143]} (55) & 0.500 / 0.646\\
 & \qllava{} & 271 & 0.502 {\tiny[0.322, 0.708]} & $-$0.073 & 0.391 {\tiny[0.211, 0.606]} & 0.414 {\tiny[0.219, 0.633]} (211) & 0.328 / 0.448\\
\addlinespace
Objects365 & \qsmall{} & 190 & 0.542 {\tiny[0.365, 0.693]} & $-$0.037 & 0.278 {\tiny[0.124, 0.435]} & 0.544 {\tiny[0.316, 0.765]} (125) & 0.379 / 0.409\\
 & \qlarge{} & 134 & 0.771 {\tiny[0.565, 0.993]} & 0.027 & 0.412 {\tiny[0.218, 0.631]} & 0.854 {\tiny[0.494, 1.182]} (69) & 0.351 / 0.388\\
 & \qfour{} & 58 & 0.577 {\tiny[0.266, 0.878]} & 0.022 & 0.386 {\tiny[0.067, 0.679]} & 0.520 {\tiny[0.186, 0.851]} (36) & 0.276 / 0.286\\
\bottomrule
\end{tabular}
\end{sidewaystable}

%% file: tables/sameclass.tex
\begin{table}[!htbp]
\centering
\caption{The read-out and other objects of the target's class, by the MS-COCO instance annotations; every item has at least one such object. Upper block: wrong-object errors with a further same-class object, the displacement of the read-out from the written box regressed on the annotated target and on the nearest other object of the same class (and the mean training box), without intercept; correct answers: the coefficient of the nearest other object. Lower block: the three observations of \cref{sec:measure} on items with a single further object of the class. Held-out list and new items: six models; RefCOCO+: three. Brackets: 95\% intervals over images (10,000 resamples), stored fit.}
\label{tab:sameclass}
\footnotesize
\setlength{\tabcolsep}{3pt}
\begin{tabular}{@{}lccc@{}}
\toprule
& Held-out list & New items & RefCOCO+\\
\midrule
Target, coefficient & 0.220 {\tiny[0.178, 0.268]} & 0.293 {\tiny[0.240, 0.357]} & 0.266 {\tiny[0.211, 0.323]}\\
Nearest same-class object, coefficient & 0.028 {\tiny[$-$0.011, 0.070]} & $-$0.009 {\tiny[$-$0.070, 0.053]} & $-$0.025 {\tiny[$-$0.081, 0.028]}\\
Difference & 0.193 {\tiny[0.132, 0.255]} & 0.302 {\tiny[0.221, 0.391]} & 0.291 {\tiny[0.219, 0.371]}\\
Correct answers: nearest competitor & $-$0.006 & $-$0.015 & $-$0.006\\
\midrule
\multicolumn{4}{@{}l}{\emph{Items with one further object of the class}}\\
Residual AUROC, wrong-object errors & 0.859 & 0.885 & 0.836\\
Read-out towards the target & 0.243 & 0.305 & 0.319\\
Second answer on the read-out (slope) & 0.918 & 1.016 & 1.139\\
\bottomrule
\end{tabular}
\end{table}

%% file: tables/o365.tex
\begin{table}[!htbp]
\centering
\caption{Images outside MS-COCO: the Objects365 part of Ref-L4 (test items with disjoint images). Upper blocks: AUROC; the detector and the probes are trained on unlabelled Objects365 items unless stated. Lower blocks: paired differences and gains with 95\% bootstrap intervals over images (10,000 resamples).}
\label{tab:o365}
\footnotesize
\resizebox{\textwidth}{!}{%
\begin{tabular}{lccc}
\toprule
 & \qsmall & \qlarge & \qfour\\
\midrule
\multicolumn{4}{l}{\emph{All errors}}\\
Detector (ours) & 0.758 & 0.692 & 0.762\\
Detector trained on RefCOCO, applied unchanged & 0.768 & 0.691 & 0.746\\
Two-teacher probe without the residual & 0.768 & 0.685 & 0.764\\
Sampling-teacher probe & 0.769 & 0.681 & 0.747\\
Coordinate log-prob & 0.669 & 0.583 & 0.682\\
Supervised linear probe, RefCOCO labels & 0.731 & 0.687 & 0.714\\
Supervised probe on $1-\mathrm{IoU}$, RefCOCO labels & 0.722 & 0.680 & 0.740\\
SAPLMA probe, RefCOCO labels & 0.680 & 0.626 & 0.654\\
\midrule
\multicolumn{4}{l}{\emph{Wrong-object errors}}\\
Detector (ours) & 0.795 & 0.666 & 0.760\\
Two-teacher probe without the residual & 0.783 & 0.612 & 0.732\\
Sampling-teacher probe & 0.765 & 0.607 & 0.712\\
Sampling-teacher probe + residual & 0.780 & 0.661 & 0.743\\
\midrule
\multicolumn{4}{l}{\emph{Paired differences}}\\
Ours $-$ coordinate log-prob & +0.089 [+0.044, +0.135] & +0.108 [+0.059, +0.158] & +0.080 [+0.034, +0.127]\\
Ours $-$ supervised probe, RefCOCO labels & +0.026 [$-$0.002, +0.055] & +0.004 [$-$0.039, +0.048] & +0.048 [+0.015, +0.081]\\
Ours trained on RefCOCO $-$ supervised probe, RefCOCO labels & 0.037 [0.012, 0.063] & 0.004 [$-$0.038, 0.046] & 0.032 [0.000, 0.064]\\
Ours trained on RefCOCO $-$ probe on $1-\mathrm{IoU}$, RefCOCO labels & 0.046 [0.019, 0.074] & 0.012 [$-$0.030, 0.053] & 0.006 [$-$0.021, 0.034]\\
Ours $-$ probe on $1-\mathrm{IoU}$, RefCOCO labels & 0.035 [0.007, 0.064] & 0.012 [$-$0.031, 0.054] & 0.022 [$-$0.008, 0.053]\\
Ours trained on RefCOCO $-$ SAPLMA probe, RefCOCO labels & 0.088 [0.054, 0.122] & 0.066 [0.019, 0.114] & 0.092 [0.046, 0.136]\\
Ours $-$ ours trained on RefCOCO & $-$0.010 [$-$0.040, +0.019] & +0.000 [$-$0.032, +0.033] & +0.016 [$-$0.006, +0.037]\\
Ours $-$ probe without residual, all errors & $-$0.010 [$-$0.027, +0.007] & +0.007 [$-$0.010, +0.024] & $-$0.002 [$-$0.021, +0.016]\\
Ours $-$ probe without residual, wrong object & +0.012 [$-$0.011, +0.034] & +0.054 [+0.028, +0.079] & +0.028 [$-$0.002, +0.057]\\
\midrule
\multicolumn{4}{l}{\emph{Read-out and probes fitted on RefCOCO: gain from the residual on wrong-object errors by written-box width,}}\\
\multicolumn{4}{l}{\emph{mean over the seven hidden-state probes and the three models}}\\
Narrower than $2\sigma$ & \multicolumn{3}{c}{0.013 [0.004, 0.021]}\\
At least $2\sigma$ wide & \multicolumn{3}{c}{0.005 [$-$0.022, 0.033]}\\
Difference & \multicolumn{3}{c}{$-$0.007 [$-$0.036, 0.021]}\\
\bottomrule
\end{tabular}}
\end{table}

%% file: tables/external.tex
\begin{table}[!htbp]
\centering
\caption{A frozen open-vocabulary detector as a second localizer, new items, \qsmall{}, \qlarge{} and \qfour{}, wrong-object errors against correct answers. The external detector (IDEA-Research/grounding-dino-tiny~\cite{liu2024groundingdino}, zero-shot) is given the same expression. Top box: $1-\mathrm{IoU}$ between the external detector's highest-scoring box and the written box (the score of the main text); support: minus the largest product of its box confidence and IoU with the written box. The external score has no training values and is standardized with its mean and standard deviation on the scored items; probe and residual as in \cref{eq:combine}, all weights 0.5. Mean over the three models; brackets: combined intervals over images and ten refits of probe and residual. For reference, the residual alone reaches 0.857 and adds 0.057 [0.037, 0.078] to the probe on these items.}
\label{tab:external}
\footnotesize
\begin{tabular}{@{}>{\raggedright\arraybackslash}p{3.7cm}cc@{}}
\toprule
 & Top box & Support\\
\midrule
External score alone (AUROC) & 0.700 [0.647, 0.752] & 0.759 [0.709, 0.810]\\
Probe $+$ external, minus probe & 0.001 [$-$0.021, 0.022] & 0.016 [$-$0.001, 0.033]\\
Probe $+$ residual, minus probe $+$ external & 0.057 [0.031, 0.083] & 0.041 [0.019, 0.064]\\
Probe $+$ external $+$ residual, minus probe $+$ external & 0.049 [0.032, 0.065] & 0.047 [0.031, 0.062]\\
\bottomrule
\end{tabular}
\end{table}

%% file: tables/ownother.tex
\begin{table}[!htbp]
\centering
\caption{Gain in AUROC on wrong-object errors when the term added to a score (weight 0.5, \cref{eq:combine}) is the model's own residual or the mean residual of the other five models on the same item, each standardized on the training items. Means over the six models (four Qwen models for the attention scores); in brackets, the combined interval of the ratio over images and ten refits on redrawn training items (post hoc). Seven probes: the label-free and supervised probes without the two trained on wrong-object labels. One other model's residual alone, added to the two-teacher probe: 0.032 on the replication list, against 0.042 with the own residual.}
\label{tab:ownother}
\small
\setlength{\tabcolsep}{3pt}
\begin{tabular}{@{}lcccccc@{}}
\toprule
& \multicolumn{3}{c}{Replication list} & \multicolumn{3}{c}{Held-out list}\\
\cmidrule(lr){2-4}\cmidrule(l){5-7}
Score & Own & Others & Ratio & Own & Others & Ratio\\
\midrule
Two-teacher probe & 0.042 & 0.047 & 1.13 {\scriptsize[0.88, 1.45]} & 0.044 & 0.040 & 0.90 {\scriptsize[0.66, 1.12]}\\
Supervised linear probe & 0.037 & 0.040 & 1.07 {\scriptsize[0.80, 1.40]} & 0.037 & 0.034 & 0.92 {\scriptsize[0.68, 1.17]}\\
Coordinate log-prob & 0.117 & 0.119 & 1.02 {\scriptsize[0.91, 1.13]} & 0.131 & 0.120 & 0.92 {\scriptsize[0.80, 1.02]}\\
Seven probes, mean & 0.039 & 0.039 & 1.01 {\scriptsize[0.77, 1.30]} & 0.043 & 0.038 & 0.89 {\scriptsize[0.67, 1.10]}\\
Fifteen scores, mean & 0.061 & 0.061 & 1.01 {\scriptsize[0.85, 1.19]} & 0.065 & 0.059 & 0.91 {\scriptsize[0.76, 1.06]}\\
\bottomrule
\end{tabular}
\end{table}

%% file: tables/ownothermodel.tex
\begin{table}[!htbp]
\centering
\caption{Gain in AUROC on wrong-object errors when the term added to the two-teacher probe (weight 0.5) is the model's own residual, the mean residual of the other five models on the same item, or the residual of a single other model (averaged over the five), per model; stored fit (post hoc).}
\label{tab:ownothermodel}
\small
\setlength{\tabcolsep}{3pt}
\begin{tabular}{@{}lcccccc@{}}
\toprule
& \multicolumn{3}{c}{Replication list} & \multicolumn{3}{c}{Held-out list}\\
\cmidrule(lr){2-4}\cmidrule(l){5-7}
Model & Own & Others & One other & Own & Others & One other\\
\midrule
\qsmall{} & 0.035 & 0.042 & 0.026 & 0.063 & 0.038 & 0.025\\
\qlarge{} & 0.076 & 0.054 & 0.040 & 0.063 & 0.078 & 0.059\\
\qfour{} & 0.052 & 0.062 & 0.044 & 0.058 & 0.049 & 0.032\\
\qeight{} & 0.044 & 0.068 & 0.049 & 0.018 & 0.033 & 0.020\\
\qivl{} & 0.025 & 0.044 & 0.031 & 0.053 & 0.040 & 0.027\\
\qllava{} & 0.017 & 0.011 & 0.004 & 0.012 & 0.002 & $-$0.004\\
\bottomrule
\end{tabular}
\end{table}

%% file: tables/controlled.tex
\begin{table}[!htbp]
\centering
\caption{Error detection with item difficulty held fixed. Natural errors: mean AUROC over the new items of the five models (MTLA (layers 8--21): four Qwen models). Across models: share of the 1{,}224 cases (1{,}146 new items answered by all five models, ten model pairs, one model right and the other wrong) in which the wrong model has the higher score percentile; 0.5 for a score that records only the difficulty the two models share (MTLA, layers 8--21: the 664 cases between Qwen models). $^{\ast}$Supervised, trained on gold labels of the 2{,}000 training items (correctness; $1-\mathrm{IoU}$ for the second linear probe). Brackets: 95\% bootstrap intervals over images ($^{\dagger}$combined with ten refits on redrawn training items); ---: per model in \cref{tab:detectdiff}.}
\label{tab:controlled}
\footnotesize
\setlength{\tabcolsep}{4pt}
\begin{tabular}{@{}>{\raggedright\arraybackslash}p{4.6cm}cc@{}}
\toprule
 & \hq{Natural}{errors} & \hq{Across}{models}\\
Score & mean AUROC & share\\
\midrule
Detector (ours) & 0.849 & 0.634\\
Two-teacher probe & 0.829 & 0.605\\
Residual & 0.747 & 0.599 [0.550, 0.648]$^{\dagger}$\\
Content-shift teacher probe & 0.821 & 0.592\\
Sampling-teacher probe & 0.808 & 0.573\\
Linear probe, correctness$^{\ast}$ & 0.834 & 0.565\\
Linear probe, $1-\mathrm{IoU}$$^{\ast}$ & 0.828 & 0.581\\
SAPLMA$^{\ast}$ & 0.787 & 0.571\\
Pre-generation probe$^{\ast}$ & 0.811 & 0.550\\
Mahalanobis distance & 0.692 & 0.555\\
Coordinate log-prob & 0.670 & 0.520\\
MTLA (layers 8--21) & 0.589 & 0.505\\
\midrule
Detector minus linear probe, correctness & --- & 0.069 [0.024, 0.114]\\
Detector minus linear probe, $1-\mathrm{IoU}$ & --- & 0.054 [0.012, 0.095]\\
Detector minus SAPLMA & --- & 0.063 [0.014, 0.113]\\
Detector minus pre-generation probe & --- & 0.084 [0.030, 0.137]\\
Detector minus residual & --- & 0.036 [$-$0.009, 0.080]\\
Detector minus coordinate log-prob & --- & 0.114 [0.062, 0.166]\\
\bottomrule
\end{tabular}
\end{table}

%% file: tables/pairsrob.tex
\begin{table}[!htbp]
\centering
\caption{Robustness of the across-models comparison (1{,}146 new items common to five models): share of discordant cases in which the wrong model scores higher, for the detector and the supervised linear probe, and their difference with 95\% interval (bootstrap over images, 10,000 resamples). The difference is positive in 14 of the 14 variants, with an interval above zero in 13.}
\label{tab:pairsrob}
\footnotesize
\begin{tabular}{@{}>{\raggedright\arraybackslash}p{3.3cm}cccc@{}}
\toprule
Variant & Cases & Detector & \hq{Supervised}{linear} & Difference\\
\midrule
Main: IoU $<0.5$, percentiles & 1{,}224 & 0.634 & 0.565 & 0.069 [0.024, 0.114]\\
\midrule
Error at IoU $<0.3$ & 832 & 0.621 & 0.559 & 0.061 [0.011, 0.111]\\
Error at IoU $<0.4$ & 1{,}038 & 0.610 & 0.559 & 0.051 [0.006, 0.097]\\
Error at IoU $<0.6$ & 1{,}480 & 0.631 & 0.587 & 0.044 [0.003, 0.083]\\
Error at IoU $<0.7$ & 1{,}764 & 0.626 & 0.583 & 0.043 [0.006, 0.081]\\
\midrule
No percentiles, standardized by training items & 1{,}224 & 0.634 & 0.570 & 0.064 [0.021, 0.107]\\
No percentiles, standardized on the common items & 1{,}224 & 0.635 & 0.569 & 0.065 [0.021, 0.109]\\
\midrule
Without \qsmall & 734 & 0.648 & 0.562 & 0.086 [0.034, 0.138]\\
Without \qlarge & 703 & 0.620 & 0.585 & 0.036 [$-$0.011, 0.083]\\
Without \qfour & 792 & 0.632 & 0.569 & 0.063 [0.010, 0.116]\\
Without \qeight & 779 & 0.642 & 0.563 & 0.080 [0.028, 0.131]\\
Without \qivl & 664 & 0.628 & 0.547 & 0.081 [0.022, 0.141]\\
\midrule
Wrong model with a wrong-object error (IoU $<0.1$) & 406 & 0.640 & 0.560 & 0.080 [0.005, 0.158]\\
Wrong model with another error ($0.1\le$ IoU $<0.5$) & 818 & 0.631 & 0.568 & 0.064 [0.009, 0.118]\\
\bottomrule
\end{tabular}
\end{table}

%% file: tables/dprobe.tex
\begin{table}[!htbp]
\centering
\caption{A difficulty probe: a ridge probe on the model's hidden state ($h^{\mathrm{coord}}$, $\lambda=1{,}000$) trained, without labels, to predict the mean detector score of the other four models on the shared unlabelled training items; layer and penalty chosen by out-of-fold correlation with that target over five folds by image (Corr.). AUROC for all errors on the new items, with the supervised linear probe and the detector, and the difficulty probe minus the supervised linear probe with 95\% interval (bootstrap over images). Computed on a separate extraction of the hidden states that lacks two of the new items per model, so the detector of \qivl{} is 0.868 here and 0.867 in \cref{tab:detect}. On \qivl{} the supervised linear probe is high itself. Discordant cases (\cref{sup:ctl}): the difficulty probe wins 0.549 [0.500, 0.599], the detector 0.086 more.}
\label{tab:dprobe}
\footnotesize
\setlength{\tabcolsep}{2pt}
\begin{tabular}{@{}lcccccc@{}}
\toprule
Model & Layer & Corr. & \hq{Difficulty}{probe} & Supervised & Detector & \hq{Difficulty $-$}{supervised}\\
\midrule
\hqsmall & 7 & 0.849 & 0.816 [0.784, 0.845] & 0.820 & 0.829 & $-$0.004 [$-$0.028, 0.020]\\
\hqlarge & 5 & 0.848 & 0.812 [0.775, 0.848] & 0.802 & 0.848 & 0.010 [$-$0.022, 0.044]\\
\hqfour & 6 & 0.888 & 0.833 [0.801, 0.863] & 0.835 & 0.858 & $-$0.002 [$-$0.024, 0.020]\\
\hqeight & 6 & 0.876 & 0.836 [0.802, 0.866] & 0.835 & 0.837 & 0.000 [$-$0.023, 0.024]\\
\hqivl & 4 & 0.826 & 0.837 [0.804, 0.869] & 0.876 & 0.868 & $-$0.039 [$-$0.067, $-$0.013]\\
\bottomrule
\end{tabular}
\end{table}

%% file: tables/spredreplicationfull.tex
\begin{table}[!htbp]
\centering
\caption{The detector on the replication list, all rows (1{,}200 RefCOCOg test items on 901 images; \cref{sec:setup}). AUROC for all errors (IoU $<0.5$) unless named. W / P / E: wrong-object, partial-overlap and wrong-extent errors. Bound: one-sided 95\% lower bound of the difference to the supervised linear probe (5th percentile of a bootstrap over images); within 0.03: whether it exceeds $-0.03$. Probe alone, residual alone: the two parts of the detector. The held-out list in \cref{tab:detectho}, the new items and RefCOCO+ in \cref{tab:detect}.}
\label{tab:spredrepfull}
\scriptsize
\setlength{\tabcolsep}{1.4pt}
\begin{tabular}{@{}lcccccc@{}}
\toprule
& \shortstack{Qwen2.5-VL\\-3B} & \shortstack{Qwen2.5-VL\\-7B} & \shortstack{Qwen3-VL\\-4B} & \shortstack{Qwen3-VL\\-8B} & \shortstack{InternVL2.5\\-8B-MPO} & \shortstack{LLaVA-1.5\\-7B}\\
\midrule
Items / errors & 1{,}200 / 176 & 1{,}199 / 152 & 1{,}198 / 149 & 1{,}197 / 132 & 1{,}197 / 127 & 1{,}200 / 457\\
Errors by type: W / P / E & 76 / 39 / 61 & 55 / 39 / 58 & 72 / 34 / 43 & 60 / 36 / 36 & 45 / 34 / 48 & 127 / 114 / 216\\
Detector & 0.792 & 0.785 & 0.795 & 0.781 & 0.810 & 0.812\\
\quad probe alone & 0.794 & 0.765 & 0.782 & 0.764 & 0.810 & 0.820\\
\quad residual alone & 0.658 & 0.699 & 0.708 & 0.698 & 0.691 & 0.717\\
Coordinate log-prob & 0.633 & 0.665 & 0.616 & 0.615 & 0.612 & 0.658\\
Supervised linear probe & 0.778 & 0.769 & 0.765 & 0.780 & 0.793 & 0.819\\
Detector $-$ coordinate log-prob & 0.159 & 0.120 & 0.179 & 0.166 & 0.198 & 0.154\\
Detector $-$ supervised: bound & $-$0.011 & $-$0.017 & 0.008 & $-$0.024 & $-$0.008 & $-$0.021\\
\quad within 0.03 & yes & yes & yes & yes & yes & yes\\
Probe alone $-$ supervised: bound & $-$0.009 & $-$0.039 & $-$0.007 & $-$0.043 & $-$0.007 & $-$0.012\\
\quad within 0.03 & yes & no & yes & no & yes & yes\\
BICR (our reimpl., 2,000 labels) & 0.727 & 0.745 & 0.770 & 0.753 & 0.684 & 0.765\\
\midrule
Wrong object: detector & 0.863 & 0.815 & 0.831 & 0.815 & 0.757 & 0.894\\
\quad probe alone & 0.828 & 0.739 & 0.779 & 0.771 & 0.732 & 0.877\\
\quad residual alone & 0.802 & 0.824 & 0.830 & 0.802 & 0.744 & 0.845\\
\quad detector $-$ supervised & 0.075 & 0.020 & 0.049 & 0.019 & 0.033 & 0.024\\
\quad BICR & 0.787 & 0.774 & 0.819 & 0.826 & 0.639 & 0.832\\
\quad BICR + residual $-$ BICR & 0.051 & 0.067 & 0.057 & 0.033 & 0.082 & 0.042\\
\bottomrule
\end{tabular}
\end{table}

%% file: tables/detectho.tex
\begin{table}[!t]
\centering
\caption{The detector on the held-out list (1{,}200 RefCOCOg test items on 983 images; \cref{sec:setup}). AUROC for all errors (IoU $<0.5$) unless named. W / P / E: wrong-object, partial-overlap and wrong-extent errors. Bound: one-sided 95\% lower bound of the difference to the supervised linear probe (5th percentile of a bootstrap over images); within 0.03: whether it exceeds $-0.03$. Probe alone, residual alone: the two parts of the detector. Wrong-object errors: scored against correct answers. Error rate kept: among the 80\% of answers with the lowest score. BICR: our reimplementation, trained on the same 2,000 labelled items (\cref{sec:related}). Intervals in \cref{tab:heldout}; the replication list in \cref{tab:spredrep}, the new items and RefCOCO+ in \cref{tab:detect}.}
\label{tab:detectho}
\scriptsize
\setlength{\tabcolsep}{1.4pt}
\begin{tabular}{@{}lcccccc@{}}
\toprule
& \shortstack{Qwen2.5-VL\\-3B} & \shortstack{Qwen2.5-VL\\-7B} & \shortstack{Qwen3-VL\\-4B} & \shortstack{Qwen3-VL\\-8B} & \shortstack{InternVL2.5\\-8B-MPO} & \shortstack{LLaVA-1.5\\-7B}\\
\midrule
Items / errors & 1{,}200 / 178 & 1{,}199 / 152 & 1{,}200 / 140 & 1{,}199 / 130 & 1{,}199 / 142 & 1{,}200 / 480\\
Errors by type: W / P / E & 68 / 45 / 65 & 52 / 36 / 64 & 64 / 38 / 38 & 57 / 33 / 40 & 63 / 39 / 40 & 121 / 121 / 238\\
Detector & 0.838 & 0.765 & 0.817 & 0.822 & 0.829 & 0.780\\
\quad probe alone & 0.823 & 0.754 & 0.794 & 0.823 & 0.807 & 0.779\\
\quad residual alone & 0.739 & 0.683 & 0.736 & 0.721 & 0.744 & 0.713\\
Coordinate log-prob & 0.670 & 0.653 & 0.634 & 0.665 & 0.596 & 0.659\\
Supervised linear probe & 0.816 & 0.739 & 0.803 & 0.832 & 0.807 & 0.794\\
Detector $-$ coordinate log-prob & 0.168 & 0.113 & 0.182 & 0.157 & 0.233 & 0.121\\
Detector $-$ supervised: bound & $-$0.004 & $-$0.007 & $-$0.012 & $-$0.034 & $-$0.003 & $-$0.031\\
\quad within 0.03 & yes & yes & yes & no & yes & no\\
Probe alone $-$ supervised: bound & $-$0.018 & $-$0.021 & $-$0.038 & $-$0.034 & $-$0.026 & $-$0.029\\
\quad within 0.03 & yes & yes & no & no & yes & yes\\
BICR (our reimpl., 2,000 labels) & 0.746 & 0.747 & 0.794 & 0.770 & 0.725 & 0.742\\
Detector $-$ BICR & 0.091 & 0.018 & 0.022 & 0.052 & 0.104 & 0.038\\
\midrule
Wrong object: detector & 0.886 & 0.757 & 0.846 & 0.832 & 0.852 & 0.884\\
\quad probe alone & 0.823 & 0.694 & 0.788 & 0.814 & 0.799 & 0.873\\
\quad residual alone & 0.865 & 0.770 & 0.839 & 0.780 & 0.841 & 0.826\\
\quad detector $-$ supervised & 0.084 & $-$0.036 & 0.052 & 0.026 & 0.054 & 0.030\\
\quad BICR & 0.778 & 0.802 & 0.806 & 0.772 & 0.705 & 0.814\\
\quad BICR + residual $-$ BICR & 0.090 & 0.030 & 0.066 & 0.045 & 0.116 & 0.045\\
\midrule
Error rate, all answers & 0.148 & 0.127 & 0.117 & 0.108 & 0.118 & 0.400\\
Error rate kept: detector & 0.076 & 0.081 & 0.062 & 0.050 & 0.064 & 0.311\\
\quad supervised & 0.085 & 0.083 & 0.061 & 0.053 & 0.065 & 0.302\\
\quad coordinate log-prob & 0.119 & 0.104 & 0.092 & 0.080 & 0.097 & 0.339\\
\bottomrule
\end{tabular}
\end{table}

%% file: tables/heldout.tex
\begin{sidewaystable}
\centering
\caption{Held-out list: 1{,}200 RefCOCOg test items on 983 images shared with none of the training, new or RefCOCO+ items (36 images also occur in three early development lists), scored with the detectors, probes and reference scores fitted earlier (\cref{sup:heldout}). AUROC for all errors and differences of the detector to the supervised linear probe and to coordinate log-prob, with 95\% intervals from a bootstrap over images (10,000 resamples); last rows: wrong-object errors against correct answers. Items: items with a parsed answer. One-sided 95\% lower bound: 5th percentile of the same bootstrap; the margin of 0.03 is that of \cref{tab:boundsupervised}.}
\label{tab:heldout}
\footnotesize
\setlength{\tabcolsep}{3pt}
\begin{tabular}{@{}lcccccc@{}}
\toprule
& \shortstack{Qwen2.5-VL\\-3B} & \shortstack{Qwen2.5-VL\\-7B} & \shortstack{Qwen3-VL\\-4B} & \shortstack{Qwen3-VL\\-8B} & \shortstack{InternVL2.5\\-8B-MPO} & \shortstack{LLaVA-1.5\\-7B}\\
\midrule
Items & 1{,}200 & 1{,}199 & 1{,}200 & 1{,}199 & 1{,}199 & 1{,}200\\
Errors & 178 & 152 & 140 & 130 & 142 & 480\\
Detector & 0.838 & 0.765 & 0.817 & 0.822 & 0.829 & 0.780\\
\quad 95\% interval & {\scriptsize[0.807, 0.867]} & {\scriptsize[0.724, 0.805]} & {\scriptsize[0.781, 0.851]} & {\scriptsize[0.781, 0.860]} & {\scriptsize[0.795, 0.862]} & {\scriptsize[0.753, 0.806]}\\
Supervised linear probe & 0.816 & 0.739 & 0.803 & 0.832 & 0.807 & 0.794\\
\quad 95\% interval & {\scriptsize[0.784, 0.847]} & {\scriptsize[0.699, 0.779]} & {\scriptsize[0.765, 0.839]} & {\scriptsize[0.797, 0.865]} & {\scriptsize[0.766, 0.845]} & {\scriptsize[0.768, 0.819]}\\
Coordinate log-prob & 0.670 & 0.653 & 0.634 & 0.665 & 0.596 & 0.659\\
\quad 95\% interval & {\scriptsize[0.627, 0.712]} & {\scriptsize[0.607, 0.698]} & {\scriptsize[0.585, 0.682]} & {\scriptsize[0.611, 0.716]} & {\scriptsize[0.539, 0.651]} & {\scriptsize[0.626, 0.690]}\\
Detector $-$ supervised linear & 0.022 & 0.026 & 0.014 & $-$0.010 & 0.022 & $-$0.014\\
\quad 95\% interval & {\scriptsize[$-$0.009, 0.051]} & {\scriptsize[$-$0.013, 0.065]} & {\scriptsize[$-$0.017, 0.046]} & {\scriptsize[$-$0.038, 0.017]} & {\scriptsize[$-$0.007, 0.053]} & {\scriptsize[$-$0.034, 0.005]}\\
\quad one-sided 95\% lower bound & $-$0.004 & $-$0.007 & $-$0.012 & $-$0.034 & $-$0.003 & $-$0.031\\
\quad lower bound above $-0.03$ & yes & yes & yes & no & yes & no\\
Probe alone $-$ supervised linear: one-sided bound & $-$0.018 & $-$0.021 & $-$0.038 & $-$0.034 & $-$0.026 & $-$0.029\\
\quad lower bound above $-0.03$ & yes & yes & no & no & yes & yes\\
Detector $-$ coordinate log-prob & 0.168 & 0.113 & 0.182 & 0.157 & 0.233 & 0.121\\
\quad 95\% interval & {\scriptsize[0.125, 0.211]} & {\scriptsize[0.062, 0.164]} & {\scriptsize[0.134, 0.233]} & {\scriptsize[0.103, 0.213]} & {\scriptsize[0.171, 0.296]} & {\scriptsize[0.086, 0.157]}\\
\midrule
Wrong-object errors & 68 & 52 & 64 & 57 & 63 & 121\\
Detector $-$ supervised linear & 0.084 & $-$0.036 & 0.052 & 0.026 & 0.054 & 0.030\\
\quad 95\% interval & {\scriptsize[0.045, 0.124]} & {\scriptsize[$-$0.091, 0.017]} & {\scriptsize[0.010, 0.095]} & {\scriptsize[$-$0.014, 0.068]} & {\scriptsize[0.007, 0.103]} & {\scriptsize[0.002, 0.060]}\\
\bottomrule
\end{tabular}
\end{sidewaystable}

%% file: tables/residheld.tex
\begin{sidewaystable}
\centering
\caption{The residual as an addition on the held-out list: AUROC of base $+\,0.5\times$ residual minus that of the base alone, with the same standardization as in \cref{tab:residadddiff}, for wrong-object errors against correct answers and for all errors, with 95\% intervals from a bootstrap over images. The six bases are those of \cref{tab:residadddiff}, five probes and coordinate log-prob. Dashes: the pre-generation probe of \qlarge{} is the same probe as its linear probe on correctness, as in \cref{tab:residadddiff}, so each block has 35 distinct model--base cells. Wrong-object errors: increment positive in 35 and with an interval above zero in 31; all errors: positive in 32, above zero in 20 and below zero in 1. $^{\ast}$Supervised with the 2,000 training items.}
\label{tab:residheld}
\scriptsize
\setlength{\tabcolsep}{2.5pt}
\begin{tabular}{@{}lcccccc@{}}
\toprule
Base & \qsmall{} & \qlarge{} & \qfour{} & \qeight{} & \qivl{} & \qllava{}\\
\midrule
\multicolumn{7}{@{}l}{\emph{Wrong-object errors}}\\
Two-teacher probe (ours) & 0.063 {\tiny[0.035, 0.093]} & 0.063 {\tiny[0.029, 0.099]} & 0.058 {\tiny[0.034, 0.084]} & 0.018 {\tiny[$-$0.007, 0.042]} & 0.053 {\tiny[0.030, 0.079]} & 0.012 {\tiny[$-$0.004, 0.029]}\\
Sampling-teacher probe & 0.074 {\tiny[0.049, 0.099]} & 0.025 {\tiny[$-$0.002, 0.051]} & 0.071 {\tiny[0.044, 0.101]} & 0.041 {\tiny[0.014, 0.071]} & 0.066 {\tiny[0.039, 0.096]} & 0.022 {\tiny[0.005, 0.040]}\\
Linear probe, correctness$^{\ast}$ & 0.057 {\tiny[0.033, 0.083]} & 0.032 {\tiny[0.008, 0.056]} & 0.046 {\tiny[0.025, 0.068]} & 0.008 {\tiny[$-$0.016, 0.032]} & 0.056 {\tiny[0.029, 0.084]} & 0.023 {\tiny[0.004, 0.043]}\\
Pre-generation probe$^{\ast}$ & 0.062 {\tiny[0.041, 0.084]} & -- & 0.029 {\tiny[0.009, 0.050]} & 0.027 {\tiny[0.002, 0.053]} & 0.089 {\tiny[0.061, 0.118]} & 0.037 {\tiny[0.017, 0.060]}\\
SAPLMA$^{\ast}$ & 0.057 {\tiny[0.035, 0.080]} & 0.069 {\tiny[0.031, 0.108]} & 0.046 {\tiny[0.025, 0.070]} & 0.032 {\tiny[0.002, 0.063]} & 0.055 {\tiny[0.034, 0.077]} & 0.039 {\tiny[0.021, 0.058]}\\
Coordinate log-prob & 0.145 {\tiny[0.112, 0.179]} & 0.111 {\tiny[0.065, 0.157]} & 0.142 {\tiny[0.104, 0.182]} & 0.113 {\tiny[0.070, 0.158]} & 0.143 {\tiny[0.108, 0.179]} & 0.132 {\tiny[0.101, 0.164]}\\
\addlinespace
\multicolumn{7}{@{}l}{\emph{All errors}}\\
Two-teacher probe (ours) & 0.015 {\tiny[$-$0.002, 0.033]} & 0.011 {\tiny[$-$0.007, 0.029]} & 0.023 {\tiny[0.005, 0.040]} & $-$0.000 {\tiny[$-$0.017, 0.016]} & 0.022 {\tiny[0.008, 0.038]} & 0.001 {\tiny[$-$0.010, 0.011]}\\
Sampling-teacher probe & 0.033 {\tiny[0.016, 0.051]} & 0.009 {\tiny[$-$0.008, 0.026]} & 0.030 {\tiny[0.011, 0.049]} & 0.011 {\tiny[$-$0.006, 0.029]} & 0.028 {\tiny[0.011, 0.046]} & 0.002 {\tiny[$-$0.010, 0.013]}\\
Linear probe, correctness$^{\ast}$ & 0.011 {\tiny[$-$0.005, 0.027]} & 0.016 {\tiny[0.000, 0.032]} & 0.009 {\tiny[$-$0.007, 0.024]} & $-$0.017 {\tiny[$-$0.033, $-$0.001]} & 0.023 {\tiny[0.007, 0.041]} & $-$0.000 {\tiny[$-$0.012, 0.012]}\\
Pre-generation probe$^{\ast}$ & 0.029 {\tiny[0.012, 0.045]} & -- & 0.004 {\tiny[$-$0.010, 0.019]} & 0.003 {\tiny[$-$0.014, 0.019]} & 0.043 {\tiny[0.025, 0.062]} & 0.013 {\tiny[0.002, 0.025]}\\
SAPLMA$^{\ast}$ & 0.022 {\tiny[0.006, 0.037]} & 0.043 {\tiny[0.020, 0.066]} & 0.012 {\tiny[$-$0.006, 0.030]} & 0.005 {\tiny[$-$0.014, 0.025]} & 0.025 {\tiny[0.009, 0.041]} & 0.017 {\tiny[0.006, 0.028]}\\
Coordinate log-prob & 0.075 {\tiny[0.054, 0.096]} & 0.058 {\tiny[0.033, 0.083]} & 0.088 {\tiny[0.062, 0.114]} & 0.070 {\tiny[0.045, 0.096]} & 0.094 {\tiny[0.071, 0.118]} & 0.067 {\tiny[0.051, 0.084]}\\
\bottomrule
\end{tabular}
\end{sidewaystable}

%% file: tables/heldoutpairs.tex
\begin{table}[!htbp]
\centering
\caption{Cross-model discordant cases on the held-out list: share in which the wrong model has the higher score percentile, measured and predicted. The five models of the main analyses, 1{,}197 items common to them on 981 images, 1{,}218 cases, predicted by \cref{eq:twocomp}. Last row: mean absolute difference between measured and predicted over the twelve scores.}
\label{tab:heldoutpairs}
\footnotesize
\begin{tabular}{@{}lcc@{}}
\toprule
Score & Measured & \cref{eq:twocomp}\\
\midrule
Detector & 0.647 & 0.667\\
Two-teacher probe & 0.614 & 0.627\\
Self-prediction residual & 0.642 & 0.657\\
Sampling-teacher probe & 0.555 & 0.545\\
Shift-teacher probe & 0.576 & 0.588\\
Supervised linear probe & 0.601 & 0.601\\
Supervised probe, $1-\mathrm{IoU}$ & 0.581 & 0.604\\
SAPLMA & 0.597 & 0.600\\
Pre-generation probe & 0.564 & 0.554\\
Coordinate log-prob & 0.564 & 0.566\\
Mahalanobis distance & 0.532 & 0.522\\
Content-shift check & 0.603 & 0.640\\
\midrule
Mean absolute difference & --- & 0.013\\
\bottomrule
\end{tabular}
\end{table}

%% file: sections/supplement_replication.tex
\section{Replication list}
\label{sup:replication}

\paragraph{The list}
\begin{sloppypar}
The replication list holds 1{,}200 RefCOCOg test items on 901 images, drawn with a fixed seed by the rule used for the held-out list (\cref{sup:heldout}) from images that occur in none of the training items, the new items, the RefCOCO+ items, the development lists or the held-out list. A script compared its image identifiers with those of every score, answer and feature file we hold (147 sources, 8{,}832 images) and found no image in common; the list was fixed by this check before any model was run on it.
\end{sloppypar}

\paragraph{Evaluation}
Rows marked post hoc were added after the list had been scored. Every probe, read-out and reference score was the one fitted earlier on the training items and was applied unchanged, with the scripts used for the held-out list. The generation log of \qivl{}, seen when the answers were retrieved, includes its error rate on the list. The numbers in the abstract and the contributions are those of this list unless another set is named. Every model has at least 45 wrong-object errors. \Cref{tab:replication} lists the comparisons on the list with their results.

\paragraph{Weaker results on this list}
Several results on this list are weaker than on the held-out list; some were not run there. Added to hidden-state probes, the residual lowers their detection of partial-overlap and wrong-extent errors, with a combined interval below zero for three and one of the seven probes, and the detector is lower than its probe on these two types, with the combined interval below zero for partial overlap (\cref{sec:plug:add}). The displacement of the read-out towards the nearest other object of the target's class has an interval above zero (0.064 [0.011, 0.116]), so the read-out leans far more towards the annotated object than towards that object, but not towards it alone. On \qfour{} the new box of a changed answer has a slope above 0.6, but the permuted control reaches 0.170, above 0.1. Against the two-position ridge probe the one-sided bound of the detector exceeds $-0.03$, but the detector is not ahead: both intervals include zero, and on wrong-object errors the detector is below it on two of the six models. The supervised linear probe is slightly above the detector on \qllava{} ($-$0.007), and BICR is slightly above it on wrong-object errors of \qeight{} ($-$0.010). With 250 labelled items the intervals of \qfour{} and \qeight{} include zero.

\paragraph{Not tested on this list}
Three results reported in the paper were not computed on this list: the scores on annotations confirmed by an independent re-check, because the re-check covers neither list; the dependence of the gain on target size in units of the read-out error, because targets narrower than twice that error are about 1\% of the RefCOCO family; and variation on the training side beyond the ten existing refits. %

{\scriptsize
\setlength{\tabcolsep}{2pt}
\setlength{\LTcapwidth}{\textwidth}
\begin{longtable}{@{}>{\raggedright\arraybackslash}p{4.4cm}>{\raggedright\arraybackslash}p{3.3cm}>{\raggedright\arraybackslash}p{3.3cm}@{}}
\caption{Comparisons on the replication list (1{,}200 RefCOCOg test items on 901 images) and their results, next to the held-out list. Intervals: image resampling, 10,000 draws (2,000 in the rows on the fifteen scores, on review and on other label-free signals); combined intervals as in \cref{sup:diff}. Bold: number of probes with a combined interval below zero, and a difference with its interval below zero. Notes a--f below the table.}
\label{tab:replication}\\
\toprule
Comparison & Held-out list & Replication list\\
\midrule
\endfirsthead
\multicolumn{3}{@{}l}{\cref{tab:replication} (continued)}\\
\toprule
Comparison & Held-out list & Replication list\\
\midrule
\endhead
\midrule
\multicolumn{3}{r@{}}{continued on the next page}\\
\endfoot
\bottomrule
\endlastfoot
\multicolumn{3}{@{}l}{\textit{The residual added to existing scores, wrong-object errors (\cref{sec:plug:add})}}\\
Residual added to each of the fifteen scores & 15 of 15; lowest 0.029 [0.012, 0.047]; 85 of 85 cells positive & 15 of 15; lowest 0.026 [0.007, 0.045]; 85 of 85 cells positive\\
\quad mean over the scores, per model & lowest lower end 0.021 (\qllava) & lowest lower end 0.020 (\qivl)\\
Gain by error type, mean over seven hidden-state probes (probes with a combined interval below 0): wrong object$^{f}$ & 0.043 (0 of 7) & 0.039 (0 of 7)\\
\quad partial overlap$^{f}$ & $-$0.002 (0 of 7) & $-$0.013 (\textbf{3 of 7})\\
\quad wrong extent & $-$0.009 (0 of 7) & $-$0.012 (\textbf{1 of 7})\\
\quad all errors & 0.014 (0 of 7) & 0.008 (0 of 7)\\
Top-scoring fifth sent to review: share of wrong-object errors caught & 0.474 $\to$ 0.574, 0.100 [0.075, 0.124]; positive for 15 of 15 & 0.469 $\to$ 0.575, 0.106 [0.080, 0.132]; positive for 15 of 15\\
\quad share of all errors caught & 0.043 [0.031, 0.055] & 0.032 [0.018, 0.047]\\
Gain from the residual minus gain from another label-free signal (five signals) & 0.052 to 0.086 & 0.056 to 0.097\\
\quad against the two-teacher probe & 0.022 [0.007, 0.037] & 0.022 [0.008, 0.037]\\
Telling wrong-object errors from other errors (wrong answers only): box geometry with the residual minus box geometry & 0.054 [0.025, 0.083] & 0.061 [0.031, 0.092]\\
\quad residual alone & 0.703 & 0.726\\
Residual added to probes reading two positions: correctness labels; wrong-object labels & 0.032 [0.015, 0.049]; 0.031 [0.018, 0.045] & 0.030 [0.013, 0.048]; 0.025 [0.013, 0.037]\\
Supervised model given both boxes and their IoU: gain from the residual (two label kinds); it alone, residual alone & 0.005, 0.003; 0.820, 0.820 & 0.011, 0.009; 0.791, 0.808\\
Residual added to BICR & 0.065 [0.051, 0.080] & 0.055 [0.040, 0.071]\\
\midrule
\multicolumn{3}{@{}l}{\textit{The detector, all errors (\cref{sec:plug:detector})}}\\
Detector minus coordinate log-prob$^{a}$ & 0.162 [0.134, 0.191]; all six above 0 & 0.163 [0.133, 0.193]; all six above 0\\
Detector against the 2,000-label linear probe: models with one-sided bound $\ge -0.03$ & 4 of 6 & 6 of 6\\
\quad probe alone & 4 of 6 & 4 of 6\\
Detector minus BICR & 0.054 (bound 0.038) & 0.055 (bound 0.039)\\
Detector minus two-position ridge probe (correctness labels): all errors; wrong-object errors & 0.006 (bound $-$0.017); 0.022 (bound $-$0.012) & 0.009 (bound $-$0.011); 0.017 (bound $-$0.017)\\
250 labelled items: detector minus supervised linear, both fitted on them (20 subsets) & not run on this list (new items: 0.020 to 0.093) & 0.016 to 0.086\\
One error type left out of training: detector minus supervised, five models; wrong object, overlap & not run on this list (new items: 0.102, 0.085) & 0.132 [0.098, 0.168], 0.069 [0.040, 0.100]; probe alone on overlap 0.788\\
\midrule
\multicolumn{3}{@{}l}{\textit{What the residual measures, wrong-object errors (\cref{sec:measure})}}\\
Residual alone & 0.820 [0.789, 0.852] & 0.808 [0.774, 0.840]\\
Own read-out against other items' read-outs$^{b}$ & other items 0.560--0.647 & other items 0.545--0.626\\
New box of a changed answer moves with the read-out: models with slope $\ge0.6$, lower end $>0.3$ and permuted $|\cdot|<0.1$$^{c}$ & 5 of 6 (\qllava{}: slope 0.502) & 5 of 6 (\qfour{}: permuted 0.170)\\
Read-out displaced towards the annotated object$^{d}$ & 0.232 [0.194, 0.275] pooled; 0.164--0.286 per model & 0.250 [0.212, 0.292] pooled; 0.192--0.313 per model\\
\quad minus towards the nearest same-class object$^{e}$ & 0.193 [0.132, 0.255]; correct answers $-$0.006 [$-$0.015, 0.004] & 0.189 [0.122, 0.269]; correct answers $-$0.004 [$-$0.015, 0.006]\\
Box decoded by the output head minus residual & not run on this list (new items: $-$0.105 to $-$0.357) & $-$0.135 to $-$0.242\\
Read-out fitted to annotated boxes minus fitted to own answers: residual alone, all errors (three models) & not run on this list & 0.011, \textbf{$-$0.019}, $-$0.003\\
\quad detector, all errors (\qllava{} added; post hoc) & not run on this list & 0.004, \textbf{$-$0.009}, $-$0.001; 0.002\\
Share of wrong-object errors among all errors & 41.0\%; \qllava{} 25.2\% & 41.8\%; \qllava{} 27.8\%\\
\end{longtable}
\par\noindent
$^{a}$ The interval of the six-model mean was computed post hoc; the per-model values do not depend on it.
$^{b}$ The own value is compared with the range of the permuted read-outs, which the script reports in place of an interval of the difference.
$^{c}$ The model not counted is \qllava{} on the held-out list and \qfour{} on the replication list.
$^{d}$ Pooled coefficient from the same-class analysis, which pools the six models (\cref{tab:sameclass}); per-model values from the per-model regression of \cref{sec:measure:pull}.
$^{e}$ On the replication list the coefficient towards the same-class object is itself above zero (0.064 [0.011, 0.116]); the row reports the difference.
$^{f}$ The rows for wrong-object and partial-overlap errors were added afterwards (post hoc).\par}

%% file: main.bbl
\begin{thebibliography}{10}
\expandafter\ifx\csname url\endcsname\relax
  \def\url#1{\texttt{#1}}\fi
\expandafter\ifx\csname href\endcsname\relax
  \def\href#1#2{#2} \def\path#1{#1}\fi

\bibitem{bai2025qwen25vl}
S.~Bai, K.~Chen, X.~Liu, J.~Wang, W.~Ge, S.~Song, K.~Dang, P.~Wang, S.~Wang,
  J.~Tang, H.~Zhong, Y.~Zhu, M.~Yang, Z.~Li, J.~Wan, et~al., {Qwen2.5-VL}
  technical report (2025).
\newblock \href {https://doi.org/10.48550/arXiv.2502.13923}
  {\path{doi:10.48550/arXiv.2502.13923}}.

\bibitem{ax2511.21631}
S.~Bai, Y.~Cai, R.~Chen, K.~Chen, X.~Chen, Z.~Cheng, L.~Deng, W.~Ding, C.~Gao,
  C.~Ge, W.~Ge, Z.~Guo, Q.~Huang, J.~Huang, F.~Huang, et~al., {Qwen3-VL}
  technical report (2025).
\newblock \href {https://doi.org/10.48550/arXiv.2511.21631}
  {\path{doi:10.48550/arXiv.2511.21631}}.

\bibitem{zhang2018excitation}
J.~Zhang, S.~A. Bargal, Z.~Lin, J.~Brandt, X.~Shen, S.~Sclaroff, Top-down
  neural attention by excitation backprop, International Journal of Computer
  Vision 126~(10) (2018) 1084--1102.
\newblock \href {https://doi.org/10.1007/s11263-017-1059-x}
  {\path{doi:10.1007/s11263-017-1059-x}}.

\bibitem{fumera2000reject}
G.~Fumera, F.~Roli, G.~Giacinto, Reject option with multiple thresholds,
  Pattern Recognition 33~(12) (2000) 2099--2101.
\newblock \href {https://doi.org/10.1016/S0031-3203(00)00059-5}
  {\path{doi:10.1016/S0031-3203(00)00059-5}}.

\bibitem{hendrycks2017msp}
D.~Hendrycks, K.~Gimpel, A baseline for detecting misclassified and
  out-of-distribution examples in neural networks, in: International Conference
  on Learning Representations (ICLR), 2017.

\bibitem{jiang2018trustscore}
H.~Jiang, B.~Kim, M.~Y. Guan, M.~Gupta, To trust or not to trust a classifier,
  in: Advances in Neural Information Processing Systems, 2018.

\bibitem{ax2606.25760}
D.~Kumar, S.~Tayebati, D.~Naik, A.~S. Rios, N.~Ahuja, O.~Tickoo, R.~Krishnan,
  A.~R. Trivedi, Uncertainty quantification for computer-use agents: A
  benchmark across vision-language models and {GUI} grounding datasets, in:
  Advances in Neural Information Processing Systems, 2026, arXiv:2606.25760.

\bibitem{ax2607.05978}
D.~Shalam, E.~B. Baruch, A.~B. Cohen, T.~Remez, Propose and attend:
  Training-free {MLLM} grounding confidence via multi-token localized attention
  (2026).
\newblock \href {https://doi.org/10.48550/arXiv.2607.05978}
  {\path{doi:10.48550/arXiv.2607.05978}}.

\bibitem{li2026pace}
Y.~Li, X.~Wu, H.~M. Meng, How many pixels is a digit worth? {P}lace-aware
  coordinate entropy for {GUI} agent confidence estimation, in: Proceedings of
  the 2026 Conference on Empirical Methods in Natural Language Processing,
  2026, pp. 10302--10315.
\newblock \href {https://doi.org/10.18653/2026.emnlp-main.448}
  {\path{doi:10.18653/2026.emnlp-main.448}}.

\bibitem{wang2023selfconsistency}
X.~Wang, J.~Wei, D.~Schuurmans, Q.~V. Le, E.~H. Chi, S.~Narang, A.~Chowdhery,
  D.~Zhou, Self-consistency improves chain of thought reasoning in language
  models, in: International Conference on Learning Representations ({ICLR}),
  2023.

\bibitem{kuhn2023semantic}
L.~Kuhn, Y.~Gal, S.~Farquhar, Semantic uncertainty: Linguistic invariances for
  uncertainty estimation in natural language generation, in: International
  Conference on Learning Representations, 2023.

\bibitem{zhang2026mvp}
Y.~Zhang, Z.~Pan, Z.~Zeng, S.~Shen, C.~Meng, L.~Zhu, {MVP}: Multiple view
  prediction improves {GUI} grounding, in: Proceedings of the IEEE/CVF
  Conference on Computer Vision and Pattern Recognition (CVPR), 2026, pp.
  27482--27492, arXiv:2512.08529.

\bibitem{ax2609.37488}
T.~Sato, T.~Sanda, K.~Maeda, T.~Ogawa, M.~Haseyama, S.~Nagashima, {FORUM}:
  Frozen outputs reconciled using model agreement for visual grounding, in:
  Asian Conference on Computer Vision (ACCV), 2026.

\bibitem{ax2505.17656}
H.~Tan, F.~Sun, S.~Liu, D.~Su, Q.~Cao, X.~Chen, J.~Wang, X.~Cai, Y.~Wang,
  H.~Shen, X.~Cheng, Too consistent to detect: A study of self-consistent
  errors in {LLMs}, in: Proceedings of the 2025 Conference on Empirical Methods
  in Natural Language Processing, 2025, pp. 4755--4765.
\newblock \href {http://arxiv.org/abs/2505.17656} {\path{arXiv:2505.17656}},
  \href {https://doi.org/10.18653/v1/2025.emnlp-main.238}
  {\path{doi:10.18653/v1/2025.emnlp-main.238}}.

\bibitem{orgad2025know}
H.~Orgad, M.~Toker, Z.~Gekhman, R.~Reichart, I.~Szpektor, H.~Kotek,
  Y.~Belinkov, {LLMs} know more than they show: On the intrinsic representation
  of {LLM} hallucinations, in: International Conference on Learning
  Representations, 2025, arXiv:2410.02707.

\bibitem{azaria2023saplma}
A.~Azaria, T.~Mitchell, The internal state of an {LLM} knows when it's lying,
  in: Findings of the Association for Computational Linguistics: EMNLP 2023,
  2023, pp. 967--976.
\newblock \href {https://doi.org/10.18653/v1/2023.findings-emnlp.68}
  {\path{doi:10.18653/v1/2023.findings-emnlp.68}}.

\bibitem{ax2605.10893}
R.~Khanmohammadi, E.~Miahi, S.~Kaur, C.~Smiley, I.~Brugere, K.~S. Thind, M.~M.
  Ghassemi, Grounded or guessing? {LVLM} confidence estimation via blind-image
  contrastive ranking, in: Proceedings of the 15th International Joint
  Conference on Natural Language Processing and the 5th Conference of the
  Asia-Pacific Chapter of the Association for Computational Linguistics (Volume
  1: Long Papers), Association for Computational Linguistics, 2026, pp.
  4645--4721.
\newblock \href {https://doi.org/10.18653/2026.aacl-long.210}
  {\path{doi:10.18653/2026.aacl-long.210}}.

\bibitem{sun2025arith}
Y.~Sun, A.~Stolfo, M.~Sachan, Probing for arithmetic errors in language models,
  in: Proceedings of the 2025 Conference on Empirical Methods in Natural
  Language Processing, Association for Computational Linguistics, 2025, pp.
  8122--8139.
\newblock \href {https://doi.org/10.18653/v1/2025.emnlp-main.411}
  {\path{doi:10.18653/v1/2025.emnlp-main.411}}.

\bibitem{elshangiti2026count}
A.~O. El-Shangiti, A.~Nurgazy, H.~AlQuabeh, N.~Rozanov, K.~Inui, The count is
  there, but misaligned: Understanding and correcting counting failures in
  {VLMs} (2026).
\newblock \href {https://doi.org/10.48550/arXiv.2607.09544}
  {\path{doi:10.48550/arXiv.2607.09544}}.

\bibitem{ax2609.11216}
M.~V.~S. Ravulapalli, S.~S. Chadha, A.~M. Hari, Legible failures: Detecting and
  repairing in-context binding errors (2026).
\newblock \href {https://doi.org/10.48550/arXiv.2609.11216}
  {\path{doi:10.48550/arXiv.2609.11216}}.

\bibitem{ax2610.13239}
X.~Zhu, K.~Lu, H.~Wu, Z.~Du, D.~Zhang, J.~Xue, Jag: Direct box prediction for
  efficient visual grounding (2026).
\newblock \href {https://doi.org/10.48550/arXiv.2610.13239}
  {\path{doi:10.48550/arXiv.2610.13239}}.

\bibitem{ax2610.35053}
M.~Talha, {LOCUS}: Locate once, check until settled. {D}raft-and-verify visual
  grounding with a box head and coordinate-token verification (2026).
\newblock \href {https://doi.org/10.48550/arXiv.2610.35053}
  {\path{doi:10.48550/arXiv.2610.35053}}.

\bibitem{ax2502.13329}
D.~Ashok, J.~May, Language models can predict their own behavior, in: Advances
  in Neural Information Processing Systems, Vol.~38, 2025, pp. 69806--69837.
\newblock \href {https://doi.org/10.52202/085713-2102}
  {\path{doi:10.52202/085713-2102}}.

\bibitem{belrose2023tunedlens}
N.~Belrose, I.~Ostrovsky, L.~McKinney, Z.~Furman, L.~Smith, D.~Halawi,
  S.~Biderman, J.~Steinhardt, Eliciting latent predictions from transformers
  with the tuned lens (2023).
\newblock \href {https://doi.org/10.48550/arXiv.2303.08112}
  {\path{doi:10.48550/arXiv.2303.08112}}.

\bibitem{kang2025heads}
S.~Kang, J.-Y. Kim, J.~Kim, S.~J. Hwang, Your large vision-language model only
  needs a few attention heads for visual grounding, in: Proceedings of the
  IEEE/CVF Conference on Computer Vision and Pattern Recognition (CVPR), 2025,
  pp. 9339--9350.
\newblock \href {https://doi.org/10.1109/cvpr52734.2025.00872}
  {\path{doi:10.1109/cvpr52734.2025.00872}}.

\bibitem{phukan2025contextuallens}
A.~Phukan, D.~Divyansh, H.~K. Morj, V.~Vaishnavi, A.~Saxena, K.~Goswami, Beyond
  logit lens: Contextual embeddings for robust hallucination detection \&
  grounding in {VLMs}, in: Proceedings of the 2025 Conference of the Nations of
  the Americas Chapter of the Association for Computational Linguistics: Human
  Language Technologies (Volume 1: Long Papers), 2025, pp. 9661--9675.
\newblock \href {https://doi.org/10.18653/v1/2025.naacl-long.488}
  {\path{doi:10.18653/v1/2025.naacl-long.488}}.

\bibitem{kossen2024sep}
J.~Kossen, J.~Han, M.~Razzak, L.~Schut, S.~Malik, Y.~Gal, Semantic entropy
  probes: Robust and cheap hallucination detection in {LLMs} (2024).
\newblock \href {https://doi.org/10.48550/arXiv.2406.15927}
  {\path{doi:10.48550/arXiv.2406.15927}}.

\bibitem{ax2604.19444}
T.~Zollo, J.~Wang, R.~Zemel, Binary agreement for efficient unsupervised
  confidence calibration of reasoning {LLMs} (2026).
\newblock \href {https://doi.org/10.48550/arXiv.2604.19444}
  {\path{doi:10.48550/arXiv.2604.19444}}.

\bibitem{chen2019whitebox}
T.~Chen, J.~Navr{\'a}til, V.~Iyengar, K.~Shanmugam, Confidence scoring using
  whitebox meta-models with linear classifier probes, in: Proceedings of the
  22nd International Conference on Artificial Intelligence and Statistics
  (AISTATS), Vol.~89 of Proceedings of Machine Learning Research, 2019, pp.
  1467--1475, arXiv:1805.05396.

\bibitem{pal2023futurelens}
K.~Pal, J.~Sun, A.~Yuan, B.~Wallace, D.~Bau, Future lens: Anticipating
  subsequent tokens from a single hidden state, in: Proceedings of the 27th
  Conference on Computational Natural Language Learning (CoNLL), 2023, pp.
  548--560, arXiv:2311.04897.
\newblock \href {https://doi.org/10.18653/v1/2023.conll-1.37}
  {\path{doi:10.18653/v1/2023.conll-1.37}}.

\bibitem{ke2025graphrec}
J.~Ke, Q.~Zhang, J.~Wang, H.~Ding, P.~Zhang, J.~Wen, Graph-based referring
  expression comprehension with expression-guided selective filtering and
  noun-oriented reasoning, Pattern Recognition 161 (2025) 111222.
\newblock \href {https://doi.org/10.1016/j.patcog.2024.111222}
  {\path{doi:10.1016/j.patcog.2024.111222}}.

\bibitem{cheng2026plrvg}
W.~Cheng, M.~Dai, W.~Yang, {PLRVG}: Progressive layer-wise refinement for
  visual grounding via deep-to-shallow decoding, Pattern Recognition 179 (2026)
  113555.
\newblock \href {https://doi.org/10.1016/j.patcog.2026.113555}
  {\path{doi:10.1016/j.patcog.2026.113555}}.

\bibitem{wang2027plvl}
J.~Wang, H.~Wang, W.~Zhang, K.~Ji, D.~Huang, Y.~Zheng, Progressive
  language-guided visual learning for multi-task visual grounding, Pattern
  Recognition 183 (2027) 115117.
\newblock \href {https://doi.org/10.1016/j.patcog.2026.115117}
  {\path{doi:10.1016/j.patcog.2026.115117}}.

\bibitem{he2023grec}
S.~He, H.~Ding, C.~Liu, X.~Jiang, {GREC}: Generalized referring expression
  comprehension (2023).
\newblock \href {https://doi.org/10.48550/arXiv.2308.16182}
  {\path{doi:10.48550/arXiv.2308.16182}}.

\bibitem{huang2027tave}
W.~Huang, K.~Chen, J.~Wang, M.~Rashid, Y.~Li, R.~Shao, Z.~Tian, Mitigating
  hallucinations in multimodal large language models via token-aware visual
  enhancement, Pattern Recognition 183 (2027) 115143.
\newblock \href {https://doi.org/10.1016/j.patcog.2026.115143}
  {\path{doi:10.1016/j.patcog.2026.115143}}.

\bibitem{wang2026safeground}
Q.~Wang, Y.~Fan, X.~E. Wang, {SafeGround}: Know when to trust {GUI} grounding
  models via uncertainty calibration (2026).
\newblock \href {https://doi.org/10.48550/arXiv.2602.02419}
  {\path{doi:10.48550/arXiv.2602.02419}}.

\bibitem{yang2026acs}
T.~Yang, Y.~Shi, R.~Sun, N.~Liu, J.~Sun, Self-improving small object grounding
  in {LVLMs} (2026).
\newblock \href {https://doi.org/10.48550/arXiv.2606.01612}
  {\path{doi:10.48550/arXiv.2606.01612}}.

\bibitem{liu2023dissonance}
K.~Liu, S.~Casper, D.~Hadfield-Menell, J.~Andreas, Cognitive dissonance: Why do
  language model outputs disagree with internal representations of
  truthfulness?, in: Proceedings of the 2023 Conference on Empirical Methods in
  Natural Language Processing, 2023.
\newblock \href {https://doi.org/10.18653/v1/2023.emnlp-main.291}
  {\path{doi:10.18653/v1/2023.emnlp-main.291}}.

\bibitem{ax2603.05465}
S.~A. Kogilathota, S.~V.~E. G, L.~Sun, J.~Zhou, {HALP}: Detecting
  hallucinations in vision-language models without generating a single token,
  in: Proceedings of the 19th Conference of the European Chapter of the
  Association for Computational Linguistics (EACL), 2026, pp. 6067--6085.
\newblock \href {https://doi.org/10.18653/v1/2026.eacl-long.287}
  {\path{doi:10.18653/v1/2026.eacl-long.287}}.

\bibitem{fime2026trajectory}
A.~A. Fime, M.~Z. Hossain, A.~Imteaj, Do {LVLMs} know when they are wrong?
  {T}rajectory-based correctness estimation from a single forward pass, in:
  Proceedings of the 2026 Conference on Empirical Methods in Natural Language
  Processing, Association for Computational Linguistics, 2026, pp.
  56033--56050.
\newblock \href {https://doi.org/10.18653/2026.emnlp-main.2468}
  {\path{doi:10.18653/2026.emnlp-main.2468}}.

\bibitem{lee2018mahalanobis}
K.~Lee, K.~Lee, H.~Lee, J.~Shin, A simple unified framework for detecting
  out-of-distribution samples and adversarial attacks, in: Advances in Neural
  Information Processing Systems, Vol.~31, 2018.

\bibitem{du2024haloscope}
X.~Du, C.~Xiao, Y.~Li, {HaloScope}: Harnessing unlabeled {LLM} generations for
  hallucination detection, in: Advances in Neural Information Processing
  Systems, Vol.~37, 2024, pp. 102948--102972, arXiv:2409.17504.
\newblock \href {https://doi.org/10.52202/079017-3270}
  {\path{doi:10.52202/079017-3270}}.

\bibitem{jiang2025vlinterp}
N.~Jiang, A.~Kachinthaya, S.~Petryk, Y.~Gandelsman, Interpreting and editing
  vision-language representations to mitigate hallucinations, in: International
  Conference on Learning Representations (ICLR), 2025, arXiv:2410.02762.

\bibitem{li2026ola}
H.~Li, Y.~Liu, J.~Zhu, S.~Yan, X.~Li, J.~Sun, Z.~Ge, Z.~Yu, Latent ordinal
  evidence, misaligned outputs: Inference-time ordinal lens alignment for
  multimodal {LLMs}, in: Proceedings of the 2026 Conference on Empirical
  Methods in Natural Language Processing, Association for Computational
  Linguistics, 2026, pp. 21328--21346.
\newblock \href {https://doi.org/10.18653/2026.emnlp-main.927}
  {\path{doi:10.18653/2026.emnlp-main.927}}.

\bibitem{ax2609.32757}
D.~E. Juanico, Readout is not recovery: Dissociating coordinate emission from
  visual-corruption repair in vision-language models (2026).
\newblock \href {https://doi.org/10.48550/arXiv.2609.32757}
  {\path{doi:10.48550/arXiv.2609.32757}}.

\bibitem{park2026vauq}
S.~Park, C.~Oh, H.~K. Choi, S.~Du, S.~Li, {VAUQ}: Vision-aware uncertainty
  quantification for {LVLM} self-evaluation, in: Findings of the Association
  for Computational Linguistics: ACL 2026, Association for Computational
  Linguistics, 2026, pp. 26534--26550.
\newblock \href {https://doi.org/10.18653/v1/2026.findings-acl.1321}
  {\path{doi:10.18653/v1/2026.findings-acl.1321}}.

\bibitem{jiang2018iounet}
B.~Jiang, R.~Luo, J.~Mao, T.~Xiao, Y.~Jiang, Acquisition of localization
  confidence for accurate object detection, in: European Conference on Computer
  Vision (ECCV), Lecture Notes in Computer Science, Springer, 2018, pp.
  816--832.
\newblock \href {https://doi.org/10.1007/978-3-030-01264-9_48}
  {\path{doi:10.1007/978-3-030-01264-9_48}}.

\bibitem{kao2018localization}
C.-C. Kao, T.-Y. Lee, P.~Sen, M.-Y. Liu, Localization-aware active learning for
  object detection, in: Computer Vision -- ACCV 2018, Lecture Notes in Computer
  Science, Springer, 2019, pp. 506--522.
\newblock \href {https://doi.org/10.1007/978-3-030-20876-9_32}
  {\path{doi:10.1007/978-3-030-20876-9_32}}.

\bibitem{zhou2027edl}
X.~Zhou, T.~Tang, Y.~Li, J.~Heikkil{\"a}, L.~Liu, G.~Kuang, Reactivating
  pre-trained backbone uncertainty for post-hoc evidential deep learning,
  Pattern Recognition 184 (2027) 115301.
\newblock \href {https://doi.org/10.1016/j.patcog.2026.115301}
  {\path{doi:10.1016/j.patcog.2026.115301}}.

\bibitem{angelopoulos2021ltt}
A.~N. Angelopoulos, S.~Bates, E.~J. Cand{\`e}s, M.~I. Jordan, L.~Lei, Learn
  then test: Calibrating predictive algorithms to achieve risk control, The
  Annals of Applied Statistics 19~(2) (2025) 1641--1662.
\newblock \href {https://doi.org/10.1214/24-AOAS1998}
  {\path{doi:10.1214/24-AOAS1998}}.

\bibitem{yu2016refcoco}
L.~Yu, P.~Poirson, S.~Yang, A.~C. Berg, T.~L. Berg, Modeling context in
  referring expressions, in: Computer Vision -- ECCV 2016, Lecture Notes in
  Computer Science, Springer International Publishing, Cham, 2016, pp. 69--85.
\newblock \href {https://doi.org/10.1007/978-3-319-46475-6_5}
  {\path{doi:10.1007/978-3-319-46475-6_5}}.

\bibitem{mao2016refcocog}
J.~Mao, J.~Huang, A.~Toshev, O.~Camburu, A.~Yuille, K.~Murphy, Generation and
  comprehension of unambiguous object descriptions, in: 2016 {IEEE} Conference
  on Computer Vision and Pattern Recognition ({CVPR}), 2016, pp. 11--20.
\newblock \href {https://doi.org/10.1109/CVPR.2016.9}
  {\path{doi:10.1109/CVPR.2016.9}}.

\bibitem{chen2024refl4}
J.~Chen, F.~Wei, J.~Zhao, S.~Song, B.~Wu, Z.~Peng, S.-H.~G. Chan, H.~Zhang,
  Revisiting referring expression comprehension evaluation in the era of large
  multimodal models, in: Proceedings of the IEEE/CVF Conference on Computer
  Vision and Pattern Recognition Workshops (CVPRW), 2025, pp. 513--524.
\newblock \href {https://doi.org/10.1109/CVPRW67362.2025.00056}
  {\path{doi:10.1109/CVPRW67362.2025.00056}}.

\bibitem{li2024vrsbench}
X.~Li, J.~Ding, M.~Elhoseiny, {VRSBench}: A versatile vision-language benchmark
  dataset for remote sensing image understanding, in: Advances in Neural
  Information Processing Systems (Datasets and Benchmarks Track), Vol.~37,
  2024, pp. 3229--3242.
\newblock \href {http://arxiv.org/abs/2406.12384} {\path{arXiv:2406.12384}},
  \href {https://doi.org/10.52202/079017-0106}
  {\path{doi:10.52202/079017-0106}}.

\bibitem{zhang2025mllmsknow}
J.~Zhang, M.~Khayatkhoei, P.~Chhikara, F.~Ilievski, {MLLMs} know where to look:
  Training-free perception of small visual details with multimodal {LLMs}, in:
  International Conference on Learning Representations (ICLR), 2025,
  arXiv:2502.17422.

\end{thebibliography}


\begin{thebibliography}{10}
\expandafter\ifx\csname url\endcsname\relax
  \def\url#1{\texttt{#1}}\fi
\expandafter\ifx\csname urlprefix\endcsname\relax\def\urlprefix{URL }\fi
\expandafter\ifx\csname href\endcsname\relax
  \def\href#1#2{#2} \def\path#1{#1}\fi

\bibitem{cui2026leftright}
E.~Cui, J.~Yang, Y.~Lin, Y.~Feng, Z.~Chen, Where do vision-language models get
  left and right? {P}osition-code counterfactuals in {Qwen2.5-VL}, {Qwen3-VL}
  and {InternVL3} (2026).
\newblock \href {https://doi.org/10.48550/arXiv.2611.14060}
  {\path{doi:10.48550/arXiv.2611.14060}}.

\bibitem{kossen2024sep}
J.~Kossen, J.~Han, M.~Razzak, L.~Schut, S.~Malik, Y.~Gal, Semantic entropy
  probes: Robust and cheap hallucination detection in {LLMs} (2024).
\newblock \href {https://doi.org/10.48550/arXiv.2406.15927}
  {\path{doi:10.48550/arXiv.2406.15927}}.

\bibitem{liu2024groundingdino}
S.~Liu, Z.~Zeng, T.~Ren, F.~Li, H.~Zhang, J.~Yang, Q.~Jiang, C.~Li, J.~Yang,
  H.~Su, J.~Zhu, L.~Zhang, Grounding {DINO}: Marrying {DINO} with grounded
  pre-training for open-set object detection, in: Computer Vision -- {ECCV}
  2024, Lecture Notes in Computer Science, 2024, pp. 38--55.
\newblock \href {https://doi.org/10.1007/978-3-031-72970-6_3}
  {\path{doi:10.1007/978-3-031-72970-6_3}}.

\bibitem{ax2607.05978}
D.~Shalam, E.~B. Baruch, A.~B. Cohen, T.~Remez, Propose and attend:
  Training-free {MLLM} grounding confidence via multi-token localized attention
  (2026).
\newblock \href {https://doi.org/10.48550/arXiv.2607.05978}
  {\path{doi:10.48550/arXiv.2607.05978}}.

\bibitem{ax2606.25760}
D.~Kumar, S.~Tayebati, D.~Naik, A.~S. Rios, N.~Ahuja, O.~Tickoo, R.~Krishnan,
  A.~R. Trivedi, Uncertainty quantification for computer-use agents: A
  benchmark across vision-language models and {GUI} grounding datasets, in:
  Advances in Neural Information Processing Systems, 2026, arXiv:2606.25760.

\bibitem{wang2025coe}
Y.~Wang, P.~Zhang, B.~Yang, D.~F. Wong, R.~Wang, Latent space
  chain-of-embedding enables output-free {LLM} self-evaluation, in:
  International Conference on Learning Representations (ICLR), 2025.

\bibitem{sriramanan2024llmcheck}
G.~Sriramanan, S.~Bharti, V.~S. Sadasivan, S.~Saha, P.~Kattakinda, S.~Feizi,
  {LLM-Check}: Investigating detection of hallucinations in large language
  models, in: Advances in Neural Information Processing Systems, Vol.~37, 2024,
  pp. 34188--34216.
\newblock \href {https://doi.org/10.52202/079017-1077}
  {\path{doi:10.52202/079017-1077}}.

\bibitem{belrose2023tunedlens}
N.~Belrose, I.~Ostrovsky, L.~McKinney, Z.~Furman, L.~Smith, D.~Halawi,
  S.~Biderman, J.~Steinhardt, Eliciting latent predictions from transformers
  with the tuned lens (2023).
\newblock \href {https://doi.org/10.48550/arXiv.2303.08112}
  {\path{doi:10.48550/arXiv.2303.08112}}.

\bibitem{jiang2025vlinterp}
N.~Jiang, A.~Kachinthaya, S.~Petryk, Y.~Gandelsman, Interpreting and editing
  vision-language representations to mitigate hallucinations, in: International
  Conference on Learning Representations (ICLR), 2025, arXiv:2410.02762.

\bibitem{lee2018mahalanobis}
K.~Lee, K.~Lee, H.~Lee, J.~Shin, A simple unified framework for detecting
  out-of-distribution samples and adversarial attacks, in: Advances in Neural
  Information Processing Systems, Vol.~31, 2018.

\bibitem{ax2604.19444}
T.~Zollo, J.~Wang, R.~Zemel, Binary agreement for efficient unsupervised
  confidence calibration of reasoning {LLMs} (2026).
\newblock \href {https://doi.org/10.48550/arXiv.2604.19444}
  {\path{doi:10.48550/arXiv.2604.19444}}.

\bibitem{azaria2023saplma}
A.~Azaria, T.~Mitchell, The internal state of an {LLM} knows when it's lying,
  in: Findings of the Association for Computational Linguistics: EMNLP 2023,
  2023, pp. 967--976.
\newblock \href {https://doi.org/10.18653/v1/2023.findings-emnlp.68}
  {\path{doi:10.18653/v1/2023.findings-emnlp.68}}.

\bibitem{ax2603.05465}
S.~A. Kogilathota, S.~V.~E. G, L.~Sun, J.~Zhou, {HALP}: Detecting
  hallucinations in vision-language models without generating a single token,
  in: Proceedings of the 19th Conference of the European Chapter of the
  Association for Computational Linguistics (EACL), 2026, pp. 6067--6085.
\newblock \href {https://doi.org/10.18653/v1/2026.eacl-long.287}
  {\path{doi:10.18653/v1/2026.eacl-long.287}}.

\bibitem{ax2605.10893}
R.~Khanmohammadi, E.~Miahi, S.~Kaur, C.~Smiley, I.~Brugere, K.~S. Thind, M.~M.
  Ghassemi, Grounded or guessing? {LVLM} confidence estimation via blind-image
  contrastive ranking, in: Proceedings of the 15th International Joint
  Conference on Natural Language Processing and the 5th Conference of the
  Asia-Pacific Chapter of the Association for Computational Linguistics (Volume
  1: Long Papers), Association for Computational Linguistics, 2026, pp.
  4645--4721.
\newblock \href {https://doi.org/10.18653/2026.aacl-long.210}
  {\path{doi:10.18653/2026.aacl-long.210}}.

\bibitem{li2026pace}
Y.~Li, X.~Wu, H.~M. Meng, How many pixels is a digit worth? {P}lace-aware
  coordinate entropy for {GUI} agent confidence estimation, in: Proceedings of
  the 2026 Conference on Empirical Methods in Natural Language Processing,
  2026, pp. 10302--10315.
\newblock \href {https://doi.org/10.18653/2026.emnlp-main.448}
  {\path{doi:10.18653/2026.emnlp-main.448}}.

\bibitem{geifman2017selective}
Y.~Geifman, R.~El-Yaniv, Selective classification for deep neural networks, in:
  Advances in Neural Information Processing Systems, Vol.~30, Curran
  Associates, Inc., 2017.

\bibitem{angelopoulos2021ltt}
A.~N. Angelopoulos, S.~Bates, E.~J. Cand{\`e}s, M.~I. Jordan, L.~Lei, Learn
  then test: Calibrating predictive algorithms to achieve risk control, The
  Annals of Applied Statistics 19~(2) (2025) 1641--1662.
\newblock \href {https://doi.org/10.1214/24-AOAS1998}
  {\path{doi:10.1214/24-AOAS1998}}.

\bibitem{angelopoulos2024crc}
A.~N. Angelopoulos, S.~Bates, A.~Fisch, L.~Lei, T.~Schuster, Conformal risk
  control, in: International Conference on Learning Representations, 2024.

\bibitem{phillips2026entropy}
E.~Phillips, F.~K. Gustafsson, S.~Wu, A.~Thakur, D.~A. Clifton, Entropy alone
  is insufficient for safe selective prediction in {LLMs}, arXiv:2603.21172
  (2026).

\bibitem{xiong2024uqllm}
M.~Xiong, A.~Santilli, M.~Kirchhof, A.~Golinski, S.~Williamson,
  \href{https://machinelearning.apple.com/research/efficient-and-effective}{Efficient
  and effective uncertainty quantification in {LLMs}}, in: NeurIPS 2024
  Workshop on Safe Generative AI, 2024.
\newline\urlprefix\url{https://machinelearning.apple.com/research/efficient-and-effective}

\bibitem{ax2609.37488}
T.~Sato, T.~Sanda, K.~Maeda, T.~Ogawa, M.~Haseyama, S.~Nagashima, {FORUM}:
  Frozen outputs reconciled using model agreement for visual grounding, in:
  Asian Conference on Computer Vision (ACCV), 2026.

\bibitem{bolya2020tide}
D.~Bolya, S.~Foley, J.~Hays, J.~Hoffman, {TIDE}: A general toolbox for
  identifying object detection errors, in: Computer Vision -- {ECCV} 2020, Vol.
  12348 of Lecture Notes in Computer Science, Springer, 2020, pp. 558--573.
\newblock \href {https://doi.org/10.1007/978-3-030-58580-8_33}
  {\path{doi:10.1007/978-3-030-58580-8_33}}.

\bibitem{ax2607.04683}
K.~N.~H. Vo, A.~Vazhentsev, A.~Shelmanov, T.~Baldwin, Y.~Kementchedjhieva,
  Failing to see or failing to know? {Attributing} errors in vision-language
  models (2026).
\newblock \href {http://arxiv.org/abs/2607.04683} {\path{arXiv:2607.04683}}.

\bibitem{chen2024refl4}
J.~Chen, F.~Wei, J.~Zhao, S.~Song, B.~Wu, Z.~Peng, S.-H.~G. Chan, H.~Zhang,
  Revisiting referring expression comprehension evaluation in the era of large
  multimodal models, in: Proceedings of the IEEE/CVF Conference on Computer
  Vision and Pattern Recognition Workshops (CVPRW), 2025, pp. 513--524.
\newblock \href {https://doi.org/10.1109/CVPRW67362.2025.00056}
  {\path{doi:10.1109/CVPRW67362.2025.00056}}.

\bibitem{chen2024inside}
C.~Chen, K.~Liu, Z.~Chen, Y.~Gu, Y.~Wu, M.~Tao, Z.~Fu, J.~Ye, {INSIDE}: {LLMs}'
  internal states retain the power of hallucination detection, in:
  International Conference on Learning Representations, 2024.
\newblock \href {http://arxiv.org/abs/2402.03744} {\path{arXiv:2402.03744}}.

\bibitem{kang2025heads}
S.~Kang, J.-Y. Kim, J.~Kim, S.~J. Hwang, Your large vision-language model only
  needs a few attention heads for visual grounding, in: Proceedings of the
  IEEE/CVF Conference on Computer Vision and Pattern Recognition (CVPR), 2025,
  pp. 9339--9350.
\newblock \href {https://doi.org/10.1109/cvpr52734.2025.00872}
  {\path{doi:10.1109/cvpr52734.2025.00872}}.

\bibitem{li2024vrsbench}
X.~Li, J.~Ding, M.~Elhoseiny, {VRSBench}: A versatile vision-language benchmark
  dataset for remote sensing image understanding, in: Advances in Neural
  Information Processing Systems (Datasets and Benchmarks Track), Vol.~37,
  2024, pp. 3229--3242.
\newblock \href {http://arxiv.org/abs/2406.12384} {\path{arXiv:2406.12384}},
  \href {https://doi.org/10.52202/079017-0106}
  {\path{doi:10.52202/079017-0106}}.

\bibitem{ashuach2026masked}
T.~Ashuach, S.~Gretz, Y.~Katz, Y.~Belinkov, L.~Ein-Dor, Masked by consensus:
  Disentangling privileged knowledge in {LLM} correctness, in: Proceedings of
  the 64th Annual Meeting of the {A}ssociation for {C}omputational
  {L}inguistics (Volume 1: Long Papers), Association for Computational
  Linguistics, 2026, pp. 10577--10596.
\newblock \href {https://doi.org/10.18653/v1/2026.acl-long.483}
  {\path{doi:10.18653/v1/2026.acl-long.483}}.

\bibitem{ax2509.24988}
H.~Xiao, V.~Patil, H.~Lee, E.~Stengel-Eskin, M.~Bansal,
  \href{https://proceedings.mlr.press/v306/xiao26l.html}{Generalized
  correctness models: Learning calibrated and cross-model correctness
  predictors from historical patterns}, in: Proceedings of the 43rd
  International Conference on Machine Learning, Vol. 306 of Proceedings of
  Machine Learning Research, PMLR, 2026, pp. 138884--138908.
\newline\urlprefix\url{https://proceedings.mlr.press/v306/xiao26l.html}

\bibitem{martinezplumed2019irt}
F.~Mart{\'i}nez-Plumed, R.~B.~C. Prud{\^e}ncio, A.~Mart{\'i}nez-Us{\'o},
  J.~Hern{\'a}ndez-Orallo, Item response theory in {AI}: Analysing machine
  learning classifiers at the instance level, Artificial Intelligence 271
  (2019) 18--42.
\newblock \href {https://doi.org/10.1016/j.artint.2018.09.004}
  {\path{doi:10.1016/j.artint.2018.09.004}}.

\bibitem{ax2509.15339}
Y.~Seo, D.~Lee, J.~Yeo, Quantifying genuine awareness in hallucination
  prediction beyond question-side shortcuts (2025).
\newblock \href {https://doi.org/10.48550/arXiv.2509.15339}
  {\path{doi:10.48550/arXiv.2509.15339}}.

\bibitem{ax2610.05572}
M.~EsfandyariDoulabi, L.~Arkoh, B.~Tadesse, V.~Patel, M.~Sharma, M.~d'Amorim,
  W.~Assun{\c{c}}{\~a}o, Disentangling task difficulty from run-level failure
  in agent failure prediction (2026).
\newblock \href {https://doi.org/10.48550/arXiv.2610.05572}
  {\path{doi:10.48550/arXiv.2610.05572}}.

\bibitem{sharma2026probeeval}
M.~Sharma, Are hallucination probes evaluated correctly? {A} critical review
  and a pilot study under 4-bit quantization (2026).
\newblock \href {https://doi.org/10.5281/zenodo.24223138}
  {\path{doi:10.5281/zenodo.24223138}}.

\bibitem{green1966sdt}
D.~M. Green, J.~A. Swets, Signal Detection Theory and Psychophysics, Wiley, New
  York, 1966.

\bibitem{dorfman1992mrmc}
D.~D. Dorfman, K.~S. Berbaum, C.~E. Metz, Receiver operating characteristic
  rating analysis: Generalization to the population of readers and patients
  with the jackknife method, Investigative Radiology 27~(9) (1992) 723--731.
\newblock \href {https://doi.org/10.1097/00004424-199209000-00015}
  {\path{doi:10.1097/00004424-199209000-00015}}.

\end{thebibliography}
